\documentclass[journal]{IEEEtran}
\usepackage{amsmath,amsfonts}
\usepackage{array}
\usepackage{textcomp}
\usepackage{stfloats}
\usepackage{url}
\usepackage{verbatim}
\usepackage{graphicx}
\usepackage{amssymb}
\usepackage{hyperref}
\usepackage{graphicx}
\usepackage{amsmath}
\usepackage{longtable}
\usepackage{algorithm} 
\usepackage{algpseudocode} 
\usepackage{mathrsfs}
\usepackage{subcaption}
\usepackage{mathtools}
\usepackage{pifont}
\usepackage{color}
\usepackage{lineno}
\usepackage{makecell} 
\usepackage{pdflscape}
\usepackage{adjustbox}
\usepackage[utf8]{inputenc}
\usepackage{tabularx}
\usepackage{blindtext}
\usepackage{longtable}
\usepackage{lscape}
\usepackage{setspace}
\usepackage{notoccite} 
\usepackage{lscape} 
\usepackage{mwe}
\usepackage{booktabs}
\usepackage{amsthm}
\newtheorem{theorem}{Theorem}[section]

\theoremstyle{definition}

\theoremstyle{definition}
\newtheorem{definition}{Definition}[section]
\newcommand{\RNum}[1]{\lowercase\expandafter{\romannumeral #1\relax}}
\newcommand{\RNumU}[1]{\uppercase\expandafter{\romannumeral #1\relax}}
\usepackage[numbers]{natbib}
\def\BibTeX{{\rm B\kern-.05em{\sc i\kern-.025em b}\kern-.08em
    T\kern-.1667em\lower.7ex\hbox{E}\kern-.125emX}}
\usepackage{balance}
\begin{document}
\title{RoBoSS: A Robust, Bounded, Sparse, and Smooth Loss Function for Supervised Learning}
\author{Mushir Akhtar, \IEEEmembership{Graduate Student Member,~IEEE}, M. Tanveer{$^*$} \IEEEmembership{Senior Member,~IEEE}, Mohd. Arshad
\thanks{ \noindent $^*$Corresponding author\\
    Mushir Akhtar, M. Tanveer and Mohd. Arshad are with the Department of Mathematics, Indian Institute of Technology Indore, Simrol, Indore, 453552, India (e-mail: phd2101241004@iiti.ac.in, mtanveer@iiti.ac.in, arshad@iiti.ac.in).}}

\maketitle

\begin{abstract}
In the domain of machine learning, the significance of the loss function is paramount, especially in supervised learning tasks. It serves as a fundamental pillar that profoundly influences the behavior and efficacy of supervised learning algorithms. Traditional loss functions, though widely used, often struggle to handle outlier-prone and high-dimensional data, resulting in suboptimal outcomes and slow convergence during training. In this paper, we address the aforementioned constraints by proposing a novel robust, bounded, sparse, and smooth (RoBoSS) loss function for supervised learning. Further, we incorporate the RoBoSS loss within the framework of support vector machine (SVM) and introduce a new robust algorithm named $\mathcal{L}_{RoBoSS}$-SVM. For the theoretical analysis, the classification-calibrated property and generalization ability are also presented. These investigations are crucial for gaining deeper insights into the robustness of the RoBoSS loss function in classification problems and its potential to generalize well to unseen data. To validate the potency of the proposed $\mathcal{L}_{RoBoSS}$-SVM, we assess it on $88$ benchmark datasets from KEEL and UCI repositories.
Further, to rigorously evaluate its performance in challenging scenarios, we conducted an assessment using datasets intentionally infused with outliers and label noise. 
Additionally, to exemplify the effectiveness of $\mathcal{L}_{RoBoSS}$-SVM within the biomedical domain, we evaluated it on two medical datasets: the electroencephalogram (EEG) signal dataset and the breast cancer (BreaKHis) dataset. The numerical results substantiate the superiority of the proposed $\mathcal{L}_{RoBoSS}$-SVM model, both in terms of its remarkable generalization performance and its efficiency in training time. The code of the $\mathcal{L}_{RoBoSS}$-SVM is publicly accessible at \url{https://github.com/mtanveer1/RoBoSS}.
\end{abstract}
\section {Introduction and Motivation} 
\IEEEPARstart{D}{ata} analysis tasks such as classification and regression fall under the umbrella of supervised machine learning (SML). SML is a powerful paradigm in machine learning wherein a model learns from labeled data to make predictions on unseen instances. Key to this process is the concept of loss functions, which quantify the discrepancy between predicted and actual outputs.
Support vector machine (SVM) \cite{cortes1995support} represents an efficient SML algorithm. It is based on the concept of structural risk minimization (SRM) and is rooted in statistical learning theory (SLT) \cite{vapnik1999nature}, providing it with a robust theoretical base and strong generalization capabilities.
In this paper, we undertake an in-depth examination of the interrelation between loss functions and the supervised learning algorithm, utilizing the framework of SVM.
\par
This study is solely focused on the binary classification task. Let the training set be defined by  $\left\{x_k,y_k\right\}_{k=1}^n$, where $x_k \in \mathbb{R}^m$ indicates the sample vector and $y_k \in\{1,-1\}$ indicates the corresponding label of the class. The aim of SVM is to construct a decision hyperplane $w^\intercal x+b=0$ with bias $b \in \mathbb{R} $ and weight vector $w \in \mathbb{R}^m $, which are estimated by training data.
When predicting the class label $\hat{y}$ for a test data point $\hat{x}$, it is assigned a value of $-1$ if $w^\intercal \hat{x}+b < 0$, and $1$ otherwise.
To determine the best hyperplane, we examine two cases within the input space: datasets that are linearly separable and those that are not.
\par
In the case of linearly separable situation, the hyperplane parameters $w$ and $b$ are determined by solving the following optimization problem:
\begin{align} \label{hardmarginSVM}
\underset{ w, b}{min} \hspace{0.2cm} &\frac{1}{2}\|w\|^2 \nonumber \\
 \text {s.t.}\hspace{0.2cm}  & y_k\left(w^\intercal x_k+b\right) \geq 1, ~\forall~ k=1,2, \ldots,n.
\end{align}
The model in equation (\ref{hardmarginSVM}) is termed the hard-margin SVM since it necessitates every training sample to be correctly classified.
\par
For linearly inseparable situation, the widely used approach permits misclassification and penalizes these violations by including the loss function, leading to the following optimization task:
\begin{align} \label{softmarginSVM}
\underset{ w, b}{min} \hspace{0.2cm} &\frac{1}{2}\|w\|^2 + \frac{\mathcal{C}}{n} \sum_{k=1}^n \mathcal{L}\biggl(1-y_k \left(w^\intercal x_k+b\right) \biggr), 
\end{align}
where $\mathcal{C} > 0$ is a trade-off parameter and $\mathcal{L}(u)$ with $u$:$=1-y_k \left(w^\intercal x_k+b\right)$ represents the loss function. Since model (\ref{softmarginSVM}) allows misclassification of samples, it is referred to as a soft-margin SVM model \cite{cortes1995support}.
\par
The loss function $\mathcal{L}(u)$ is an essential component of support vector machine,  which controls the robustness and sparsity of SVM. The ``$0$-$1$" loss function is defined as an ideal loss function \cite{cortes1995support} that assigns a fixed loss of $1$ to all misclassified samples and no loss to correctly classified samples. It is defined as follows:
\begin{align}    
\mathcal{L}_{0-1}(u)=
\begin{cases}
1, & u > 0, \\
0, & u \leq 0.
\end{cases}
\end{align}
However, solving SVM with $0$-$1$ loss function is NP-hard \cite{natarajan1995sparse, amaldi1998approximability}, since it is discontinuous and non-convex.
\begin{figure*}
\centering
\subcaptionbox{\label{fig:Hinge_loss }}{%
\includegraphics[width=0.32\textwidth,keepaspectratio]{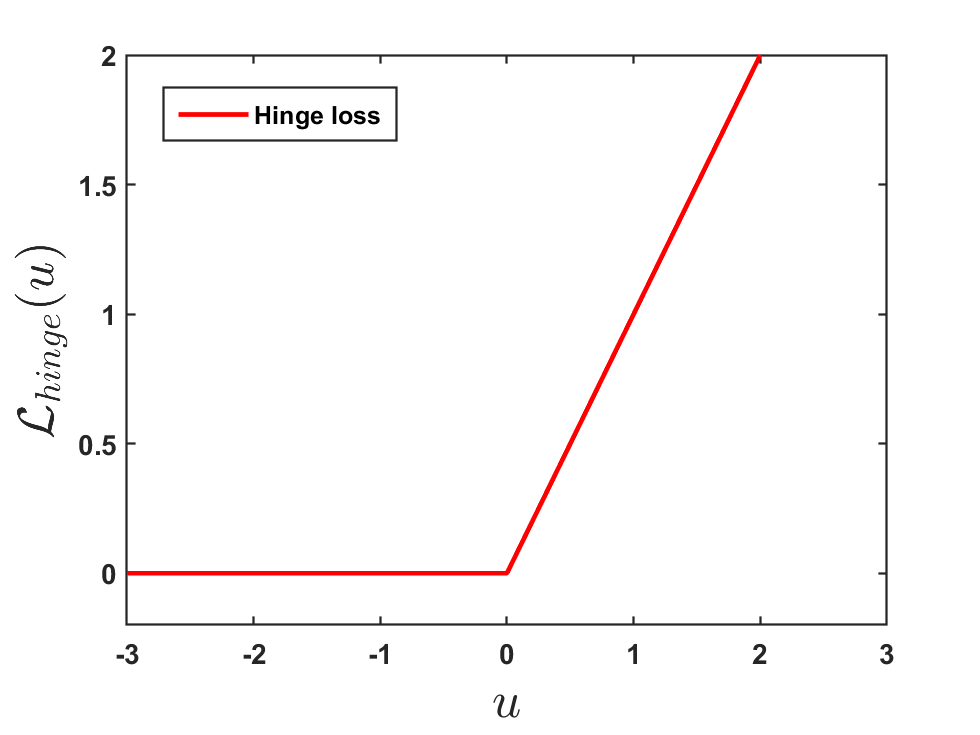}}
\hfill
\subcaptionbox{\label{fig:Pinball loss }} {%
\includegraphics[width=0.32\textwidth,keepaspectratio]{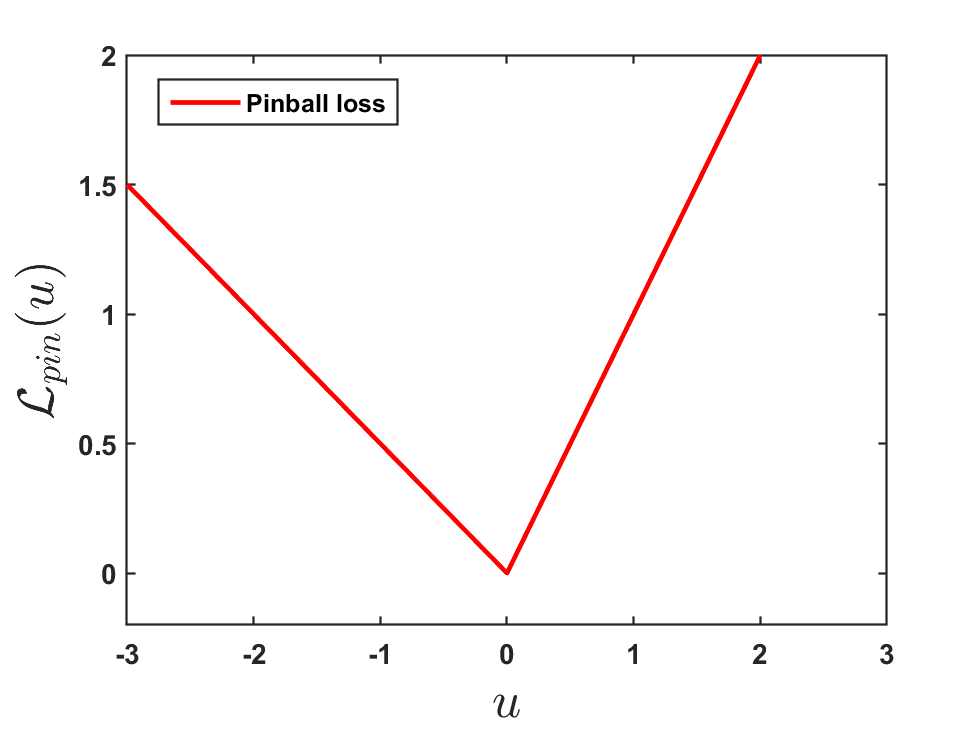}}
\hfill
\subcaptionbox{\label{fig:Truncated hinge loss}} {%
\includegraphics[width=0.32\textwidth,keepaspectratio]{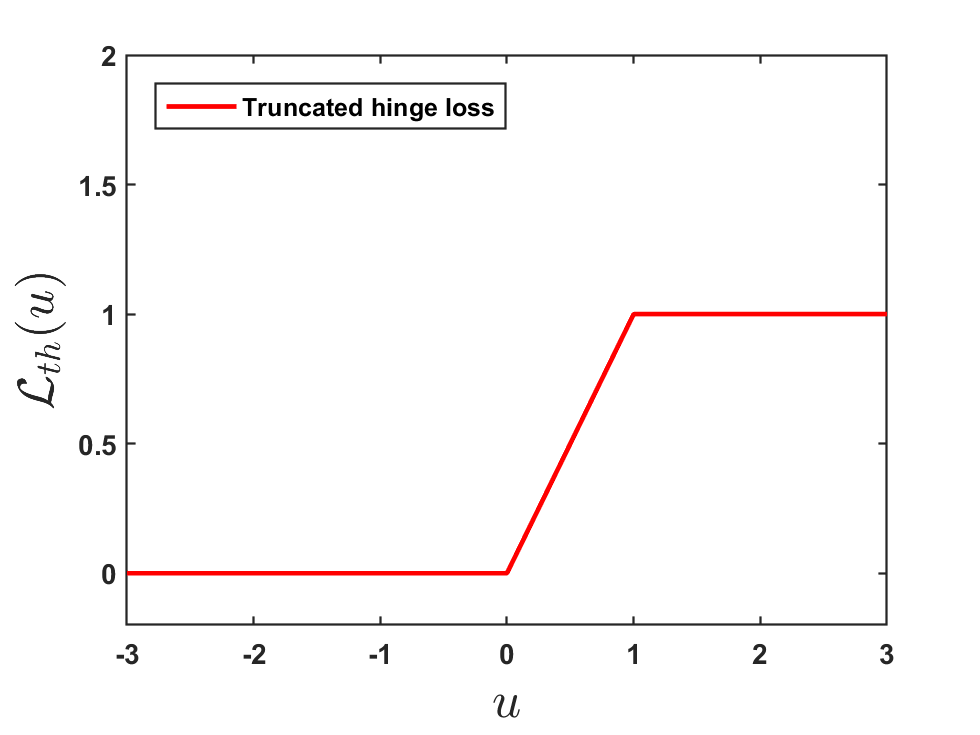}}
\\      
\subcaptionbox{\label{fig:Truncated pinball loss}} {%
\includegraphics[width=0.32\textwidth,keepaspectratio]{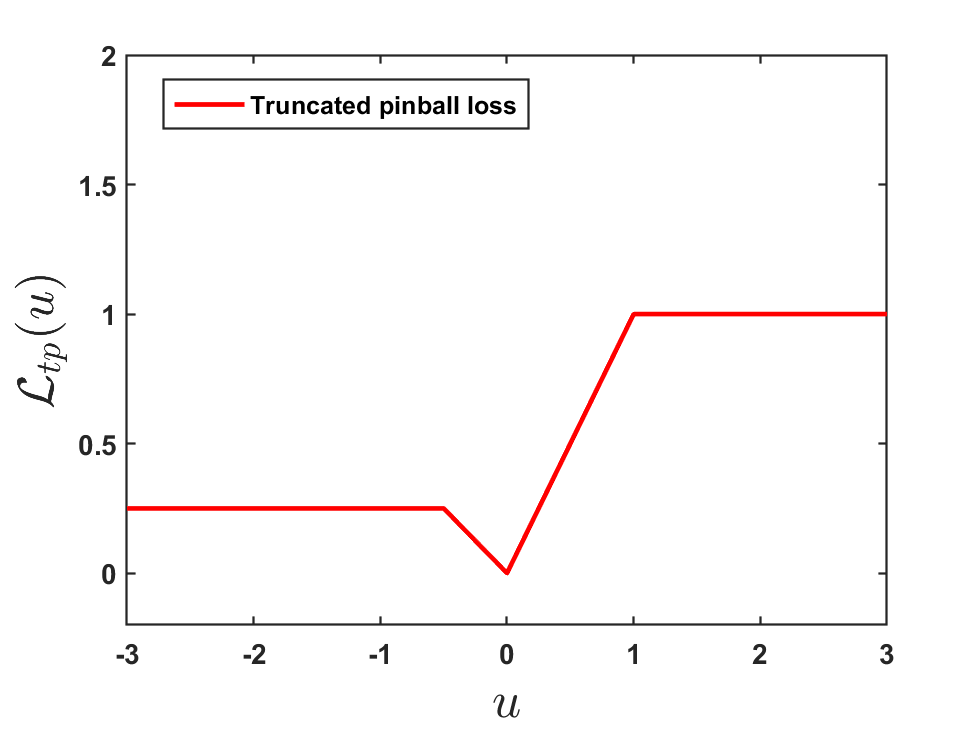}}
\hfill
\subcaptionbox{\label{fig:RoBoSS loss with fixed lambda}} {%
\includegraphics[width=0.32\textwidth,keepaspectratio]{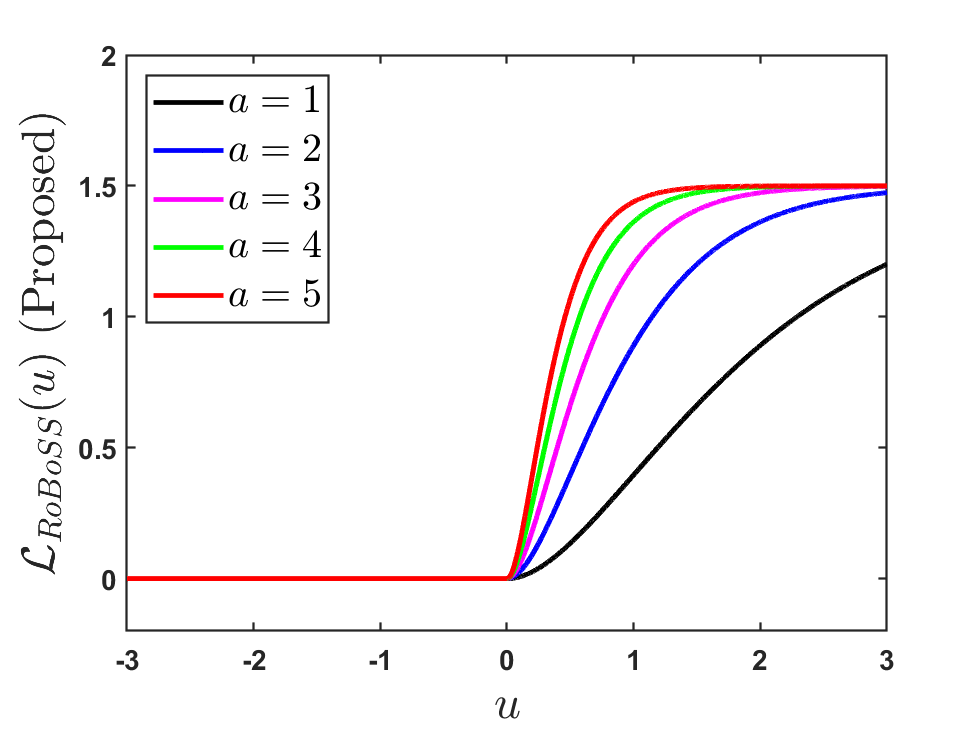}}
\hfill
\subcaptionbox{\label{fig: wiRoBoSth fixed a}} {%
\includegraphics[width=0.32\textwidth,keepaspectratio]{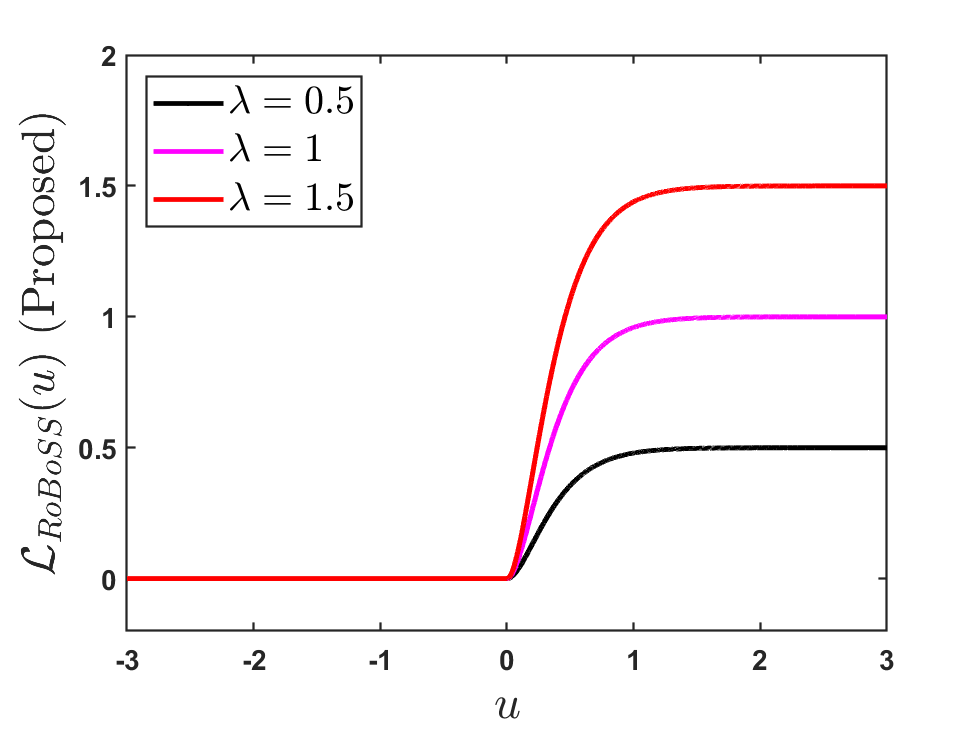}}
\caption{(a) Hinge loss function. (b) Pinball loss function with $\tau=0.5$. (c) Truncated hinge loss with $\delta=1$. (d) Truncated pinball loss with $\tau=0.5$, $\delta_1=1$, and $\delta_2=0.25$. (e) Proposed RoBoSS loss with fixed $\lambda=1.5$ and different values of $a$. (f) Proposed RoBoSS loss with fixed $a=5$ and different values of $\lambda$.}
\label{fig: baseline and proposed Loss}
\end{figure*}
For the development of SVM, a great deal of work has gone into constructing new loss functions to obtain new effective soft-margin SVM models. Here, we briefly reviewed a few renowned loss functions, which are sufficient to serve as inspiration for the rest of this paper.
\par
The first soft-margin SVM model is hinge loss SVM ($\mathcal{L}_{hinge}$-SVM) \cite{cortes1995support}, which utilizes the hinge loss function $\mathcal{L}_{hinge}(u)$ (see Fig. \ref{fig:Hinge_loss }), and is defined as: 
\begin{align}    
\mathcal{L}_{hinge}(u)=
\begin{cases}
u, & u > 0, \\
0, & u \leq 0. 
\end{cases}
\end{align}
The hinge loss function is convex, non-smooth, and unbounded.
To improve the efficacy of $\mathcal{L}_{hinge}$-SVM, \citet{huang2013support} studied pinball loss SVM ($\mathcal{L}_{pin}$-SVM), which utilizes pinball loss function $\mathcal{L}_{pin}(u)$ {(see Fig. \ref{fig:Pinball loss }) and is defined as: 
\begin{align}    
\mathcal{L}_{pin}(u)=
\begin{cases}
u, & u > 0, \\
-\tau u, & u \leq 0, 
\end{cases}
\end{align}
where $\tau \in \left[0,1\right]$. For $\tau=0$, the pinball loss function is reduced to the hinge loss function. For $\tau \in \left(0,1\right]$, it also provides penalty to correctly classified samples, which diminishes the sparseness \cite{tanveer2021sparse}. The pinball loss is likewise characterized by its convexity, non-smooth nature, and lack of boundedness.
Some other convex loss functions are least square loss function \cite{suykens1999least}, generalized hinge loss function \cite{bartlett2008classification}, LINEX loss function \cite{ma2019linex}, huberized pinball loss function \cite{zhu2020support}, and so on.
\par
The convexity of loss functions is acknowledged as highly regarded due to its computational benefits. Specifically, convex loss functions possess unique optima, are easy to use, and can be efficiently optimized using convex optimization tools. However, the convex loss functions provide poor approximations of $0$-$1$ loss function and exhibit a lack of robustness to outliers due to their unbounded nature, which makes the corresponding classifier susceptible to being overly influenced or dominated by outliers \cite{zhao2010convex}. To improve the robustness, various bounded loss functions are suggested in the literature. In order to increase the robustness of $\mathcal{L}_{hinge}$-SVM, \citet{wu2007robust} developed truncated hinge loss function $\mathcal{L}_{th}(u)$ (see Fig. \ref{fig:Truncated hinge loss}), which is formulated as: 
\begin{align}    
\mathcal{L}_{th}(u)=
\begin{cases}
\delta, & u \geq \delta, \\
u, & u \in \left(0,\delta\right),\\
0, & u \leq 0,
\end{cases}
\end{align}
where $\delta \geq 1$.
It is non-convex, non-smooth, and bounded. Other relevant research focuses on the development of new algorithms for solving truncated hinge loss SVM, such as the branch and bound algorithm \cite{brooks2011support}, the convex-concave procedure (CCCP) \cite{shen2003psi}, and so on.
To enhance the robustness and sparseness of $\mathcal{L}_{pin}$-SVM, \citet{yang2018support} proposed the truncated pinball loss function $\mathcal{L}_{tp}(u)$ (see Fig. \ref{fig:Truncated pinball loss}), and is defined as:
\begin{align}    
\mathcal{L}_{tp}(u)=
\begin{cases}
\delta_1, & u \geq \delta_1, \\
 u, & u \in \left[0 ,\delta_1\right),\\
-\tau u, &  u \in \left(-\delta_2/\tau ,0\right),\\
\delta_2, & u \leq -\delta_2/\tau,
\end{cases}
\end{align}
where $\tau \in [0,1]$, and $\delta_1, \delta_2 > 0$. It gives a fixed loss $\delta_1$ for samples with $u \geq -\delta_1$, which enhances the robustness and a fixed loss $\delta_2$ for samples with $u \leq -\delta_2/\tau$, which adds the sparseness to $\mathcal{L}_{pin}$-SVM. It is also non-convex, non-smooth, and bounded. The optimization of truncated pinball loss SVM is addressed by the popular and efficient CCCP algorithm. The non-convex and non-smooth nature of the aforementioned loss functions poses significant challenges in terms of computational optimization for solving corresponding SVM models.
\par
Motivated by the previous works, the main focus of this paper is to construct a new robust, bounded, sparse, and smooth loss function for supervised learning.
To improve the robustness, sparsity, and smoothness of the aforementioned losses, we design a new loss function named RoBoSS loss function (see Fig. \ref{fig:RoBoSS loss with fixed lambda} and \ref{fig: wiRoBoSth fixed a}), which is described as:
\begin{align} \label{proposedloss}     
\mathcal{L}_{RoBoSS}(u)=
\begin{cases}
\lambda\{1-(au+1)exp(-au)\}, & u > 0, \\
0, & u \leq 0,
\end{cases}
\end{align}
where $a, \lambda >0$ represent the shape and bound parameters, respectively.
Further, we amalgamate the proposed RoBoSS loss in SVM and introduce a new robust SVM model termed $\mathcal{L}_{RoBoSS}$-SVM. By replacing $\mathcal{L}(\cdot)$ by $\mathcal{L}_{RoBoSS}(\cdot)$ in (\ref{softmarginSVM}) yields us to get the proposed $\mathcal{L}_{RoBoSS}$-SVM model, which is given by
\begin{align} \label{ProposedSVM}
\underset{ w, b}{min} \hspace{0.2cm} &\frac{1}{2}\|w\|^2 + \frac{\mathcal{C}}{n} \sum_{k=1}^n \mathcal{L}_{RoBoSS}\biggl(1-y_k \left(w^\intercal x_k+b\right) \biggr).
\end{align}
The non-convex nature of the proposed loss function poses challenges for optimizing the $\mathcal{L}_{RoBoSS}$-SVM by the Wolfe-dual method. However, the smooth nature of $\mathcal{L}_{RoBoSS}$-SVM  enables the use of gradient-based fast optimization techniques for solving the model. In this paper, we utilize the Nestrov accelerated gradient (NAG) based framework to solve the optimization problem of $\mathcal{L}_{RoBoSS}$-SVM. NAG is known for its low computational complexity and efficiency in handling large-scale problems \cite{ruder2016overview}.
The main contributions of this work can be summarized as follows:
\begin{itemize}
    \item We introduce an innovative advancement in the field of supervised learning: the RoBoSS (Robust, Bounded, Sparse, and Smooth) loss function.
    \item We explored the theoretical aspects of the RoBoSS loss and showed it possesses two crucial properties: classification-calibration and a bound on generalization error. These results not only emphasize the robustness of the RoBoSS loss function but also provide valuable insights into its performance and applicability.
    \item We fuse the RoBoSS loss within the SVM framework and introduce a novel SVM model coined as $\mathcal{L}_{RoBoSS}$-SVM. The resulting $\mathcal{L}_{RoBoSS}$-SVM model harnesses the inherent strengths of both the RoBoSS loss function and the SVM algorithm, leading to an advanced and versatile machine learning tool.
    \item We carried out numerical experiments on 88 benchmark KEEL and UCI datasets from diverse domains. The outcomes validate the effectiveness of the $\mathcal{L}_{RoBoSS}$-SVM model when compared to the baseline models.
    \item Furthermore, to showcase the prowess of the $\mathcal{L}_{RoBoSS}$-SVM in the biomedical domain, we executed additional evaluations on two biomedical datasets: the electroencephalogram (EEG) signal dataset and the breast cancer (BreaKHis) dataset. These experiments provide evidence of the model's efficiency in the biomedical realm.
\end{itemize}
\section{Proposed work}
In this work, we present a significant advancement in supervised learning: a new loss function characterized by robustness, boundedness, sparsity, and smoothness, termed the RoBoSS loss (see Fig. \ref{fig:RoBoSS loss with fixed lambda} and \ref{fig: wiRoBoSth fixed a}). This innovative approach represents a substantial stride in optimizing the training process of machine learning models. The equation (\ref{proposedloss}) provides the mathematical formulation of the RoBoSS loss introduced in this study. The RoBoSS loss function, as put forth in this work, exhibits the subsequent characteristics:
\begin{itemize}
    \item It is robust and sparse. By setting an upper bound $\lambda$ and capping the loss for samples with $u>0$ beyond a certain margin, robustness is enhanced. Additionally, it assigns a fixed loss of $0$ for all samples with $u \leq 0$, thereby introducing sparsity.
    \item It is smooth, bounded, and non-convex.
    \item It has two beneficial parameters, $a$ and $\lambda$, known as the shape parameter and bounding parameter, respectively. The shape parameter ($a$) controls the intensity of the penalty, while the bounding parameter ($\lambda$) defines the limits for the loss values.
    \item For $\lambda=1$, when $a  \rightarrow +\infty$, it approaches the ``$0-1$" loss function in a pointwise manner.
\end{itemize}
\par
The RoBoSS loss function addresses multiple crucial aspects of supervised learning simultaneously. By encompassing robustness, it ensures the stability of the learning process even in the presence of outliers. The bounded nature of the RoBoSS loss function restricts the impact of extreme values, preventing the loss from growing unbounded. Incorporating sparsity, the RoBoSS loss function promotes the utilization of only the samples that are misclassified or near the decision boundary, resulting in parsimonious models. Moreover, the RoBoSS loss function is designed with a focus on smoothness, facilitating a gradual and consistent optimization process. This smoothness property promotes avoiding abrupt changes during parameter updates, leading to more stable and efficient convergence during training. Next, to highlight the advantages of the proposed RoBoSS loss function, we provide a thorough comparison with existing loss functions:
\begin{enumerate}
\item \textbf{Robustness to outliers:}
Traditional loss functions, including the hinge loss, pinball loss, and LINEX loss, are both unbounded and convex. Although convexity provides certain advantages, the unbounded nature of these functions renders them highly sensitive to outliers. Conversely, the RoBoSS loss function is bounded, greatly improving its robustness to outliers. The bounding parameter $\lambda$ ensures that the loss value does not increase indefinitely for any sample, thereby preventing outliers from disproportionately influencing the model training. This prevents the model from being unduly influenced by extreme values, ensuring a more balanced learning process.
\item \textbf{Flexibility in penalty assignment:} Existing loss functions like the hinge loss and pinball loss, and their variants do not possess a shape parameter and assign a uniform loss value to misclassified samples, regardless of the dataset's characteristics. This uniform penalty approach can be suboptimal when dealing with diverse data distributions, as it does not allow for adjustments based on the specific requirements of different datasets. The RoBoSS loss function offers flexibility in managing different data distributions through its shape parameter $a$. This parameter allows for tuning the strength of the penalty assigned to misclassified samples. By adjusting $a$, one can control the severity of the penalization for misclassifications, providing the ability to adapt the loss function to various data characteristics (see Fig. \ref{fig:RoBoSS loss with fixed lambda}). This flexibility is particularly advantageous when dealing with heterogeneous datasets, as it enables the model to be more responsive to the specific needs of different data distributions.
\item \textbf{Sparsity:} Sparse models are often easier to analyze because they rely on fewer support vectors \cite{wang2024fastC}. The hinge loss, while sparse, lacks boundedness and can lead to suboptimal results on outlier-prone datasets. Pinball loss and LINEX loss, on the other hand, sacrifice both sparsity and boundedness. Wave loss \cite{akhtar2024advancing}, while bounded, lacks the sparsity property. However, the proposed RoBoSS loss strikes a balance by being both sparse and bounded. It enhances sparsity by assigning zero loss to all correctly classified samples. This characteristic ensures that only the most relevant samples contribute to the model's training, leading to simpler models.
\item \textbf{Smoothness:} The non-smooth nature of traditional loss functions such as hinge loss, pinball loss, truncated hinge loss, and truncated pinball loss can lead to challenges in optimization, often requiring specialized algorithms for convergence. The proposed RoBoSS loss function, with its inherent smoothness, allows for the use of gradient-based fast optimization techniques. This smoothness avoids abrupt changes during parameter updates, promoting a more consistent optimization trajectory. The smooth nature of RoBoSS ensures stable updates during training, leading to efficient and effective model convergence.
\end{enumerate}
To succinctly illustrate the advantages of the proposed RoBoSS loss function, we have included a summary table (Table \ref{tab:Comparison of loss function}) comparing the key characteristics of various state-of-the-art loss functions with RoBoSS.
\begin{table}[]
\caption{Illustrates the key attributes of different state-of-the-art loss functions with the proposed RoBoSS loss function, highlighting their robustness, sparsity, boundedness, convexity, and smoothness.}
\label{tab:Comparison of loss function}
\resizebox{\columnwidth}{!}{%
\begin{tabular}{|l|l|l|l|l|l|}
\hline
\textbf{Loss function $\downarrow$\textbackslash{} Characteristic $\rightarrow$} &
  \multicolumn{1}{c|}{\textbf{Robust to outliers}} &
  \multicolumn{1}{c|}{\textbf{Sparse}} &
  \multicolumn{1}{c|}{\textbf{Bounded}} &
  \multicolumn{1}{c|}{\textbf{Convex}} &
  \multicolumn{1}{c|}{\textbf{Smooth}} \\ \hline
\textbf{Hinge loss \cite{cortes1995support}}             &~~~~~~~~~~\textcolor{red}{\ding{55}}  &\textcolor{blue}{\ding{51}}  &\textcolor{red}{\ding{55}}  &\textcolor{blue}{\ding{51}}  &\textcolor{red}{\ding{55}}  \\ \hline
\textbf{Pinball loss \cite{huang2013support}}           &~~~~~~~~~~\textcolor{red}{\ding{55}}  &\textcolor{red}{\ding{55}}  &\textcolor{red}{\ding{55}}  &\textcolor{blue}{\ding{51}}  &\textcolor{red}{\ding{55}}  \\ \hline
\textbf{Truncated hinge loss \cite{wu2007robust}}   &~~~~~~~~~~\textcolor{blue}{\ding{51}}  &\textcolor{blue}{\ding{51}}  &\textcolor{blue}{\ding{51}}  &\textcolor{red}{\ding{55}}  &\textcolor{red}{\ding{55}}  \\ \hline
\textbf{Truncated pinball loss \cite{shen2017support}} &~~~~~~~~~~\textcolor{blue}{\ding{51}}  &\textcolor{red}{\ding{55}}  &\textcolor{blue}{\ding{51}}  &\textcolor{red}{\ding{55}}  &\textcolor{red}{\ding{55}}  \\ \hline
\textbf{LINEX loss \cite{ma2019linex}}             &~~~~~~~~~~\textcolor{red}{\ding{55}}  &\textcolor{red}{\ding{55}}  &\textcolor{red}{\ding{55}}  &\textcolor{blue}{\ding{51}}  & \textcolor{blue}{\ding{51}} \\ \hline
\textbf{QTSELF loss \cite{zhao2022asymmetric}}             &~~~~~~~~~~\textcolor{red}{\ding{55}}  &\textcolor{red}{\ding{55}}  &\textcolor{red}{\ding{55}}  &\textcolor{red}{\ding{55}}  & \textcolor{blue}{\ding{51}} \\ \hline
\textbf{Wave loss \cite{akhtar2024advancing}}             &~~~~~~~~~~\textcolor{blue}{\ding{51}}  &\textcolor{red}{\ding{55}}  &\textcolor{blue}{\ding{51}}  &\textcolor{red}{\ding{55}}  & \textcolor{blue}{\ding{51}} \\ \hline
\textbf{RoBoSS loss (Proposed)}          &~~~~~~~~~~\textcolor{blue}{\ding{51}}  &\textcolor{blue}{\ding{51}}  &\textcolor{blue}{\ding{51}}  &\textcolor{red}{\ding{55}}  &\textcolor{blue}{\ding{51}}  \\ \hline
\end{tabular}%
}
\end{table}
\par
Now, by amalgamating the RoBoSS loss function (\ref{proposedloss}) within the least squares SVM framework, we introduce a novel SVM model termed $\mathcal{L}_{RoBoSS}$-SVM. For simplicity, we adopt the notation $w$ to represent $\left[w^\intercal,b\right]$ and $x_i$ to represent $\left[x_i,1\right]^\intercal$, henceforth. The $\mathcal{L}_{RoBoSS}$-SVM model is delineated as follows: 
\begin{align} \label{proposedSVMmodel}
\underset{ w, \xi}{min} \hspace{0.2cm}~ &\frac{1}{2}\|w\|^2+ \frac{\mathcal{C}}{n} \sum_{k=1}^n \lambda\biggl(1-(a\{\xi_k\}_{+}+1)exp(-a\{\xi_k\}_{+})\biggr), \nonumber  \\
\text {s.t.}\hspace{0.2cm}  & y_k\left(w^\intercal \psi(x_k)\right) = 1-\xi_k, ~\forall~ k=1,2, \ldots,n, 
\end{align}
where $\{\xi_k\}_{+} = \xi_k$ if $\xi_k>0$ and $0$ otherwise, $\mathcal{C}>0$ is the regularization parameter, $a$ and $\lambda$ are the loss parameters, and the function $\psi(\cdot)$ represents the feature mapping corresponding to the kernel function.\\
While kernel functions are typically used to handle non-linear problems through dual problem formulation, the non-convex nature of $\mathcal{L}_{RoBoSS}$-SVM makes this approach formidable.
To empower the non-linear adaptation capability of $\mathcal{L}_{RoBoSS}$-SVM, we utilize the representer theorem \cite{dinuzzo2012representer}.
Using the representer theorem \cite{dinuzzo2012representer}, the corresponding solution can be stated as:
\begin{align} \label{representer theorem}
    w= \sum_{k=1}^n \beta_k \psi(x_k),
\end{align}
where $\beta = \left(\beta_1, \ldots, \beta_n \right)^\intercal$ represents the coefficient vector. By substituting the value of $w$ from equation (\ref{representer theorem}) into equation (\ref{proposedSVMmodel}), we derive
\begin{align} \label{proposed dual}
{\underset{\beta}{min}}~ f(\beta)=&\sum_{k=1}^n \sum_{j=1}^n \frac{1}{2} \beta_k \beta_j \mathcal{K}\left(x_k, x_j\right) \nonumber\\
&+ \frac{\mathcal{C}}{n} \sum_{k=1}^n \lambda\biggl(1-(a\{\xi_k\}_{+}+1)exp(-a\{\xi_k\}_{+})\biggr),
\end{align}
where $\xi_k=y_k\left(\sum_{j=1}^n \beta_{j} \mathcal{K}\left(x_k, x_j\right)\right)-1 $, and $\mathcal{K}\left(x_{k},x_{j}\right)=\psi\left(x_{k}\right) \cdot \psi\left(x_{j}\right)$ is the kernel function.
\section{Theoretical evaluation of the proposed RoBoSS loss function}
Assume that the training data $z$ = $\left\{x_k,y_k\right\}_{k=1}^n$ is drawn independently from a probability measure $\mathcal{P}$. The probability measure $\mathcal{P}$ is defined on $X \times Y$, where $X \subseteq \mathbb{R}^m$ represents the input space and $Y =\{-1,1\}$ is the label space. The primary objective of the classification task is to produce a function $\mathcal{C}: X \rightarrow Y$ that reduces the related risks. The risk related with $\mathcal{C}$ is defined as follows:
$$
{\cal R}({\cal C})=\int_{X}{\cal P} (y\ne{\cal C}(x)\vert x)d{\cal P}_{X},
$$
where ${\cal P} (y\vert x)$ represents the conditional probability distribution of $\mathcal{P}$ given $x$ and $d{\cal P}_{X}$ is the marginal distribution of  $\mathcal{P}$ on $x$. Further, ${\cal P} (y\vert x)$ adheres to a binary distribution, expresses as the likelihoods ${\rm Prob}(y=1 \vert x)$ and ${\rm Prob}(y=-1 \vert x)$. To simplify, we denote ${\rm Prob}(y=1\vert x)$ as $P(x)$ and ${\rm Prob}(y=-1\vert x)$ as $1-P(x)$. Now, for $P(x) \ne 1/2$, the Bayes classifier function ($f_{\mathcal{C}}(x)$) assigns a value of $1$ if $P(x) > 1/2$ and $-1$ if $P(x) < 1/2$.
It can be demonstrated that the Bayes classifier achieves the minimum classification risk \cite{huang2013support}.
Practically, we aim to identify a function $f{:} X \rightarrow \mathbb{R}$ that can generate a binary classifier. In this case, the classification risk becomes $\int_{X\times Y}\mathcal{L}_{mis}(yf(x)) d{\cal P}$, where $\mathcal{L}_{mis}(yf(x))$ is the misclassification loss defined as
\begin{align}      
\mathcal{L}_{mis}(yf(x))=
\begin{cases}
0, & yf(x) > 0, \\
1, & yf(x) \leq 0.
\end{cases}
\end{align}
Therefore, minimizing the misclassification error will result in a function whose sign corresponds to the Bayes classifier \cite{huang2013support}. Now, the expected risk of a classifier \( f: X \rightarrow \mathbb{R} \) for any given loss function \(\mathcal{L}\) can be expressed as:
\begin{align}
    {\cal R}_{\mathcal{L},{\cal P}}(f)=\int_{X\times Y}\mathcal{L}(1-yf(x)) d{\cal P}.
\end{align}
The function \( f_{L,P} \), which achieves the lowest expected risk among all measurable functions, can be described as follows:
\begin{align}{}
    f_{\mathcal{L},\mathcal{P}}(x)=\arg\min_{\mathclap{f(x) \in \mathbb{R}}}~\int_{Y}\mathcal{L}\left(1-yf(x)\right) d\mathcal{P} (y\vert x),~\forall x\in X.
\end{align}
Then, for the RoBoSS loss ($\mathcal{L}_{RoBoSS}(\cdot)$), we can obtain Theorem \ref{theorem classification calibrated}, demonstrating that the RoBoSS loss is classification-calibrated \cite{bartlett2006convexity}. It is a desirable property for a loss function and requires that the minimizer of the function ${\cal R}_{\mathcal{L},{\cal P}}(f)$ shares the sign as of the Bayes classifier. Classification calibration, as introduced by \citet{bartlett2006convexity}, provides a framework for evaluating the statistical efficacy of a loss function. It ensures that the probabilities predicted by the model are closely aligned with the true event probabilities, thereby enhancing the fidelity of the model's predictions.
\begin{theorem}\label{theorem classification calibrated}
The proposed loss $\mathcal{L}_{RoBoSS}(u)$ is classification-calibrated, i.e., $f_{\mathcal{L}_{RoBoSS},\mathcal{P}}$ has the same sign as the Bayes classifier.
\end{theorem}
\begin{proof}
After simple calculation, we obtain that
\begin{align*}
&\int_{Y}\mathcal{L}_{RoBoSS}\left(1-yf(x)\right) d\mathcal{P} (y\vert x)\\
&=\mathcal{L}_{RoBoSS}(1-f(x)) P(x)+\mathcal{L}_{RoBoSS}(1+f(x)) (1-P(x))\\&=
\begin{cases}
g_1(x) P(x), & f(x) \leq -1, \\
(g_1(x) - g_2(x)) P(x) + g_2(x),& -1 < f(x) < 1,\\
g_2(x) (1-P(x)), & f(x) \geq 1,
\end{cases}
\end{align*}
where $g_1(x)= \lambda\biggl(1-(a(1-f(x))+1)exp(-a(1-f(x)))\biggr) $ and $g_2(x)= \lambda\biggl(1-(a(1+f(x))+1)exp(-a(1+f(x)))\biggr)$. \\
\begin{figure*}
\centering
\subcaptionbox{\label{fig:Calibrated1 }} { %
      \includegraphics[width=0.45\textwidth,keepaspectratio]{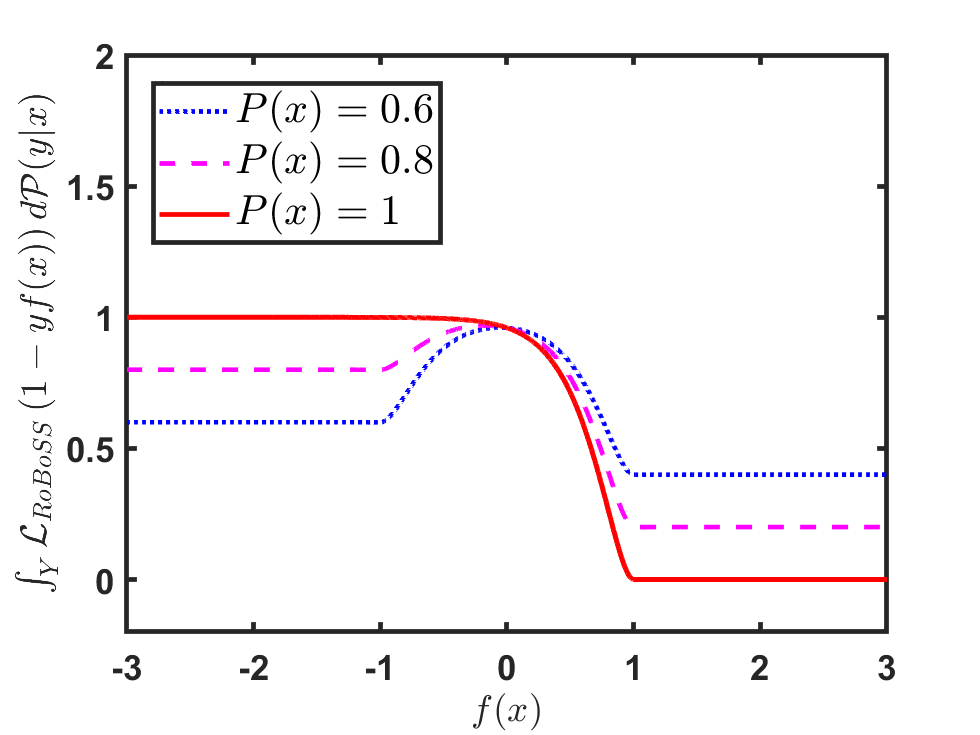}}
      \hfill
      \subcaptionbox{\label{fig:calibrated2 }} { %
      \includegraphics[width=0.45\textwidth,keepaspectratio]{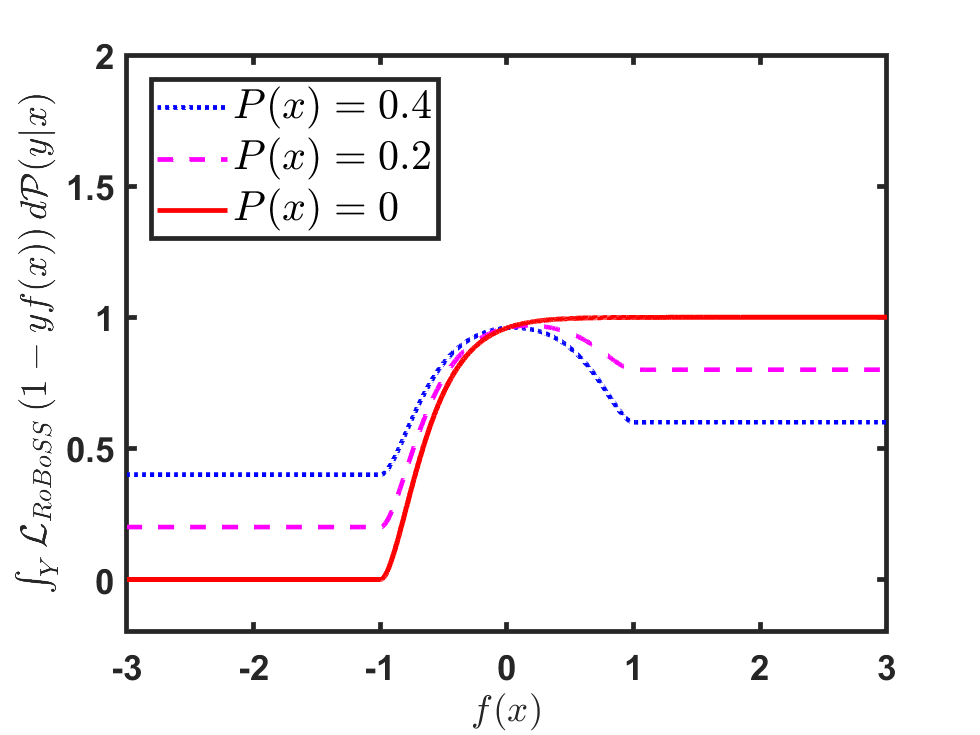}}
      \caption{Demonstrate the graph of $\int_{Y}\mathcal{L}_{RoBoSS}\left(1-yf(x)\right) d\mathcal{P} (y\vert x)$ with respect to $f(x)$ for different $P(x)$ values. (a) For $P(x)$ $>$ $1/2$  and (b) for $P(x)$ $<$ $1/2$.}
    \label{fig:Classification calibrated figures}
 \end{figure*}
 Fig. \ref{fig:Calibrated1 } and \ref{fig:calibrated2 } show the graph of $\int_{Y}\mathcal{L}_{RoBoSS}\left(1-yf(x)\right) d\mathcal{P} (y\vert x)$ over $f(x)$ when $P(x)$ $>$ $1/2$ and $P(x)$ $<$ $1/2$, respectively. It is evident from Fig. \ref{fig:Classification calibrated figures} that, for $P(x)$ $>$ $1/2$, the minimum value of $\int_{Y}\mathcal{L}_{RoBoSS}\left(1-yf(x)\right) d\mathcal{P} (y\vert x)$ is obtained for the positive value of $f(x)$, and for $P(x)$ $<$ $1/2$, it is obtained for the negative value of $f(x)$. Therefore, we can conclude that the function corresponding to the RoBoSS loss, which minimizes the expected risk overall measurable functions, has the same sign as the Bayes classifier.\\
Hence, the proposed loss $\mathcal{L}_{RoBoSS}(u)$ is classification-calibrated. 
\end{proof}
Further, we investigate the generalization ability of $\mathcal{L}_{RoBoSS}$-SVM. First, we define the Rademacher complexity, which measures the complexity of a class of functions.
\begin{definition} Rademacher Complexity \cite{bartlett2002rademacher}\\
    Let $\mathcal{X}$:=$\{x_1, x_2,\ldots,x_p\}$ be drawn independently from $d{\cal P}_{X}$ and $\mathcal{G}$ be a class of functions from $X$ to $\mathbb{R}$. Define the random variable 
\begin{align*}
\hat{R}_p(\mathcal{G}):=\mathbb{E}\left[\sup _{g \in \mathcal{G}}\left|\frac{2}{p} \sum_{k=1}^p \theta_k g\left(x_k\right)\right|\mid\mathcal{X}\right],  
\end{align*}
where $\theta_1, \theta_2,\ldots,\theta_p$ are independent discrete uniform $\{\pm1\}$-valued random variables. Then the Rademacher complexity of $\mathcal{G}$ is $R_p(\mathcal{G})$ = $\mathbb{E} \hat{R}_p(\mathcal{G})$.
\end{definition}
Now, let the expected risk and  empirical risk of RoBoSS loss be denoted by $\mathcal{R}(f_c)$ and $\mathcal{R}_z(f_c)$, respectively, and defined as
\begin{align*}
  \mathcal{R}(f_c)=\int_{X\times Y}\mathcal{L}_{RoBoSS}(1-yf(x)) d{\cal P},
\end{align*}
\begin{align*}
 \mathcal{R}_{z}(f_c)= \frac{1}{n}\sum_{k=1}^n \mathcal{L}_{RoBoSS}(1-yf(x)).  
\end{align*}
Then the generalization ability of $\mathcal{L}_{RoBoSS}$-SVM can be stated as the convergence of $\mathcal{R}_{z}(f_c)$ to $\mathcal{R}(f_c)$ when the sample size $n$ tends to infinity, where $f_c$ is the classifier elicited by (\ref{proposedSVMmodel}). 
\begin{theorem}
Let $f_c$ be the classifier produced by $\mathcal{L}_{RoBoSS}$-SVM. Then for any $0< \varepsilon < 1$, with confidence $1-\varepsilon$, the following inequality holds
\begin{align*}
\mathcal{R}(f_c) -  \mathcal{R}_{z}(f_c) \leq   \frac{4 \lambda}{\sqrt{n \mathcal{C}}} + \sqrt{\frac{8 \ln(1/\varepsilon)}{n}}.
\end{align*}
\end{theorem}

\begin{proof}
For classifier $f_c$, obtained by (\ref{proposedSVMmodel}) with the regularization parameter $\mathcal{C}$, we have
$$
\mathcal{C}\left\|f_c^{\mathcal{L}_{RoBoSS}}\right\|_{\mathcal{K}}^{2} \leq \lambda^{2},
$$
which implies $\left\|f_c^{\mathcal{L}_{RoBoSS}}\right\|_{\mathcal{K}} \leq \lambda / \sqrt{\mathcal{C}}$ \cite{feng2016robust}. Now, using Theorem 8 of \cite{bartlett2002rademacher}, for any $0<\varepsilon<1$, we have
\begin{align} \label{th1}
\mathcal{R}\left(f_c^{\mathcal{L}_{RoBoSS}}\right)-\mathcal{R}_{\mathbf{z}}\left(f_c^{\mathcal{L}_{RoBoSS}}\right) \leq R_{n}(\mathcal{J})+\sqrt\frac{8 \ln (1 / \varepsilon)} {n},
\end{align}
where the  set $\mathcal{J}$ is defined as
$$
\begin{aligned}
\mathcal{J}:= & \biggl\{ j \mid j(x, y):=\phi(1-y f(x))-\phi(0), f \in \mathcal{J}_{\mathcal{K}},\biggr. \\
& \left.\|f\|_{\mathcal{K}} \leq \lambda / \sqrt{\mathcal{C}},(x, y) \in X \times Y\right\}.
\end{aligned}
$$
Again, Theorem 12 of \cite{bartlett2002rademacher} yields that
$$
\begin{aligned}
& R_{n}(\mathcal{J}) \leq 2 R_{n}\left(\mathcal{G}_{\mathcal{C}}\right) \text { with } \\
& \mathcal{G}_{\mathcal{C}}=\left\{f \mid f \in \mathcal{J}_{\mathcal{K}},\|f\|_{\mathcal{K}} \leq \lambda \sqrt{\log \left(1+\lambda^{-2}\right) / \mathcal{C}}\right\} .
\end{aligned}
$$
Also from \cite{mendelson2003few}, we have
\begin{align} \label{th2}
R_{n}\left(\mathcal{G}_{\mathcal{C}}\right) \leq \frac{2 \lambda}{\sqrt{n \mathcal{C}}}.
\end{align}
Hence, from (\ref{th1}) and (\ref{th2}), for any $0<\varepsilon<1$, we have
$$
\mathcal{R}(f_c^{\mathcal{L}_{RoBoSS}}) -  \mathcal{R}_{z}(f_c^{\mathcal{L}_{RoBoSS}}) \leq   \frac{4 \lambda}{\sqrt{n \mathcal{C}}} + \sqrt{\frac{8 \ln(1/\varepsilon)}{n}}.
$$    
\end{proof}
\section{Optimization of $\mathcal{L}_{RoBoSS}$-SVM}
To solve the optimization problem (\ref{proposed dual}), we adopt the framework based on the Nestrov accelerated gradient (NAG) algorithm. It is an extension of the stochastic gradient descent (SGD) method that incorporates momentum to accelerate convergence. In SGD, a small batch of samples (mini-batch) is used for each iteration during the training of a model. This approach offers several advantages, including reduced computational requirements and improved speed, particularly when dealing with large-scale problems. However, SGD has some drawbacks, such as getting stuck in local optima during its process of convergence due to the randomness of the mini-batch. To improve SGD, many researchers introduced accelerated variance in SGD \cite{johnson2013accelerating, yuan2020federated}. The momentum method \cite{qian1999momentum} is a practical approach that helps SGD to accelerate in the relevant direction and dampen the oscillation. It does this by combining the update vector of the previous time step with the current update vector.\\
The NAG algorithm is an extension of the momentum method that further improves convergence by incorporating a ``look-ahead" mechanism \cite{nesterov1983method}. It gives an approximation of the future position of the parameters and then calculates the gradient with respect to the approximate future position of the model parameters. One challenge for NAG is to choose an appropriate learning rate during the training. If the learning rate is set to a very low value,  the algorithm's convergence speed becomes sluggish. On the contrary, using a high learning rate is likely to cause the algorithm to overshoot the optimal
point or even fail to converge. An intuitive approach is to begin with a slightly higher learning rate and then gradually reduce it during the learning process according to a predefined schedule.  Taking inspiration from the simulated annealing approach \cite{kirkpatrick1983optimization}, we employ
 the exponential decay method for adjusting the learning rate as $\alpha_{new} = \alpha_{old} \exp(-\eta t)$, where $\eta$ is a hyperparameter that determines the extent of the learning rate's decay at each iteration, while $t$ represents the current iteration number.\\
Now, we solve (\ref{proposed dual}) by employing the NAG-based framework. The method employed to solve (\ref{proposed dual}) is thoroughly described in Algorithm \ref{NAG algorithm}. After obtaining the optimal $\beta$,
the subsequent  decision function is used to classify a test sample $\hat{x}$:
\begin{align} \label{non-linear decision function}
\hat{y}= {\operatorname{\text{sign}}}\left(\sum_{j=1}^s \beta_{j} \mathcal{K}\left(x_j, \hat{x}\right)\right).
\end{align}
\par
To elucidate the integration of RoBoSS loss into the SVM framework, Fig. \ref{fig:flowchart} presents a flowchart of the proposed $\mathcal{L}_{RoBoSS}$-SVM model that encapsulates our methodology. This visual representation serves as both a guide to our methodology and a demonstration of the strategic integration of its components.
\begin{figure}
    \centering
    \includegraphics[width=1.025\linewidth]{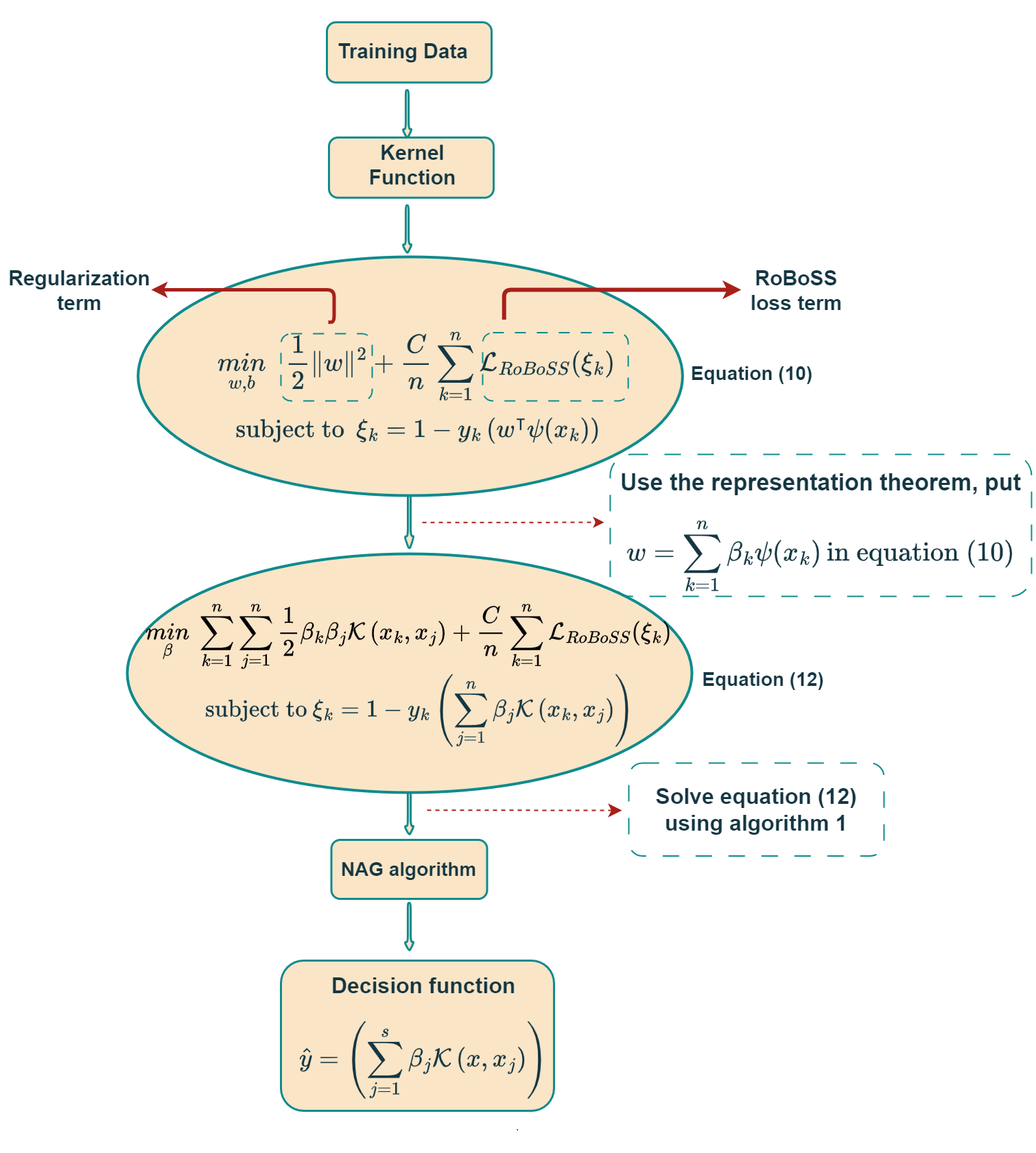}
    \caption{Flowchart of the proposed $\mathcal{L}_{RoBoSS}$-SVM model. It depicts the essential stages of the proposed $\mathcal{L}_{RoBoSS}$-SVM, demonstrating the process from raw input data to the final decision function. This flowchart shows the incorporation of the RoBoSS loss function and illustrates the use of the representer theorem and the NAG optimization algorithm.}
    \label{fig:flowchart}
\end{figure}

\begin{algorithm} 
 \caption{NAG-based algorithm to solve $\mathcal{L}_{RoBoSS}$-SVM}
 \label{NAG algorithm}
 \textbf{Input:}\\
 The dataset:  $\left\{x_k,y_k\right\}_{k=1}^n$, $y_k \in\{-1,1\}$;\\
 The parameters: Regularization parameter $\mathcal{C}$, RoBoSS loss parameters $\lambda$ and $a$, mini-batch size $s$, learning rate decay factor $\eta$, momentum parameter $r$, maximum iteration number $N$;\\
 Initialize: model parameter $\beta_0$, velocity $\upsilon_0$, learning rate $\alpha$;
 \textbf{Output:}\\ 
 The classifiers parameters: $\beta$;\\
1: Select $s$ samples $\left\{x_k,y_k\right\}_{k=1}^s$ uniformly at random.\\
2: Computing $\xi_k$ :
 \begin{align}
 \xi_k= 1 -y_k\left(\sum_{j=1}^s \beta_{j} \mathcal{K}\left(x_k, x_j\right)\right),~ k=1, \cdots, s;
 \end{align}
3: Temporary update: $\widetilde{\beta_t} = \beta_t +r v_t$;\\
4: Compute gradient:
 \begin{align}
 \nabla f(\widetilde{\beta_t})=\mathcal{K} \widetilde{\beta_t} - \frac{\mathcal{C}}{s} \lambda\sum_{j=1}^s   a^2 \xi_j \exp\left(-a \xi_j\right) y_j \mathcal{K}_j,
 \end{align}
where $\mathcal{K}$ denotes the kernel matrix and $\mathcal{K}_{j}$ denotes the $j^{th}$ row of $\mathcal{K}$.\\
5: Update velocity: $v_t=r v_{t-1}-\alpha_{t-1} \nabla f(\widetilde{\beta_t})$;\\
6: Update model parameter: $\widetilde{\beta_{t+1}}= \widetilde{\beta_t} + v_t$;\\
7: Update learning rate: $\alpha_{t+1}= \alpha_t \exp(-\eta t)$;\\
8: Update current iteration number: $t=t+1$.\\
\textbf{Until:}\\
 $t=N$.\\
\textbf{Return:} $\beta_t$.
 \end{algorithm}
\subsection{Computational analysis}
In this subsection, we provide an analysis of the computational complexity of the NAG algorithm (Algorithm \ref{NAG algorithm}) utilized to solve the optimization problem of the proposed $\mathcal{L}_{RoBoSS}$-SVM. 
Consider \( l \) and \( m \) represent the sample and feature counts, respectively, in the training data, while \( N \) indicates the count of iterations.
The computational complexity associated with updating the variable $v$ can be expressed as $\mathcal{O}\left((N+m) l^2\right)$. Similarly, the complexity for updating the parameter $\beta$ is $\mathcal{O}(N l)$, lastly the complexity of updating the parameter $\alpha$ is $O(N)$ \cite{tang2024advancing}. Combining these complexities, the total computational complexity for the NAG algorithm in solving the $\mathcal{L}_{RoBoSS}$-SVM can be summarized as $\mathcal{O}\left((N+m) l^2\right)$. It is important to note that the computational complexity of the traditional SVM model is \(\mathcal{O}(l^3)\). This highlights that the proposed $\mathcal{L}_{RoBoSS}$-SVM, with its computational complexity of \(\mathcal{O}\left((N+m) l^2\right)\), offers a significant improvement in efficiency over the traditional SVM model, particularly as the number of samples \(l\) increases. This enhanced efficiency underscores the practicality and scalability of the $\mathcal{L}_{RoBoSS}$-SVM for large-scale datasets.
\section{Experimental Results}
This section discusses the results produced by the numerical experiment conducted in this study. We compare the proposed $\mathcal{L}_{RoBoSS}$-SVM against five baseline loss function-based SVMs, namely $\mathcal{L}_{hinge}$-SVM \cite{cortes1995support}, $\mathcal{L}_{pin}$-SVM \cite{huang2013support}, $\mathcal{L}_{LINEX}$-SVM \cite{ma2019linex}, $\mathcal{L}_{qtse}$-SVM \cite{zhao2022asymmetric}, and $\mathcal{L}_{wave}$-SVM \cite{akhtar2024advancing}. The detailed experimental setup is meticulously detailed in Section S.I of the supplement material file.
\subsection{Evaluation on KEEL and UCI datasets}
Here, we present the experimental results on $88$ real-world datasets downloaded from the KEEL \cite{derrac2015keel} and UCI \cite{dua2017uci} repositories. Based on the sample size, we split the datasets into two categories: (D1) datasets with samples under or equal to $5000$, and (D2) datasets with samples over $5000$. There are $79$ and $9$ datasets in the D1 and D2 categories, respectively. 
\par
Table \ref{tab:small-table} presents the average accuracy, training time, and rank of the models on $79$ D1 category datasets.
The proposed $\mathcal{L}_{RoBoSS}$-SVM stands out with an average accuracy of 86.35\% and a standard deviation of 5.06, the highest and most consistent performance among all the models. This suggests that $\mathcal{L}_{RoBoSS}$-SVM not only excels in overall accuracy but also maintains stable performance across diverse datasets, highlighting its reliability. In comparison, the baseline models $\mathcal{L}_{hinge}$-SVM, $\mathcal{L}_{pin}$-SVM, $\mathcal{L}_{LINEX}$-SVM, $\mathcal{L}_{qtse}$-SVM, and $\mathcal{L}_{wave}$-SVM show lower average accuracies of 83.16\%, 84.26\%, 82.53\%, 82.18\%, and 83.21\%, respectively, with higher standard deviations, indicating more variability in their performance. In terms of training time, $\mathcal{L}_{RoBoSS}$-SVM exhibits the best average training time of 0.0012 seconds. While the baseline models have longer average training times, with $\mathcal{L}_{hinge}$-SVM at 0.1304, $\mathcal{L}_{pin}$-SVM at 0.1909, $\mathcal{L}_{LINEX}$-SVM at 0.0031, $\mathcal{L}_{qtse}$-SVM at 0.0019, and $\mathcal{L}_{wave}$-SVM at 0.0037 seconds. The average rank further underscores the superiority of $\mathcal{L}_{RoBoSS}$-SVM, with the lowest average rank of 2.16, indicating its consistent high performance across various datasets. In contrast, the baseline models have higher average ranks: $\mathcal{L}_{hinge}$-SVM at 3.35, $\mathcal{L}_{pin}$-SVM at 2.96, $\mathcal{L}_{LINEX}$-SVM at 3.96, $\mathcal{L}_{qtse}$-SVM at 4.45, and $\mathcal{L}_{wave}$-SVM at 4.12. These results collectively highlight the advantages of the proposed $\mathcal{L}_{RoBoSS}$-SVM model, showcasing its ability to deliver higher accuracy, faster training times, and more consistent performance compared to traditional SVM models. The bounded and sparse characteristics of the RoBoSS loss function help in mitigating the influence of outliers and ensuring that the model prioritizes the most critical samples. This leads to better generalization and efficiency, making $\mathcal{L}_{RoBoSS}$-SVM a robust and effective choice for various supervised learning tasks.
The detailed results for each of the 79 datasets and the corresponding best parameters are available in Tables S.V and S.VIII of the supplement material file, respectively.
The results for each of the $9$ D2 category datasets are presented in Table \ref{tab: large dataset table}. These results demonstrate the superior performance of the proposed $\mathcal{L}_{RoBoSS}$-SVM across several datasets.  For instance, on the Musk2 dataset, $\mathcal{L}_{RoBoSS}$-SVM  achieves an accuracy of 84.59\% with a standard deviation of 34.46, which is identical to $\mathcal{L}_{pin}$-SVM, $\mathcal{L}_{LINEX}$-SVM, and $\mathcal{L}_{wave}$-SVM but with a significantly faster training time. On the Ringnorm dataset, $\mathcal{L}_{RoBoSS}$-SVM  achieves the highest accuracy of 52.22\% with a standard deviation of 0.9, outperforming all other models. Similarly, on the Twonorm dataset, $\mathcal{L}_{RoBoSS}$-SVM attains the highest accuracy of 52.24\% with a standard deviation of 2.01, again outperforming the competing models. The average accuracy of the $\mathcal{L}_{RoBoSS}$-SVM model across all nine datasets is 74.35\%, which is higher compared to the baseline models: $\mathcal{L}_{hinge}$-SVM at 66.16\%, $\mathcal{L}_{pin}$-SVM at 68.15\%, $\mathcal{L}_{LINEX}$-SVM at 69.73\%, $\mathcal{L}_{qtse}$-SVM at 73.2\%, and $\mathcal{L}_{wave}$-SVM at 73.3\%. The overall results convincingly demonstrate the superiority of the $\mathcal{L}_{RoBoSS}$-SVM model over traditional SVM models. These results validate the potential of the RoBoSS loss function in enhancing the robustness and efficiency of SVM models, making $\mathcal{L}_{RoBoSS}$-SVM a highly effective choice for complex and large-scale classification tasks.

\begin{tiny}
\begin{table*}[h]
\centering
\caption{The average results of $\mathcal{L}_{RoBoSS}$-SVM along with the compared models on $79$ D1 category KEEL and UCI datasets.}
 \resizebox{15cm}{!}{
\label{tab:small-table}
\begin{tabular}{lcccccc}
\hline
 Model & $\mathcal{L}_{hinge}$-SVM \cite{cortes1995support}             & $\mathcal{L}_{pin}$-SVM \cite{huang2013support}           & $\mathcal{L}_{LINEX}$-SVM \cite{ma2019linex}         & $\mathcal{L}_{qtse}$-SVM \cite{zhao2022asymmetric} & $\mathcal{L}_{wave}$-SVM \cite{akhtar2024advancing} & $\mathcal{L}_{RoBoSS}$-SVM$^{\dagger}$ \\ \hline
Avg. Acc. $\pm$ Avg. Std. & 83.16$\pm$7.04 & \underline{84.26$\pm$6.44} & 82.53$\pm$7.39 & 82.18$\pm$7.91 & 83.21$\pm$6.6& \textbf{86.35$\pm$5.06} \\ \hline
Avg. time     & 0.1304    & 0.1909     & 0.0031     & \underline{0.0019} & 0.0037    & \textbf{0.0012}     \\ \hline
Avg. rank     & 3.35     & \underline{2.96}     & 3.96     & 4.45 & 4.12    & \textbf{2.16}     \\ \hline
\multicolumn{7}{l}{Acc., Avg., and Std. stand for accuracy, average, and standard deviation, respectively. $^{\dagger}$ signifies the proposed model.}\\
 \multicolumn{7}{l}{Boldface highlights the top-performing model, while underlining indicates the second-best model. }
\end{tabular}%
}
\end{table*}
\end{tiny}

\begin{table*}[h]
\centering
\caption{The classification accuracies and training times of the $\mathcal{L}_{RoBoSS}$-SVM along with the compared models on $9$ D2 category KEEL and UCI datasets.}
\label{tab: large dataset table}
\resizebox{17cm}{!}{%
\begin{tabular}{llccccc}
\hline
Model & $\mathcal{L}_{hinge}$-SVM \cite{cortes1995support}             & $\mathcal{L}_{pin}$-SVM \cite{huang2013support}           & $\mathcal{L}_{LINEX}$-SVM \cite{ma2019linex}         & $\mathcal{L}_{qtse}$-SVM \cite{zhao2022asymmetric} & $\mathcal{L}_{wave}$-SVM \cite{akhtar2024advancing} & $\mathcal{L}_{RoBoSS}$-SVM$^{\dagger}$          \\ \hline
\begin{tabular}[c]{@{}l@{}}Dataset \\ (samples, features)\end{tabular} &
  Acc. $\pm$ Std., time &
  \multicolumn{1}{l}{Acc. $\pm$ Std., time} &
  \multicolumn{1}{l}{Acc. $\pm$ Std., time} &
  \multicolumn{1}{l}{Acc. $\pm$ Std., time} &
  \multicolumn{1}{l}{Acc. $\pm$ Std., time} &
  \multicolumn{1}{l}{Acc. $\pm$ Std., time} \\ \hline
\begin{tabular}[c]{@{}l@{}}Musk2\\ (6598, 166)\end{tabular} &
  \multicolumn{1}{c}{80$\pm$44.72, 18.7863} &
  \textbf{84.59$\pm$34.46}, 22.9542 &
  \textbf{84.59$\pm$34.46}, 0.0048 &
  \underline{81.02$\pm$17.85}, 0.0023 &
  \textbf{84.59$\pm$34.46}, 0.0035 &
  \textbf{84.59$\pm$34.46}, 0.0029 \\
\begin{tabular}[c]{@{}l@{}}Ringnorm\\ (7400, 20)\end{tabular} &
  \multicolumn{1}{c}{50.5$\pm$1.29, 8.0734} &
  50.95$\pm$0.88, 14.7853 &
  51.15$\pm$0.62, 0.0033 &
  51.03$\pm$1.03, 0.0019 &
  \underline{51.19$\pm$0.54}, 0.0027 &
  \textbf{52.22$\pm$0.9}, 0.0018 \\
\begin{tabular}[c]{@{}l@{}}Twonorm\\ (7400, 20)\end{tabular} &
  \multicolumn{1}{c}{50.61$\pm$0.82, 5.0918} &
  50.8$\pm$0.52, 28.8532 &
  50.78$\pm$0.9, 0.0031 &
  \underline{50.92$\pm$1.35}, 0.0019 &
  50.88$\pm$2.19, 0.0027 &
  \textbf{52.24$\pm$2.01}, 0.0024 \\
\begin{tabular}[c]{@{}l@{}}EEG Eye State\\ (14980, 14)\end{tabular} &
  \multicolumn{1}{c}{55.12$\pm$25.92, 127.686} &
  61.78$\pm$22.46, 192.8241 &
  68.93$\pm$16.06, 0.0033 &
  69.71$\pm$15.36, 0.002 &
  \underline{70.35$\pm$15.6}, 0.0039 &
  \textbf{71.2$\pm$13.68}, 0.0017 \\
\begin{tabular}[c]{@{}l@{}}Magic\\ (19020,10)\end{tabular} &
  \multicolumn{1}{c}{82.84$\pm$9.8, 443.3522} &
  82.88$\pm$9.72, 217.4496 &
  65.3$\pm$25.25, 0.0043 &
  \textbf{95.16$\pm$10.82}, 0.0021 &
  90.38$\pm$17.47, 0.0032 &
  \underline{95.16$\pm$33.91}, 0.0023 \\
\begin{tabular}[c]{@{}l@{}}Credit Default\\ (30000, 23)\end{tabular} &
 \textbf{77.89$\pm$1.56}, 247.5376 &
  \underline{77.88$\pm$1.56}, 1415.7059 &
  \underline{77.88$\pm$1.56}, 0.0062 &
  \underline{77.88$\pm$1.56}, 0.0034 &
  \underline{77.88$\pm$1.56}, 0.0229 &
  \underline{77.88$\pm$1.56}, 0.0069 \\
\begin{tabular}[c]{@{}l@{}}Adult\\ (48842, 14)\end{tabular} &
  ~~~~~~~~~~~~~~* &
  * &
  \underline{76.41$\pm$1.8}, 0.0106 &
  76.07$\pm$0.25, 0.0042 &
  76.19$\pm$2.33, 0.0085 &
  \textbf{77.94$\pm$1.51}, 0.0051 \\
\begin{tabular}[c]{@{}l@{}}Connect4\\ (67557, 42)\end{tabular} & ~~~~~~~~~~~~~~* & * &
  \underline{75.38$\pm$3.78}, 0.0118 &
  \underline{75.38$\pm$3.78}, 0.0057 &
  \underline{75.38$\pm$3.78}, 0.0103 &
  \textbf{75.4$\pm$3.75}, 0.0082 \\
\begin{tabular}[c]{@{}l@{}}Miniboone\\ (130064, 50)\end{tabular} &
  ~~~~~~~~~~~~~~* &
  * &
  77.17$\pm$18.82, 0.0156 &
  81.67$\pm$17.11, 0.0144 &
  \textbf{82.85$\pm$6.98}, 0.0156 &
  \underline{82.5$\pm$7.94}, 0.0118 \\ \hline
Avg Acc. $\pm$ Avg. Std. &
  66.16$\pm$14.02 &
  \multicolumn{1}{l}{68.15$\pm$11.6} &
  69.73$\pm$11.47 &
  73.2$\pm$7.68 &
  \underline{73.3$\pm$9.43} &
  \textbf{74.35$\pm$11.08} \\ \hline
  \multicolumn{7}{l}{Acc., Avg., and Std. stand for accuracy, average, and standard deviation, respectively. $^{\dagger}$ signifies the proposed model.}\\
  \multicolumn{7}{l}{* denote that MATLAB encounters an ``out of memory" error.}\\
\multicolumn{7}{l}{Boldface highlights the top-performing model, while underlining indicates the second-best model.}
\end{tabular}%
}
\end{table*}
\subsection{Evaluation on datasets with introduced outliers and label noise}
To rigorously assess the robustness and generalization capabilities of the proposed $\mathcal{L}_{RoBoSS}$-SVM model, it is essential to evaluate its performance under challenging conditions. Real-world data often contains outliers and label noise, which can significantly impact the accuracy and reliability of machine learning models. Therefore, conducting an evaluation on datasets with artificially introduced outliers and label noise provides a comprehensive understanding of the model's resilience to such anomalies. In this study, we selected five diverse datasets, namely cylinder\_bands, ionosphere, spectf, titanic, stalog\_australian\_credit. The methodology for introducing outliers and label noise into training dataset is discussed in Section S.I of the supplement material file.
\par
Table \ref{tab:Outlier-table} displays the classification accuracy of the $\mathcal{L}_{RoBoSS}$-SVM model alongside the compared models $\mathcal{L}_{hinge}$-SVM, $\mathcal{L}_{pin}$-SVM, $\mathcal{L}_{LINEX}$-SVM, $\mathcal{L}_{qtse}$-SVM, and $\mathcal{L}_{wave}$-SVM across datasets with 5\%, 10\%, 20\%, and 30\% outliers. The proposed $\mathcal{L}_{RoBoSS}$-SVM model consistently surpasses the compared models. Specifically, in four out of five datasets, the $\mathcal{L}_{RoBoSS}$-SVM model achieves the top position, while in one dataset, it secures the second-best position compared to the baseline models. This superior performance can be attributed to the bounded nature of the RoBoSS loss function, which mitigates the influence of extreme values, thereby maintaining high accuracy even when datasets are significantly contaminated with outliers. In terms of overall accuracy, the $\mathcal{L}_{RoBoSS}$-SVM model shows a total average accuracy of 76.67\%, outperforming the baseline models $\mathcal{L}_{hinge}$-SVM (71.35\%), $\mathcal{L}_{pin}$-SVM (72.75\%), $\mathcal{L}_{LINEX}$-SVM (73.27\%), $\mathcal{L}_{qtse}$-SVM (70.3\%), and $\mathcal{L}_{wave}$-SVM (75.07\%). This consistency across different levels of outliers underscores the $\mathcal{L}_{RoBoSS}$-SVM robustness and stability in handling outlier-prone data.
In the context of evaluating the robustness of the RoBoSS-SVM in the presence of label noise, we have observed notable results, as detailed in Table \ref{tab:Noisy-label-table}. Specifically, $\mathcal{L}_{wave}$-SVM achieves the best accuracy on two datasets and the second-best accuracy on three datasets. Similarly, $\mathcal{L}_{pin}$-SVM attains the best accuracy on three datasets and the second-highest result on two datasets. In terms of total average accuracy across all five datasets and noise ratios, $\mathcal{L}_{wave}$-SVM outperforms the baseline models with an average accuracy of 72.47\%. Meanwhile, $\mathcal{L}_{RoBoSS}$-SVM achieves the second-best total average accuracy of 71.67\%. These findings emphasize that $\mathcal{L}_{wave}$-SVM and $\mathcal{L}_{pin}$-SVM are particularly effective in environments with label noise, while the proposed $\mathcal{L}_{RoBoSS}$-SVM also demonstrates competitive performance. These outcomes align with existing literature, which suggests that loss functions incorporating penalties for correctly classified samples tend to be more effective in managing label noise \cite{huang2013support, akhtar2024advancing}. This insight underscores the potential for further enhancing the efficiency of the $\mathcal{L}_{RoBoSS}$-SVM by exploring modifications to the RoBoSS loss function. Future research could focus on redesigning the RoBoSS loss function to also penalize correctly classified samples to a certain extent while maintaining a balance with its sparsity property. This approach could improve its performance in scenarios with prevalent label noise.
\begin{table*}[]
 \centering
\caption{The classification accuracy of the proposed $\mathcal{L}_{RoBoSS}$-SVM along with the compared models on datasets with varying levels of outliers.}
\label{tab:Outlier-table}
\resizebox{17cm}{!}{%
\begin{tabular}{lccccccc}
\hline
Dataset & Outliers & $\mathcal{L}_{hinge}$-SVM \cite{cortes1995support}             & $\mathcal{L}_{pin}$-SVM \cite{huang2013support}           & $\mathcal{L}_{LINEX}$-SVM \cite{ma2019linex}         & $\mathcal{L}_{qtse}$-SVM \cite{zhao2022asymmetric} & $\mathcal{L}_{wave}$-SVM \cite{akhtar2024advancing} & $\mathcal{L}_{RoBoSS}$-SVM$^{\dagger}$ \\ \hline
cylinder\_bands & 5\% & 64.79 & 64.79 & 63.66 & 60.87 & 66.75 & 68.53 \\
 & 10\% & 64.79 & 64.79 & 65.41 & 61.26 & 66.75 & 69.33 \\
 & 20\% & 67.93 & 64.79 & 60.87 & 61.26 & 66.75 & 68.53 \\
 & 30\% & 63.22 & 66.75 & 64.82 & 61.26 & 66.75 & 67.95 \\ \hline
Avg. & \multicolumn{1}{l}{} & 65.19 & 65.28 & 63.69 & 61.17 & \underline{66.75} & \textbf{68.59} \\ \hline
ionosphere & 5\% & 65.8 & 68.36 & 78.38 & 69.07 & 84.64 & 88.06 \\
 & 10\% & 65.8 & 72.35 & 76.66 & 64.43 & 85.21 & 87.48 \\
 & 20\% & 70.66 & 75.2 & 77.81 & 64.14 & 83.22 & 88.07 \\
 & 30\% & 67.48 & 76.93 & 79.24 & 64.43 & 82.91 & 87.76 \\ \hline
Avg. & \multicolumn{1}{l}{} & 67.43 & 73.21 & 78.02 & 65.52 & \underline{83.99} & \textbf{87.84} \\ \hline
spectf & 5\% & 79.34 & 79.34 & 79.34 & 79.34 & 79.34 & 79.34 \\
 & 10\% & 79.34 & 79.34 & 79.34 & 79.34 & 79.34 & 79.34 \\
 & 20\% & 79.34 & 79.34 & 79.34 & 79.34 & 79.34 & 79.72 \\
 & 30\% & 79.34 & 79.34 & 79.34 & 79.34 & 79.34 & 79.34 \\ \hline
Avg. & \multicolumn{1}{l}{} & \underline{79.34} & \underline{79.34} & \underline{79.34} & \underline{79.34} & \underline{79.34} & \textbf{79.44} \\ \hline
titanic & 5\% & 76.33 & 77.69 & 77.33 & 77.64 & 77.92 & 79.05 \\
 & 10\% & 76.33 & 78.28 & 77.87 & 77.55 & 77.33 & 79.05 \\
 & 20\% & 77.33 & 78.28 & 77.1 & 76.01 & 77.33 & 79.05 \\
 & 30\% & 77.33 & 77.33 & 77.33 & 76.83 & 76.83 & 79.05 \\ \hline
Avg. & \multicolumn{1}{l}{} & 76.83 & \underline{77.89} & 77.41 & 77.01 & 77.35 & \textbf{79.05} \\ \hline
statlog\_australian\_credit & 5\% & 67.97 & 67.97 & 67.83 & 68.55 & 67.97 & 68.12 \\
 & 10\% & 67.97 & 67.83 & 67.83 & 68.41 & 67.83 & 68.26 \\
 & 20\% & 67.83 & 68.12 & 67.97 & 68.55 & 67.97 & 68.26 \\
 & 30\% & 68.12 & 68.12 & 67.83 & 68.41 & 67.97 & 68.84 \\ \hline
Avg. & \multicolumn{1}{l}{} & 67.97 & 68.01 & 67.86 & \textbf{68.48} & 67.93 & \underline{68.37} \\ \hline
Total Avg. & \multicolumn{1}{l}{} & 71.35 & 72.75 & 73.27 & 70.3 & \underline{75.07} & \textbf{76.67} \\ \hline
\multicolumn{8}{l}{Here, Avg. denotes the average, and $^{\dagger}$ signifies the proposed model.}\\
\multicolumn{8}{l}{Boldface highlights the top-performing model, while underlining indicates the second-best model.}
\end{tabular}}
\end{table*}


\begin{table*}[p]
\centering
\caption{The classification accuracy of the proposed $\mathcal{L}_{RoBoSS}$-SVM along with the compared models on datasets with varying levels of label noise.}
\label{tab:Noisy-label-table}
\resizebox{17cm}{!}{%
\begin{tabular}{lccccccc}
\hline
Dataset & Noise & $\mathcal{L}_{hinge}$-SVM \cite{cortes1995support}             & $\mathcal{L}_{pin}$-SVM \cite{huang2013support}           & $\mathcal{L}_{LINEX}$-SVM \cite{ma2019linex}         & $\mathcal{L}_{qtse}$-SVM \cite{zhao2022asymmetric} & $\mathcal{L}_{wave}$-SVM \cite{akhtar2024advancing} & $\mathcal{L}_{RoBoSS}$-SVM$^{\dagger}$ \\ \hline
cylinder\_bands & 5\% & 61.07 & 61.26 & 60.87 & 60.87 & 62.85 & 61.85 \\
 & 10\% & 61.07 & 66.75 & 62.47 & 60.87 & 63.46 & 64.41 \\
 & 20\% & 60.87 & 61.26 & 60.87 & 62.24 & 62.87 & 60.87 \\
 & 30\% & 61.26 & 64.79 & 60.87 & 60.87 & 60.87 & 61.08 \\ \hline
Avg. & \multicolumn{1}{l}{} & 61.07 & \textbf{63.52} & 61.27 & 61.22 & \underline{62.51} & 62.05 \\ \hline
ionosphere & 5\% & 66.14 & 67.79 & 65.84 & 68.45 & 81.35 & 75.64 \\
 & 10\% & 69.28 & 69.75 & 71.79 & 64.14 & 75.51 & 73.51 \\
 & 20\% & 65.48 & 72.05 & 71.53 & 66.95 & 74.66 & 72.66 \\
 & 30\% & 70.37 & 70.37 & 68.37 & 66.16 & 74.39 & 72.39 \\ \hline
Avg. & \multicolumn{1}{l}{} & 67.82 & \underline{69.99} & 69.38 & 66.42 & \textbf{76.48} & 73.55 \\ \hline
spectf & 5\% & 79.34 & 79.34 & 79.34 & 79.34 & 79.34 & 79.34 \\
 & 10\% & 79.34 & 79.34 & 79.34 & 79.34 & 79.34 & 79.34 \\
 & 20\% & 79.34 & 79.34 & 79.34 & 79.34 & 79.71 & 79.34 \\
 & 30\% & 79.34 & 79.34 & 79.34 & 79.34 & 79.34 & 79.34 \\ \hline
Avg. & \multicolumn{1}{l}{} & \underline{79.34} & \underline{79.34} & \underline{79.34} & \underline{79.34} & \textbf{79.44} & \underline{79.34} \\ \hline
titanic & 5\% & 77.1 & 77.1 & 77.1 & 72.07 & 78.43 & 77.33 \\
 & 10\% & 76.4 & 77.1 & 77.33 & 71.84 & 77.33 & 77.92 \\
 & 20\% & 76.1 & 77.1 & 73.69 & 73.69 & 77.1 & 73.69 \\
 & 30\% & 75.33 & 76.33 & 72.78 & 73.92 & 72.42 & 73.01 \\ \hline
Avg. & \multicolumn{1}{l}{} & 76.23 & \textbf{76.91} & 75.22 & 72.88 & \underline{76.32} & 75.49 \\ \hline
statlog\_australian\_credit & 5\% & 67.97 & 68.97 & 67.83 & 67.97 & 68.5 & 67.83 \\
 & 10\% & 68.12 & 69.12 & 67.83 & 67.83 & 67.83 & 67.83 \\
 & 20\% & 68.12 & 68.12 & 67.83 & 67.83 & 68.23 & 68.12 \\
 & 30\% & 67.83 & 67.83 & 67.97 & 67.83 & 67.83 & 67.97 \\ \hline
Avg. & \multicolumn{1}{l}{} & 68.01 & \textbf{68.51} & 67.86 & 67.86 & \underline{68.1} & 67.93 \\ \hline
Total Avg. & \multicolumn{1}{l}{} & 70.49 & 71.65 & 70.62 & 69.54 & \textbf{72.47} & \underline{71.67} \\ \hline
\multicolumn{8}{l}{Here, Avg. denotes the average, and $^{\dagger}$ signifies the proposed model.}\\
\multicolumn{8}{l}{Boldface highlights the top-performing model, while underlining indicates the second-best model.}
\end{tabular}}
\end{table*}

\begin{table*}[h]
\caption{The average results of $\mathcal{L}_{RoBoSS}$-SVM along with the compared models on the EEG dataset.}
\centering
\label{tab:EEG-table}
\resizebox{17cm}{!}{%
\begin{tabular}{lcccccc}
\hline
Model & $\mathcal{L}_{hinge}$-SVM \cite{cortes1995support}             & $\mathcal{L}_{pin}$-SVM \cite{huang2013support}           & $\mathcal{L}_{LINEX}$-SVM \cite{ma2019linex}         & $\mathcal{L}_{qtse}$-SVM \cite{zhao2022asymmetric} & $\mathcal{L}_{wave}$-SVM \cite{akhtar2024advancing} & $\mathcal{L}_{RoBoSS}$-SVM$^{\dagger}$ \\ \hline
Avg Acc. $\pm$ Avg. Std.  & 73.8±6.17 & 74.05±6.29 & 74.06±6.91 & 55.19±2.7 & \underline{74.73±6.98} & \textbf{79.42±6.28} \\ \hline
Avg. time & 0.0051 & 0.0036 & \underline{0.0017} & 0.002 & 0.003 & \textbf{0.0014} \\ \hline
Avg. rank & 3.86 & 3.63 & 3.31 & 6 & \underline{3.14} & \textbf{1.06} \\ \hline
\multicolumn{7}{l}{Acc., Avg., and Std. stand for accuracy, average, and standard deviation, respectively. $^{\dagger}$ signifies the proposed model.}\\
 \multicolumn{7}{l}{Boldface highlights the top-performing model, while underlining indicates the second-best model. }
\end{tabular}%
}
\end{table*}

\begin{table*}[h]
\caption{The average results of $\mathcal{L}_{RoBoSS}$-SVM along with the compared models on the BreaKHis dataset.}
\centering
\label{tab:breast cancer table}
\resizebox{17cm}{!}{%
\begin{tabular}{lcccccc}
\hline
Model & $\mathcal{L}_{hinge}$-SVM \cite{cortes1995support}             & $\mathcal{L}_{pin}$-SVM \cite{huang2013support}           & $\mathcal{L}_{LINEX}$-SVM \cite{ma2019linex}         & $\mathcal{L}_{qtse}$-SVM \cite{zhao2022asymmetric} & $\mathcal{L}_{wave}$-SVM \cite{akhtar2024advancing} & $\mathcal{L}_{RoBoSS}$-SVM$^{\dagger}$ \\ \hline
Avg. Acc. $\pm$ Avg. Std. & 59.28±5.99 & 60.09±4.96 & \underline{60.41±5.16} & 59.6±4.6 & 60.32±4.15 & \textbf{63.25±5.03} \\ \hline
Avg. time & 0.0056 & 0.0278 & 0.0037 & \textbf{0.0011} & 0.005 & \underline{0.0016} \\ \hline
Avg. rank & 4.22 & \underline{3.47} & 3.72 & 4.59 & 3.63 & \textbf{1.38} \\ \hline
\multicolumn{7}{l}{Acc., Avg., and Std. stand for accuracy, average, and standard deviation, respectively. $^{\dagger}$ signifies the proposed model.}\\
 \multicolumn{7}{l}{$^{\dagger}$ signifies the proposed model while Boldface highlights the top-performing model, while underlining indicates the second-best model. }
\end{tabular}%
}
\end{table*}

\subsection{Evaluation on Biomedical datasets}
In this subsection, we provide the experimental results on publicly available biomedical datasets. Specifically, the electroencephalogram (EEG) signal dataset and the breast cancer (BreaKHis) dataset.
\par
The EEG data \cite{andrzejak2001indications} includes five sets: $A$, $B$, $O$, $C$, and $S$. Each contains $100$ single-channel EEG signals that were sampled at $173.61$ hertz with a duration of $23.6$ seconds. The sets $O$ and $C$ stand for the subject's eyes open and closed signals, respectively. Sets $A$ and $B$ provide the EEG signal that represents the subject's interictal state. The seizure activity signal is contained in set $S$. The feature selection process is the same as opted in \cite{ganaie2022eeg}. The average experimental results on EEG datasets are displayed in Table \ref{tab:EEG-table}. The results highlight the superior performance of the proposed $\mathcal{L}_{RoBoSS}$-SVM model. The $\mathcal{L}_{RoBoSS}$-SVM achieves an average accuracy of 79.42\% with a standard deviation of 6.28, outperforming all baseline models. The closest competitor, $\mathcal{L}_{wave}$-SVM, has an average accuracy of 74.73\% with a standard deviation of 6.98. The $\mathcal{L}_{RoBoSS}$-SVM model also demonstrates the lowest average training time of 0.0014 seconds and the best average rank of 1.06, indicating its overall efficiency and robustness. The detailed results on each of the EEG datasets and the corresponding best parameters are available in Tables S.VI and S.IX of the supplement material file, respectively.
\par
Further, we evaluate the models on BreaKHis dataset \cite{spanhol2015dataset}. We employed $1240$ scans from the dataset at a magnification of 400 times. These scans are classified as either benign or malignant. The benign category includes four subclasses: phyllodes tumor (PT) with 115 scans, adenosis (AD) with 106 scans, fibroadenoma (FD) with 237 scans, and tubular adenoma (TA) with 130 scans. The malignant category is divided into lobular carcinoma (LC) with 137 scans, papillary carcinoma (PC) with 138 scans, ductal carcinoma (DC) with 208 scans, and mucinous carcinoma (MC) with 169 scans. To extract features, we employ the same process as outlined in \cite{gautam2020minimum}. The average experimental results on the BreaKHis dataset are shown in Table \ref{tab:breast cancer table}.
The outcomes illustrate the dominance of the $\mathcal{L}_{RoBoSS}$-SVM model, achieving an average accuracy of 63.25\% with a standard deviation of 5.03. This performance surpasses that of all baseline models, with $\mathcal{L}_{wave}$-SVM being the closest at 60.32\% accuracy and a standard deviation of 4.15. Other models like $\mathcal{L}_{hinge}$-SVM, $\mathcal{L}_{pin}$-SVM, $\mathcal{L}_{LINEX}$-SVM, and $\mathcal{L}_{qtse}$-SVM exhibit lower accuracies of 59.28\%, 60.09\%, 60.41\%, and 59.6\%, respectively. Moreover, $\mathcal{L}_{RoBoSS}$-SVM records an average training time of 0.0016 seconds and an average rank of 1.38, indicating its superior performance and efficiency in handling complex biomedical data. The detailed results for each of the BreaKHis datasets and the 
 corresponding best parameters are available in Tables S.VII and S.X of the supplement material file, respectively.
\par
To further support the improved effectiveness of the proposed $\mathcal{L}_{RoBoSS}$-SVM model, we performed a statistical analysis of the models. The comprehensive results of this analysis can be found in Section S.II of the supplement material file.}
\par
Furthermore, to understand the impact of the loss hyperparameters \(a\) and \(\lambda\) on the performance of \(\mathcal{L}_{RoBoSS}\)-SVM, we conducted a sensitivity analysis. The detailed results of this analysis are provided in Section S.III of the supplement material file. This analysis highlights the intricate relationship between the hyperparameters $a$ and $\lambda$, and the model's accuracy. The key observations can be summarized as follows: (1) The parameter \(a\) plays a crucial role in determining the robustness and performance of the model. Higher values of \(a\) generally lead to improved accuracy, suggesting that the loss function's shape significantly impacts the model's ability to generalize. (2) The bounding parameter \(\lambda\) influences the model's performance, though its impact varies across datasets. For some datasets, the choice of \(\lambda\) is critical, while for others, the model remains relatively stable across a wide range of \(\lambda\) values. (3) The interplay between \(a\) and \(\lambda\) is dataset-dependent, highlighting the need for dataset-specific tuning of these hyperparameters to achieve optimal performance. In conclusion, the sensitivity analysis underscores the importance of careful tuning of the loss hyperparameters \(a\) and \(\lambda\) to achieve optimal performance with the $\mathcal{L}_{RoBoSS}$-SVM model.

\section{Conclusions and Future work}
In conclusion, this paper introduced a novel and innovative loss function, RoBoSS, designed to address critical challenges in supervised learning paradigms. The RoBoSS loss function is characterized by its robustness, boundedness, sparsity, and smoothness, making it a promising tool for enhancing the performance of various machine learning tasks. The theoretical analysis of the RoBoSS loss function demonstrates its remarkable properties, including classification-calibration and a rigorous generalization error bound. These theoretical insights establish RoBoSS as a reliable choice for constructing robust models in supervised learning scenarios. Furthermore, by incorporating the RoBoSS loss function into the framework of SVM, we proposed a novel $\mathcal{L}_{RoBoSS}$-SVM  model. This new model not only inherits the well-known strengths of traditional SVM but also significantly bolsters their robustness and performance. The numerical findings on a diverse range of datasets, including KEEL, UCI, EEG, and breast cancer datasets, decisively support the efficacy of the proposed $\mathcal{L}_{RoBoSS}$-SVM model.
\par
In future work, researchers could focus on developing adaptive methods to dynamically and efficiently adjust the loss hyperparameters \(a\) and \(\lambda\) during the training process, eliminating the need for manual tuning. The loss function is the backbone of machine learning and deep learning models, guiding the model's training process. The choice of loss function determines how well the model learns from the data, how it handles outliers and noise, and how effectively it generalizes to unseen data. Given the nice theoretical properties of the RoBoSS loss function, future research can explore its integration with cutting-edge deep learning and machine learning models. This exploration could lead to the development of novel algorithms that achieve superior performance in various applications.
\section*{Acknowledgment}
This project is funded by the Science and Engineering Research Board (SERB), Government of India, under the Mathematical Research Impact-Centric Support (MATRICS) scheme, grant number MTR/2021/000787. Mohd. Arshad receives funding support from SERB under the Core Research Grant (CRG/2023/001230). Mushir Akhtar acknowledges the Council of Scientific and Industrial Research (CSIR), New Delhi, for providing fellowship for his research under grant number 09/1022(13849)/2022-EMR-I.

\bibliographystyle{IEEEtranN}
\bibliography{refs.bib}

\begin{IEEEbiography}
[{%
\includegraphics[width=1in,height=1.25in,clip,keepaspectratio]{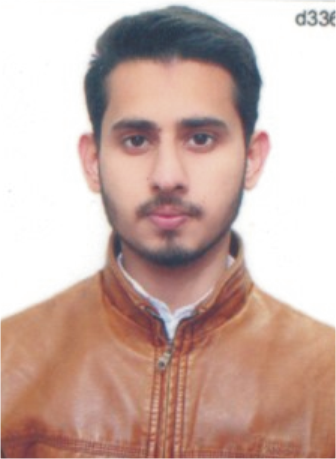}%
}]
{Mushir Akhtar} is currently a Ph.D. scholar in the Department of Mathematics at the Indian Institute of Technology Indore, India, under the supervision of Prof. M. Tanveer and Dr. Mohd. Arshad. He received his B.Sc. and M.Sc. degrees in Mathematics from CCS University, Meerut, India, in 2018 and 2020, respectively, and is a Gold Medalist in M.Sc. Mathematics. His research focuses on developing innovative loss functions for machine learning and deep learning models, with a particular emphasis on applications in the biomedical domain.
\end{IEEEbiography}

\begin{IEEEbiography}
[{%
\includegraphics[width=1in,height=1.25in,clip,keepaspectratio]{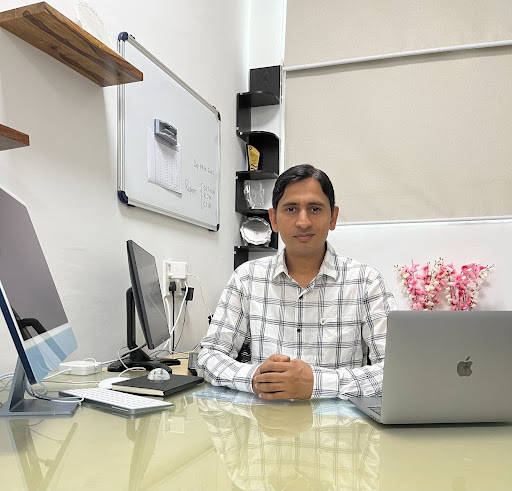}%
}]
{M. Tanveer} is Professor and Ramanujan Fellow at the Department of Mathematics of the Indian Institute of Technology Indore. Prior to that, he worked as a Postdoctoral Research Fellow at the Rolls-Royce@NTU Corporate Lab of the Nanyang Technological University, Singapore. He received the Ph.D degree in Computer Science from the Jawaharlal Nehru University, New Delhi, India. His research interests include support vector machines, optimization, machine learning, deep learning, applications to Alzheimer's disease and dementia. He has published over 160 refereed journal papers of international repute. His publications have over 7100 citations with h index 41 (Google Scholar, September 2024). Recently, he has been listed in the world's top $2\%$ scientists in the study carried out by Stanford University, USA. He has served on review boards for more than 100 scientific journals and served for scientific committees of various national and international conferences. He is the recipient of the INNS Aharon Katzir Young Investigator Award for 2024, IIT Indore Best Research Paper Award for 2023, Asia Pacific Neural Network Society Young Researcher Award for 2022, 29th ICONIP Best Research Paper Award for 2022. He is/was the Associate/Action Editor of IEEE Transactions on Neural Networks and Learning Systems (2022 - 2024), Pattern Recognition, Elsevier (2021 - ), Neural Networks, Elsevier (2022 - ), Engineering Applications of Artificial Intelligence, Elsevier (2022 - ), Neurocomputing, Elsevier (2022 - ), Cognitive Computation, Springer (2022 - ), Applied Soft Computing, Elsevier (2022 - ). He is/was Guest Editor in Special Issues of several journals including IEEE Transactions on Fuzzy Systems, ACM Transactions of Multimedia (TOMM), Applied Soft Computing, Elsevier, IEEE Journal of Biomedical Health and Informatics, IEEE Transactions on Emerging Topics in Computational Intelligence and Annals of Operations Research, Springer. He has also co-edited one book in Springer on machine intelligence and signal analysis. He has organized many international/ national conferences/ symposiums/ workshops as General Chair/ Organizing Chair/ Coordinator, and delivered talks as Keynote/Plenary/invited speaker in many international conferences and Symposiums. He has organized several special sessions in reputed conferences including WCCI, IJCNN, IEEE SMC, IEEE SSCI, ICONIP. Amongst other distinguished, international conference chairing roles, he was the General Chair for 29th International Conference on Neural Information Processing (ICONIP2022) (the world's largest and top technical event in Computational Intelligence). Tanveer is currently the Principal Investigator (PI) or Co-PI of 12 major research projects funded by Government of India including Department of Science and Technology (DST), Science \& Engineering Research Board (SERB) and Council of Scientific \& Industrial Research (CSIR), MHRD-SPARC, ICMR. He is an Elected Board of Governors of Asian Pacific Neural Network Society (APNNS) and Elected INSA Associate Fellow. He was recently honored with the prestigious INSA Distinguished Lecture Award for 2024.
\end{IEEEbiography}

\begin{IEEEbiography}
[{%
\includegraphics[width=1in,height=1.25in,clip,keepaspectratio]{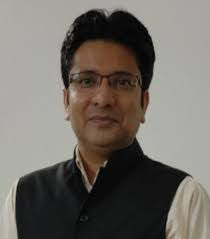}%
}]
{Mohd. Arshad} is an Assistant Professor in the Department of Mathematics at the Indian Institute of Technology (IIT) Indore, India. Prior to his tenure at IIT Indore, he contributed significantly to the Department of Statistics and Operations Research at Aligarh Muslim University, India. Earning his Ph.D. in Statistics from the prestigious Indian Institute of Technology Kanpur, India, in 2015, Dr. Arshad’s academic journey has been marked by excellence. He distinguished himself early on as a Gold Medalist in M.Sc. Statistics from CSJM University Kanpur, India. Dr. Arshad’s research spans various statistics and data science domains, including ranking and selection, copula modelling, machine learning, data science, estimation theory, and generalized order statistics. His contributions to these areas are reflected in over 40 research papers published in esteemed international journals. At IIT Indore, Dr. Arshad leads the Statistical Modeling and Simulation (SMS) Research Group, comprising Ph.D., postdoc, and UG/PG project students. Together, they tackle theoretical and computational challenges in statistics and data science, contributing to cutting-edge advancements in the field. Beyond research, Dr. Arshad is actively involved in academia, serving on the editorial boards of multiple journals and scientific societies. He has also co-authored and edited two books, further cementing his impact on the scholarly community. With over a decade of teaching experience, Dr. Arshad has imparted knowledge and inspired countless students through undergraduate and postgraduate courses. His dedication to education and research underscores his commitment to advancing the frontiers of statistical science.

\end{IEEEbiography}

\end{document}


\title{Supplementary Material for the Manuscript ``RoBoSS: A Robust, Bounded, Sparse, and Smooth Loss Function for Supervised Learning"}
\author{Mushir Akhtar, \IEEEmembership{Graduate Student Member,~IEEE}, M. Tanveer{$^*$}, \IEEEmembership{Senior Member,~IEEE}, Mohd. Arshad}


\maketitle

\section{Experimental setup and parameter selection}
The experiments are conducted using MATLAB R2023a on a Windows 10 system equipped with an Intel(R) Xenon(R) Platinum 8260 CPU running at 2.30 GHz and 256 GB of RAM.
To map the input samples into a higher-dimensional space, the Gaussian kernel function is used. It is defined as $\kappa\left(x_k,x_j\right)=\exp\left(-\left\|x_k-x_j\right\|^2 / \sigma^2\right)$ , where $\sigma$ is the kernel parameter. Before training, each dataset is normalized in the interval $\left[-1,1\right]$. For each model, the penalty parameter $\mathcal{C}$ and the kernel parameter $\sigma$ are chosen from the range $\{10^i\,|\,i=-6:1:6\}$. For $\mathcal{L}_{pin}$-SVM, the hyperparameter $\tau$ is selected from $\{0,0.3,0.5,0.7,0.9\}$. For  $\mathcal{L}_{LINEX}$-SVM, and $\mathcal{L}_{qtse}$-SVM  the range of loss hyperparameter is taken the same as in \cite{ma2019linex} and \cite{zhao2022asymmetric}, respectively. For $\mathcal{L}_{wave}$-SVM, the loss hyperparametr $a$ is selected from the range $\left[-5:1:5\right]$ and the bounding parameter $\lambda$ is fixed to $1$. For the proposed $\mathcal{L}_{RoBoSS}$-SVM, the loss parameters $a$ and $\lambda$ are chosen from the range $\left[0:0.1:5\right]$ and  $\left[0.1:0.1:2\right]$, respectively. The parameters for the NAG-based algorithm are experimentally set as: (\RNum{1}) initial model parameter $\beta_0=0.01$, (\RNum{2}) initial velocity  $v_0=0.01$, (\RNum{3}) initial learning rate $\alpha=0.1$, (\RNum{4}) learning decay factor $\eta=0.1$, (\RNum{5}) momentum parameter $r=0.6$, (\RNum{6}) two distinct minibatch size configurations are used based on the size of the dataset: $s = 2^2$ for datasets with less than $100$ samples and $s = 2^5$ for datasets with $100$ or more samples, (\RNum{7}) maximum iteration number $N=1000$.
\par
The choice of hyperparameters has a significant impact on the models' performance. In order to optimize them, we employ $5$-fold cross-validation along with grid search. In this, the dataset is randomly split into five disjoint subsets, where one subset serves as the test set and the remaining four are used for training. For each set of hyperparameters, we determined the testing accuracy for all five subsets separately. Then, for each hyperparameter set, we calculate the mean testing accuracy by taking the average of these five accuracy values. The best mean testing accuracy is chosen as the testing accuracy of the model.
\par 
The performance of the models is evaluated using the accuracy metric, which is defined as
\begin{align*}{}
 \text {Accuracy}=\frac{\text { TP+TN  }}{\text { TP+TN+FP+FN}} \times 100,  
\end{align*}
where TP, TN, FP, and FN are true positive, true negative, false positive, and false negative, respectively. To further analyze the model's performance, we also evaluate the rank and training time. The reported times reflect only the duration required to train the models using the best hyperparameters.
\par
The detailed procedure for artificially introducing outliers and label noise into the training dataset is outlined as follows: \\
\textbf{Methodology for Introducing Outliers:}\\
Outliers have been systematically introduced into the datasets at varying levels: 5\%, 10\%, 20\%, and 30\% of the total number of samples. For each dataset, the number of outliers has been calculated based on the specified percentage. Random samples have then been selected to serve as outliers. For each chosen sample, a feature has been randomly selected, and its value has been altered by multiplying it by an outlier factor of 10. This systematic introduction of outliers allows us to rigorously test the robustness and stability of the $\mathcal{L}_{RoBoSS}$-SVM model in handling data contamination.\\
\textbf{Methodology for Introducing Label Noise:}\\
Label noise has been introduced by randomly flipping the labels of a certain percentage of the samples. Specifically, noise levels of 5\%, 10\%, 20\%, and 30\% have been used. For each dataset, the number of labels to be flipped has been determined based on the specified noise level. Random samples have been selected, and their labels have been inverted to create noise.
\section{Statistical Analysis of Results}
For statistical evaluation, we employed the Friedman test followed by the Nemenyi post hoc test to assess the relative performance of these models.\\
\textbf{Friedman test:} The Friedman test \cite{friedman1940comparison} is employed to statistically analyze the significance of the models. In this test, each model is ranked on each dataset separately, with the best-performing model securing rank $1$, the second-best model getting rank $2$, and so on. Under the null hypothesis, all the models are equivalent, i.e., the average rank of each model is equal. The Friedman statistic follows the chi-squared $\chi^2_F$ distribution with $p-1$ degrees of freedom (d.f.), where $p$ denotes the number of models and is given by:
\begin{align}{} \label{chisquareequation}
\chi_F^2=\frac{12 D}{p(p+1)}\left[\sum_e R_e^2-\frac{p(p+1)^2}{4}\right],
\end{align}
where $D$ denotes the number of datasets and $R_e$ is the mean rank of $e^{th}$ of the $p$ models. The Friedman statistic is undesirably conservative and thus a better statistic is derived by \citet{iman1980approximations} as:
\begin{align}{} \label{ffequation}
F_F=\frac{(D-1) \chi_F^2}{D(p-1)-\chi_F^2},
\end{align}
which follows $F$ distribution with $((p-1),(p-1)(D-1))$ d.f.. From the statistical $F$-distribution table, at $5 \%$ level of significance, we find the value of $F((p-1),(p-1)(D-1))$. If $F_F > F((p-1),(p-1)(D-1)) $, we reject the null hypothesis. In this case, substantial differences exist among the models. Table \ref{tab:Friedman test-table} presents the results of the Friedman test on D1 category UCI and KEEL datasets, the EEG dataset, and the BreaKHis dataset. The outcomes demonstrate that significant differences exist among the proposed $\mathcal{L}_{RoBoSS}$-SVM and baseline models.\\
\textbf{Nemenyi post hoc test:} In Nemenyi post hoc test \cite{demvsar2006statistical}, all models are compared pairwise. The performance of the two models is substantially different if the corresponding mean ranks differ by a certain threshold value (critical difference, $C.D.$). If the difference between comparing models mean ranks exceeds $C.D$., the model with a higher mean rank is statistically better than the model with a lower mean rank. The $C.D.$ is calculated as:
\begin{align}{}
    C.D.=q_\alpha \sqrt{\frac{p(p+1)}{6D}},
\end{align}
where $q_\alpha$ are based on the studentized range statistic divided by $\sqrt{2}$ and called critical value for the two-tailed Nemenyi test. At 5 \% level of significance, we can simply calculate that the values of $C.D.$ for D1 category UCI and KEEL datasets, the EEG dataset, and the BreaKHis dataset are $0.85$, $1.33$, and $1.88$, respectively. Tables \ref{tab:Nementi-table-for D1}, \ref{tab:Nemenyi-table-EEG}, and \ref{tab:Nemenyi-table-BreaKHis} present the results of the Nemenyi post hoc test on D1 category UCI and KEEL datasets, the EEG dataset, and the BreaKHis dataset, respectively.
\begin{table}[]
\centering
\caption{\small{Illustrate the results of the Friedman test on D1 category UCI and KEEL datasets, the EEG dataset, and the BreaKHis dataset.}}
\label{tab:Friedman test-table}
\resizebox{9cm}{!}{%
\begin{tabular}{|l|c|c|c|c|c|c|}
\hline
Dataset              & $p$ & $D$  & $\chi^2_F$& $F_F$     & $F((p-1),(p-1)(D-1))$ & \begin{tabular}[c]{@{}c@{}}Significant difference\\ (As per Friedman test)\end{tabular}  \\ \hline
D1 category dataset & 6 & 79 & 81.442     & 20.26  &   2.24       & Yes                    \\ \hline
EEG dataset          & 6 & 32 & 114.43  & 77.844 &   2.27      & Yes                    \\ \hline
BreaKHis dataset     & 6 & 16 & 28.969     & 8.515   &   2.35      & Yes                    \\ \hline
\end{tabular}%
}
\end{table}


\begin{table}[]
\centering
\caption{\small{Differences in the rankings of the proposed $\mathcal{L}_{RoBoSS}$-SVM model
against baseline models on D1 category UCI and KEEL datasets.}}
\label{tab:Nementi-table-for D1}
\resizebox{9cm}{!}{%
\begin{tabular}{|l|c|c|c|}
\hline
Model & Average rank & Rank difference & \begin{tabular}[c]{@{}c@{}}Significant difference\\ (As per Nemenyi post hoc test)\end{tabular} \\ \hline
$\mathcal{L}_{hinge}$-SVM \cite{cortes1995support}      & 3.35 & 1.19 & Yes \\ \hline
$\mathcal{L}_{pin}$-SVM \cite{huang2013support}   & 2.96 & 0.8 & No  \\ \hline
$\mathcal{L}_{LINEX}$-SVM \cite{ma2019linex}   & 3.96 & 1.8 & Yes \\ \hline
$\mathcal{L}_{qtse}$-SVM \cite{zhao2022asymmetric}  & 4.45 & 2.29 & Yes \\ \hline
$\mathcal{L}_{wave}$-SVM \cite{akhtar2024advancing}  & 4.12 & 1.96 & Yes \\ \hline
$\mathcal{L}_{RoBoSS}$-SVM (Proposed) & 2.16 & -    & N/A \\ \hline
\end{tabular}%
}
\end{table}

\begin{table}[]
\centering
\caption{\small{Differences in the rankings of the proposed $\mathcal{L}_{RoBoSS}$-SVM model
against baseline models on the EEG dataset.}}
\label{tab:Nemenyi-table-EEG}
\resizebox{9cm}{!}{%
\begin{tabular}{|l|c|c|c|}
\hline
Model & Average rank & Rank difference & \begin{tabular}[c]{@{}c@{}}Significant difference\\ (As per Nemenyi post hoc test)\end{tabular} \\ \hline
$\mathcal{L}_{hinge}$-SVM \cite{cortes1995support}       & 3.86 & 2.8 & Yes \\ \hline
$\mathcal{L}_{pin}$-SVM \cite{huang2013support}    & 3.63 & 2.57 & Yes \\ \hline
$\mathcal{L}_{LINEX}$-SVM \cite{ma2019linex}    & 3.31    & 2.25    & Yes \\ \hline
$\mathcal{L}_{qtse}$-SVM \cite{zhao2022asymmetric}    & 6    & 4.94    & Yes \\ \hline
$\mathcal{L}_{wave}$-SVM \cite{akhtar2024advancing}   & 3.14    & 2.08    & Yes \\ \hline
$\mathcal{L}_{RoBoSS}$-SVM (Proposed) & 1.06    & -    & N/A \\ \hline
\end{tabular}%
}
\end{table}
\begin{table}[]
\centering
\caption{\small{Differences in the rankings of the proposed $\mathcal{L}_{RoBoSS}$-SVM model
against baseline models on the BreaKHis dataset.}}
\label{tab:Nemenyi-table-BreaKHis}
\resizebox{9cm}{!}{%
\begin{tabular}{|l|c|c|c|}
\hline
Model & Average rank & Rank difference & \begin{tabular}[c]{@{}c@{}}Significant difference\\ (As per Nemenyi post hoc test)\end{tabular} \\ \hline
$\mathcal{L}_{hinge}$-SVM \cite{cortes1995support}      & 4.22 & 2.84 & Yes \\ \hline
$\mathcal{L}_{pin}$-SVM \cite{huang2013support}   & 3.47 & 2.09 & Yes \\ \hline
$\mathcal{L}_{LINEX}$-SVM \cite{ma2019linex}    & 3.72    & 2.34    & Yes \\ \hline
$\mathcal{L}_{qtse}$-SVM \cite{zhao2022asymmetric}     & 4.59    & 3.21    & Yes \\ \hline
$\mathcal{L}_{wave}$-SVM \cite{akhtar2024advancing}     & 3.63    & 2.25    & Yes \\ \hline
$\mathcal{L}_{RoBoSS}$-SVM (Proposed) & 1.38    & -    & N/A \\ \hline
\end{tabular}%
}
\end{table}

\section{Sensitivity analysis}
\begin{figure*}[htp]
\begin{minipage}{.5\linewidth}
\centering
\subfloat[abalone9-18]{\label{sentivity1} \includegraphics[scale=0.55]{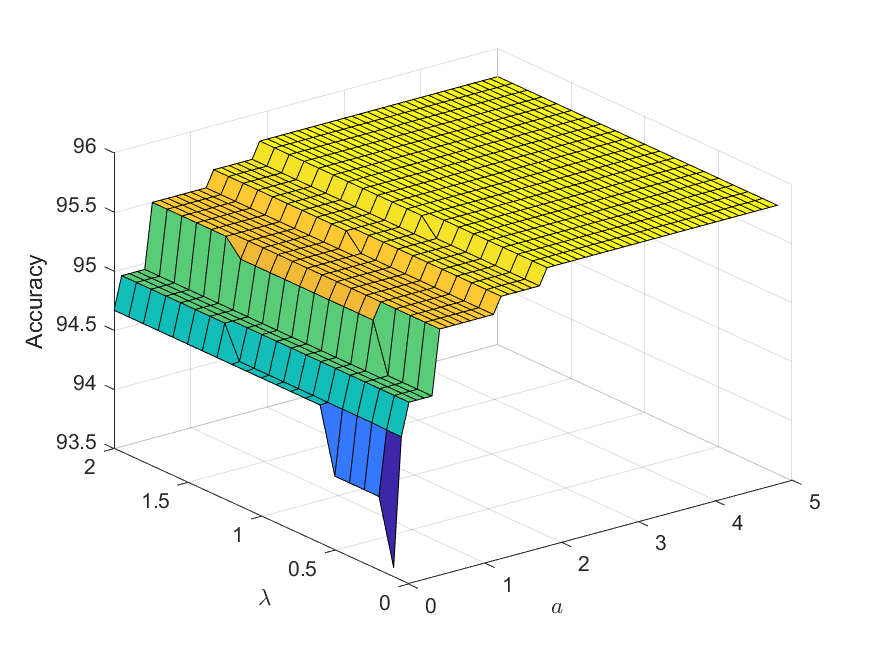}}
\end{minipage}
\begin{minipage}{.5\linewidth}
\centering
\subfloat[echocardiogram]{\label{sentivity2} \includegraphics[scale=0.55]{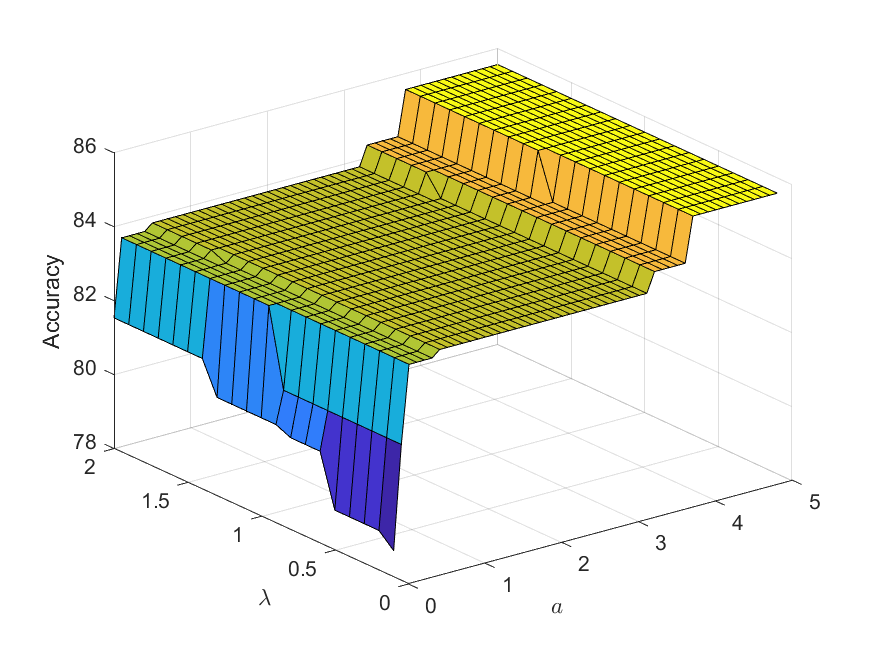}}
\end{minipage}
\begin{minipage}{.5\linewidth}
\centering
\subfloat[titanic]{\label{sentivity3} \includegraphics[scale=0.55]{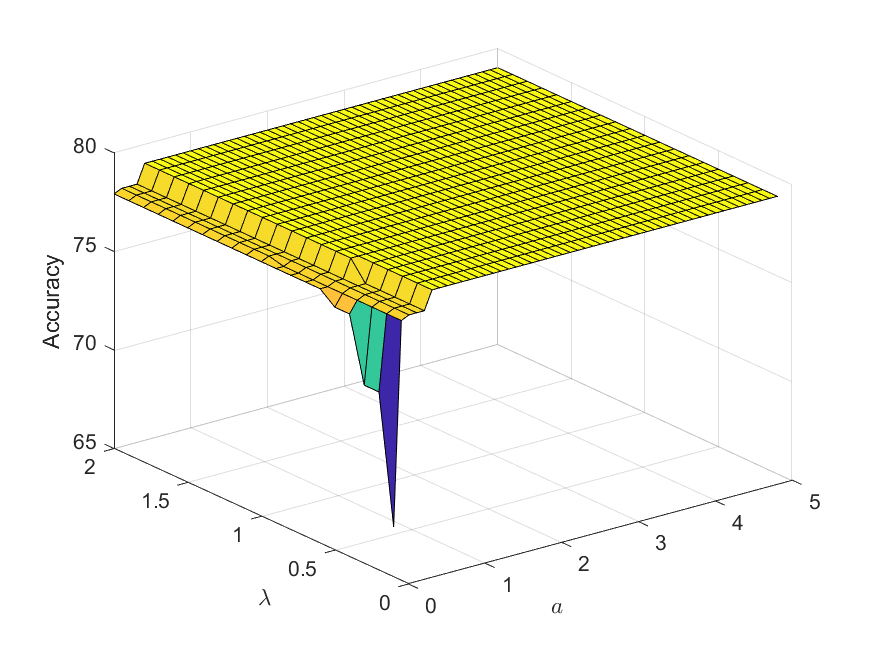}}
\end{minipage}
\begin{minipage}{.5\linewidth}
\centering
\subfloat[ecoli3]{\label{sentivity4} \includegraphics[scale=0.55]{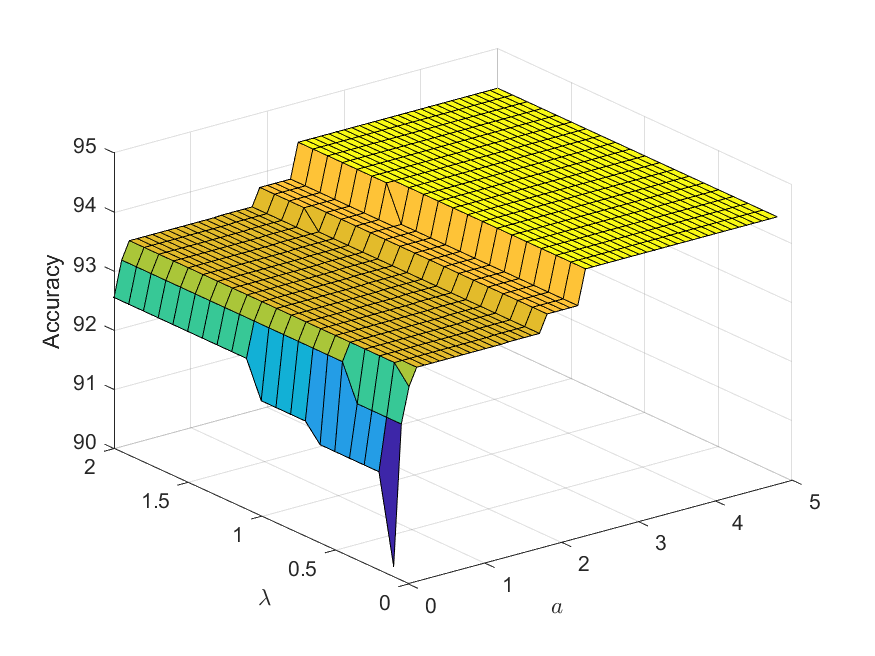}}
\end{minipage}
\caption{Sensitivity analysis of the \(\mathcal{L}_{RoBoSS}\)-SVM model with respect to the loss hyperparameters \(a\) and \(\lambda\) on abalone9-18, echocardiogram, titanic, and ecoli3 datasets. The 3D surface plots illustrate the model's accuracy variations as \(a\) and \(\lambda\) are systematically adjusted while keeping other hyperparameters fixed at their optimal values.}
\label{effect of parameter lambda and a}
\end{figure*}

To understand the impact of the loss hyperparameters $a$ and $\lambda$ on the performance of the proposed $\mathcal{L}_{RoBoSS}$-SVM model, we conduct a sensitivity analysis using four diverse datasets: abalone9-18, echocardiogram, titanic, and ecoli3. The values of $a$ and $\lambda$ are varied systematically within the specified range while the other hyperparameters are fixed at their optimal values. For each combination of $a$ and $\lambda$, the model's accuracy is recorded. The results are plotted in three-dimensional surface plots to visualize the sensitivity of the model's accuracy to changes in $a$ and $\lambda$. The sensitivity plots for each of the four datasets are presented in Fig. \ref{effect of parameter lambda and a}. These plots highlight the intricate relationship between the hyperparameters $a$ and $\lambda$, and the model's accuracy. Fig. \ref{sentivity1} reveals that the accuracy of the abalone9–18 dataset stabilizes at higher values of $a$, with $\lambda$ having a moderate influence. The model exhibits robustness across a wide range of $\lambda$ values when $a$ is sufficiently large. For the echocardiogram (see Fig. \ref{sentivity2}), the accuracy shows a strong dependency on $a$, with higher values leading to improved performance. The impact of $\lambda$ is less pronounced but still noticeable. The sensitivity plot for the titanic dataset (see Fig. \ref{sentivity3}) indicates a consistent accuracy across various values of $a$ and $\lambda$, with a notable dip in performance at lower values of both parameters. The model achieves its highest accuracy when both $a$ and $\lambda$ are set to higher values. For ecoli3 dataset (see Fig. \ref{sentivity4}), the model's accuracy is highly sensitive to changes in both $a$ and $\lambda$, with specific parameter combinations resulting in significantly higher accuracy. This analysis provides valuable insights into the behavior of the $\mathcal{L}_{RoBoSS}$-SVM model across different datasets. The key observations can be summarized as follows: (1) The parameter \(a\) plays a crucial role in determining the robustness and performance of the model. Higher values of \(a\) generally lead to improved accuracy, suggesting that the loss function's shape significantly impacts the model's ability to generalize. (2) The bounding parameter \(\lambda\) influences the model's performance, though its impact varies across datasets. For some datasets, the choice of \(\lambda\) is critical, while for others, the model remains relatively stable across a wide range of \(\lambda\) values. (3) The interplay between \(a\) and \(\lambda\) is dataset-dependent, highlighting the need for dataset-specific tuning of these hyperparameters to achieve optimal performance. In conclusion, the sensitivity analysis underscores the importance of careful tuning of the loss hyperparameters \(a\) and \(\lambda\) to achieve optimal performance with the $\mathcal{L}_{RoBoSS}$-SVM model.

\begin{tiny}
\begin{table*}[h]
 \centering
 \caption{\vspace{0.1mm}The average classification accuracies, training times, and ranks of the proposed $\mathcal{L}_{RoBoSS}$-SVM and baseline models on each $79$ D1 category UCI and KEEL datasets.}
 \resizebox{15cm}{!}{
\label{tab:small-table}
\begin{tabular}{lcccccc}
\hline
Model & $\mathcal{L}_{hinge}$-SVM \cite{cortes1995support}             & $\mathcal{L}_{pin}$-SVM \cite{huang2013support}           & $\mathcal{L}_{LINEX}$-SVM \cite{ma2019linex}         & $\mathcal{L}_{qtse}$-SVM \cite{zhao2022asymmetric} & $\mathcal{L}_{wave}$-SVM \cite{akhtar2024advancing} & $\mathcal{L}_{RoBoSS}$-SVM$^{\dagger}$ \\ \hline
\begin{tabular}[c]{@{}l@{}}Dataset\\ samples, features\end{tabular} & Acc. $\pm$ Std. , time & Acc. $\pm$ Std. , time & Acc. $\pm$ Std. , time & Acc. $\pm$ Std. , time &Acc. $\pm$ Std. , time & Acc. $\pm$ Std. , time \\ \hline
\begin{tabular}[c]{@{}l@{}}acute inflammation\\  120,6\end{tabular} & \textbf{100±0}, 0.0049 & \textbf{100±0}, 0.0047 & 90±22.36, 0.0024 & 75±37.27, 0.0016 & \textbf{100±0}, 0.0033 & \underline{98.33±3.73}, 0.001 \\
\begin{tabular}[c]{@{}l@{}}balloons\\  16,4\end{tabular} & 53.33±50.55, 0.0079 & 80±18.26, 0.002 & 73.33±27.89, 0.0023 & 86.67±18.26, 0.0015 & \underline{89.67±17.08}, 0.0004 & \textbf{100±0}, 0.0011 \\
\begin{tabular}[c]{@{}l@{}}fertility\\  100,9\end{tabular} & 88±10.37, 0.0054 & 89±9.62, 0.0075 & 88±10.37, 0.0017 & 88±10.37, 0.0008 & \underline{90±10.37}, 0.0024 & \textbf{92±7.58}, 0.0009 \\
\begin{tabular}[c]{@{}l@{}}molec biol promoter\\  106,57\end{tabular} & 56.75±10.8, 0.0081 & 90.48±21.3, 0.0052 & 71.39±42.68, 0.0018 & 90.48±21.3, 0.0009 & \underline{91.23±19.34}, 0.0033 & \textbf{92.38±17.04}, 0.0008 \\
\begin{tabular}[c]{@{}l@{}}parkinsons\\  195,22\end{tabular} & 79.49±23.15, 0.0117 & 80±21.48, 0.0043 & 80±21.48, 0.0018 & 80±21.48, 0.0008 & \underline{82.46±16.07}, 0.0024 & \textbf{89.74±9.59}, 0.0009 \\
\begin{tabular}[c]{@{}l@{}}pittsburg bridges T OR D\\  102,7\end{tabular} & \underline{86.14±13.95}, 0.0078 & \underline{86.14±13.95}, 0.0029 & \underline{86.14±13.95}, 0.0028 & \underline{86.14±13.95}, 0.0008 & \underline{86.14±13.95}, 0.0028 & \textbf{92.14±5.69}, 0.0008 \\
\begin{tabular}[c]{@{}l@{}}breast cancer\\  286,9\end{tabular} & 70.18±44.62, 0.0047 & 70.18±44.62, 0.009 & 70.18±44.62, 0.0033 & 70.62±24.14, 0.0037 & \underline{73.78±17.13}, 0.0043 & \textbf{81.81±12.26}, 0.0011 \\
\begin{tabular}[c]{@{}l@{}}breast cancer wisc prog\\  198,33\end{tabular} & 76.35±8.95, 0.0036 & 76.35±8.95, 0.0057 & \underline{77.35±8.32}, 0.0047 & 76.35±8.95, 0.0016 & 76.35±8.95, 0.0025 & \textbf{78.82±7.94}, 0.0011 \\
\begin{tabular}[c]{@{}l@{}}congressional voting\\  435,16\end{tabular} & 62.07±3.04, 0.013 & 62.3±2.62, 0.0153 & 61.38±1.74, 0.0037 & 62.07±2.15, 0.0028 & \underline{62.38±1.74}, 0.0027 & \textbf{63.22±3.25}, 0.0011 \\
\begin{tabular}[c]{@{}l@{}}echocardiogram\\  131,10\end{tabular} & 75.44±11.83, 0.0042 & 77.01±7.85, 0.0051 & 77.04±7.35, 0.0036 & 75.58±11.04, 0.0017 & \underline{77.81±8.01}, 0.0027 & \textbf{85.44±9.63}, 0.0011 \\
\begin{tabular}[c]{@{}l@{}}haberman survival\\  306,3\end{tabular} & 73.49±8.48, 0.0046 & 73.49±8.48, 0.0088 & \underline{73.82±8.2}, 0.003 & 73.82±8.68, 0.003 & 73.49±8.48, 0.0028 & \textbf{76.11±8.71}, 0.0011 \\
\begin{tabular}[c]{@{}l@{}}hepatitis\\  155,19\end{tabular} & 83.23±11.27, 0.0039 & \underline{83.87±7.21}, 0.0035 & 81.94±8.1, 0.0036 & 81.29±11.5, 0.0017 & 82.58±11.08, 0.0028 & \textbf{89.03±8.1}, 0.0011 \\
\begin{tabular}[c]{@{}l@{}}horse colic\\  368,25\end{tabular} & \textbf{80.17±4.3}, 0.0074 & \textbf{80.17±4.3}, 0.0106 & 76.09±3.24, 0.0035 & 65.77±6.65, 0.0017 & 76.08±3.38, 0.0029 & \underline{79.59±9.5}, 0.0011 \\
\begin{tabular}[c]{@{}l@{}}ionosphere\\  351,33\end{tabular} & 64.71±21.68, 0.0104 & 67.79±6.61, 0.028 & \textbf{81.78±9.51}, 0.0045 & 69.61±23.45, 0.0021 & 71.35±5.09, 0.0045 & \underline{72.56±4.84}, 0.0012 \\
\begin{tabular}[c]{@{}l@{}}planning\\  182,12\end{tabular} & 71.38±8.85, 0.0038 & 71.38±8.85, 0.0042 & 71.38±8.85, 0.0027 & 71.38±8.85, 0.0028 & \underline{75.79±12.43}, 0.0029 & \textbf{86.91±8.61}, 0.0014 \\
\begin{tabular}[c]{@{}l@{}}spect\\  265,22\end{tabular} & 64.15±6.67, 0.0043 & 65.28±6.62, 0.007 & 65.66±9.09, 0.0033 & 58.49±14.06, 0.0025 & \underline{71.38±8.85}, 0.0024 & \textbf{73.6±3.89}, 0.0012 \\
\begin{tabular}[c]{@{}l@{}}spectf\\  267,44\end{tabular} & \textbf{79.34±20.89}, 0.0039 & \textbf{79.34±20.89}, 0.0061 & \textbf{79.34±20.89}, 0.0032 & \textbf{79.34±20.89}, 0.0027 & 70.28±10.39, 0.0028 & \underline{72.08±4.3}, 0.0012 \\
\begin{tabular}[c]{@{}l@{}}statlog heart\\  270,13\end{tabular} & 77.04±1.66, 0.0062 & 77.04±1.66, 0.0103 & 78.15±4.01, 0.0031 & 72.96±3.84, 0.0019 & \underline{79.34±20.89}, 0.0028 & \textbf{80.48±10.33}, 0.0016 \\
\begin{tabular}[c]{@{}l@{}}bupa or liver-disorders\\  345,6\end{tabular} & 71.88±3.64, 0.0074 & 71.88±3.64, 0.02 & 62.61±8.9, 0.0035 & 62.32±8.64, 0.003 & \underline{76.3±7.1}, 0.0027 & \textbf{80.37±1.66}, 0.0011 \\
\begin{tabular}[c]{@{}l@{}}cleve\\  297,13\end{tabular} & \textbf{78.77±7.4}, 0.0078 & \textbf{78.77±7.4}, 0.0165 & \underline{76.07±5.26}, 0.0036 & 74.36±8.95, 0.0026 & 64.93±6.51, 0.005 & 69.28±3.75, 0.0014 \\
\begin{tabular}[c]{@{}l@{}}crossplane130\\  130,2\end{tabular} & 70.77±8.85, 0.004 & 70.77±8.85, 0.0057 & 70±8.34, 0.003 & 64.62±9.96, 0.0017 & \underline{75.4±4.24}, 0.0037 & \textbf{80.79±4.34}, 0.0012 \\
\begin{tabular}[c]{@{}l@{}}crossplane150\\  150,2\end{tabular} & 62±12.16, 0.0038 & 62±12.16, 0.0081 & 67.33±7.6, 0.0029 & 66.67±8.82, 0.0017 & \underline{70±9.18}, 0.0038 & \textbf{74.62±5.83}, 0.0012 \\
\begin{tabular}[c]{@{}l@{}}ecoli-0-1-4-6vs5\\  280,6\end{tabular} & \textbf{97.5±2.71}, 0.006 & \textbf{97.5±2.71}, 0.0161 & \underline{96.07±2.65}, 0.0034 & 96.07±3.43, 0.0018 & 73.33±5.58, 0.0037 & 74±4.35, 0.0015 \\
\begin{tabular}[c]{@{}l@{}}ecoli-0-1-4-7vs2-3-5-6\\  336,7\end{tabular} & \underline{96.73±0.67}, 0.0078 & \underline{96.73±0.67}, 0.0262 & 93.75±1.93, 0.0037 & 94.93±3.28, 0.0022 & 96.43±2.82, 0.0051 & \textbf{98.21±1.26}, 0.0015 \\
\begin{tabular}[c]{@{}l@{}}ecoli-0-1-4-7vs5-6\\  332,6\end{tabular} & \textbf{98.19±0.68}, 0.0078 & \textbf{98.19±0.68}, 0.0187 & 95.18±1.95, 0.0029 & 95.17±2.72, 0.0019 & 95.04±2.37, 0.0037 & \underline{96.12±3.27}, 0.0013 \\
\begin{tabular}[c]{@{}l@{}}ecoli-0-1vs2-3-5\\  244,7\end{tabular} & \underline{96.73±3.09}, 0.0045 & \underline{96.73±3.09}, 0.0304 & 94.68±3.7, 0.0034 & 94.66±4.94, 0.0025 & 95.49±1.84, 0.004 & \textbf{97.59±1.35}, 0.0012 \\
\begin{tabular}[c]{@{}l@{}}ecoli-0-1vs5\\  240,6\end{tabular} & \textbf{97.92±2.55}, 0.0055 & \textbf{97.92±2.55}, 0.0144 & 96.25±3.73, 0.0028 & 95.42±2.28, 0.0017 & 95.09±4.22, 0.0053 & \underline{97.14±3.98}, 0.0013 \\
\begin{tabular}[c]{@{}l@{}}ecoli-0-2-3-4vs5\\  202,7\end{tabular} & \underline{98.5±3.35}, 0.004 & \underline{98.5±3.35}, 0.02 & 97.5±3.54, 0.0036 & 94.55±2.11, 0.003 & 95.42±4.52, 0.0037 & \textbf{98.75±1.14}, 0.0011 \\
\begin{tabular}[c]{@{}l@{}}ecoli-0-2-6-7vs3-5\\  224,7\end{tabular} & \underline{96.44±5.79}, 0.0042 & \underline{96.44±5.79}, 0.0112 & 91.99±6.39, 0.003 & 95.11±5.07, 0.0017 & 95.54±2.1, 0.0039 & \textbf{98±3.26}, 0.0011 \\
\begin{tabular}[c]{@{}l@{}}ecoli-0-3-4-6vs5\\  205,7\end{tabular} & \textbf{97.07±3.18}, 0.004 & \textbf{97.07±3.18}, 0.0151 & 95.61±4.01, 0.0031 & 94.63±2.04, 0.0017 & 93.28±6.52, 0.0048 & \underline{96±5.53}, 0.0011 \\
\begin{tabular}[c]{@{}l@{}}ecoli-0-3-4-7vs5-6\\  257,7\end{tabular} & \textbf{97.66±0.89}, 0.0052 & \textbf{97.66±0.89}, 0.0129 & 94.95±2.6, 0.0037 & 94.53±2.58, 0.0029 & 95.12±3.45, 0.0048 & \underline{97.56±2.44}, 0.0011 \\
\begin{tabular}[c]{@{}l@{}}ecoli-0-4-6vs5\\  203,6\end{tabular} &\underline{97.05±2.67}, 0.0037 & \underline{97.05±2.67}, 0.0206 & 95.57±3.61, 0.0019 & 94.07±3.31, 0.0017 & 94.17±4.09, 0.0056 & \textbf{97.65±1.64}, 0.0011 \\
\begin{tabular}[c]{@{}l@{}}ecoli-0-6-7vs3-5\\  222,7\end{tabular} & \underline{96.4±3.02}, 0.004 & \underline{96.4±3.02}, 0.0137 & 93.68±2.98, 0.0032 & 92.83±3.94, 0.0019 & 93.61±2.14, 0.0038 & \textbf{98.01±2.09}, 0.0011 \\
\begin{tabular}[c]{@{}l@{}}ecoli-0-6-7vs5\\  220,6\end{tabular} & \textbf{97.27±1.9}, 0.0042 & \textbf{97.27±1.9}, 0.011 & 93.64±6.31, 0.0028 & 94.55±4.13, 0.0015 & 94.19±5.34, 0.0038 & \underline{96.87±2.99}, 0.0011 \\
\begin{tabular}[c]{@{}l@{}}ecoli0137vs26\\  311,7\end{tabular} & \underline{96.15±3.3}, 0.007 & \underline{96.15±2.65}, 0.0158 & 94.86±2.62, 0.0029 & 94.55±3.3, 0.0015 & 93.64±6.89, 0.0057 & \textbf{96.82±2.59}, 0.0011 \\
\begin{tabular}[c]{@{}l@{}}ecoli01vs5\\  240,7\end{tabular} & 98.33±1.74, 0.005 & 98.33±1.74, 0.015 & \textbf{99.17±1.14}, 0.0038 & \underline{98.75±1.86}, 0.0015 & 95.19±3.39, 0.0039 & 97.12±2.35, 0.0011 \\
\begin{tabular}[c]{@{}l@{}}ecoli3\\  336,7\end{tabular} & 92.85±3.87, 0.0084 & 92.85±3.87, 0.0267 & 93.75±3.23, 0.003 & 91.37±3.23, 0.0014 & \underline{98.75±2.8}, 0.0039 & \textbf{99.58±0.93}, 0.0011 \\
\begin{tabular}[c]{@{}l@{}}ecoli4\\  336,7\end{tabular} & \textbf{98.52±1.48}, 0.0084 & \textbf{98.52±1.48}, 0.0269 & 97.32±1.25, 0.0036 & \underline{97.92±1.7}, 0.0014 & 93.74±3.87, 0.004 & 94.94±3.75, 0.0011 \\
\begin{tabular}[c]{@{}l@{}}glass2\\  214,9\end{tabular} & 92.05±2.12, 0.0036 & 92.05±2.12, 0.0092 & 92.05±2.12, 0.0029 & 92.05±2.12, 0.0015 & \underline{97.32±1.94}, 0.004 & \textbf{99.11±0.81}, 0.0011 \\
\begin{tabular}[c]{@{}l@{}}glass4\\  214,9\end{tabular} & \textbf{97.19±3.05}, 0.005 & \textbf{97.19±3.05}, 0.0139 & 96.25±2.7, 0.0034 & \underline{96.27±2.64}, 0.0015 & 92.49±4.6, 0.0036 & 93.91±3.58, 0.0011 \\
\begin{tabular}[c]{@{}l@{}}glass5\\  214,9\end{tabular} & 96.73±2.07, 0.0053 & 96.73±2.07, 0.0184 & 95.79±1.95, 0.0029 & \underline{96.74±2.65}, 0.0015 & 96.28±3.89, 0.0008 & \textbf{98.14±1.95}, 0.0011 \\
\begin{tabular}[c]{@{}l@{}}haber\\  306,3\end{tabular} & 73.82±8.03, 0.0087 & 73.82±8.03, 0.0242 & 73.82±8.03, 0.0038 & 73.49±8.48, 0.0015 & \underline{97.66±1.64}, 0.0006 & \textbf{98.14±1.95}, 0.0011 \\
\begin{tabular}[c]{@{}l@{}}haberman\\  306,3\end{tabular} & 73.82±8.03, 0.0089 & 73.82±8.03, 0.0306 & \underline{74.15±7.64}, 0.0037 & 73.49±8.48, 0.0015 & 73.82±8.03, 0.0006 & \textbf{74.8±7.29}, 0.0011 \\
\begin{tabular}[c]{@{}l@{}}iono\\  351,33\end{tabular} & \underline{80.91±5.76}, 0.0079 & \underline{80.91±5.76}, 0.02 & \textbf{81.51±10.99}, 0.0025 & 72.38±22.53, 0.0017 & 74.15±7.64, 0.0007 & 74.8±7.29, 0.0011 \\
\begin{tabular}[c]{@{}l@{}}led7digit-0-2-4-5-6-7-8-9vs1\\  443,7\end{tabular} & \textbf{96.17±0.99}, 0.0114 & \textbf{96.17±0.99}, 0.0294 & \underline{95.27±2.55}, 0.0017 & 94.81±1.73, 0.0015 & 74.7±15.87, 0.0045 & 84.93±10.49, 0.0012 \\
\begin{tabular}[c]{@{}l@{}}new-thyroid1\\  215,5\end{tabular} & 95.35±1.64, 0.0042 & 95.35±1.64, 0.0119 & \underline{97.21±1.04}, 0.0016 & \textbf{97.67±2.33}, 0.0015 & 94.37±4.56, 0.0043 & 96.85±1.84, 0.0011 \\
\begin{tabular}[c]{@{}l@{}}shuttle-6vs2-3\\  230,9\end{tabular} & \textbf{100±0}, 0.005 & \textbf{100±0}, 0.0176 & \textbf{100±0}, 0.0017 & \textbf{100±0}, 0.0015 & 98.14±1.95, 0.006 & \underline{99.53±1.04}, 0.0011 \\
 \hline
\end{tabular}%
}
\end{table*}
\end{tiny}

\begin{tiny}
\begin{table*}[]
\ContinuedFloat
 \centering
 \caption{\vspace{0.1mm}The average classification accuracies, training times, and ranks of the proposed $\mathcal{L}_{RoBoSS}$-SVM and baseline models on each $79$ D1 category UCI and KEEL datasets.}
 \resizebox{15cm}{!}{
\label{tab:small-table}
\begin{tabular}{lcccccc}
\hline
Model & $\mathcal{L}_{hinge}$-SVM \cite{cortes1995support}             & $\mathcal{L}_{pin}$-SVM \cite{huang2013support}           & $\mathcal{L}_{LINEX}$-SVM \cite{ma2019linex}         & $\mathcal{L}_{qtse}$-SVM \cite{zhao2022asymmetric} & $\mathcal{L}_{wave}$-SVM \cite{akhtar2024advancing} & $\mathcal{L}_{RoBoSS}$-SVM$^{\dagger}$ \\ \hline
\begin{tabular}[c]{@{}l@{}}Dataset\\ samples, features\end{tabular} & Acc. $\pm$ Std. , time & Acc. $\pm$ Std. , time & Acc. $\pm$ Std. , time & Acc. $\pm$ Std. , time &Acc. $\pm$ Std. , time & Acc. $\pm$ Std. , time \\ \hline
\begin{tabular}[c]{@{}l@{}}votes\\  435,16\end{tabular} & 89.89±4.97, 0.023 & 90.11±5.37, 0.0316 & \underline{92.41±3.69}, 0.0024 & 85.75±7.48, 0.0015 & \textbf{100±0}, 0.0054 & \textbf{100±0}, 0.0011 \\
\begin{tabular}[c]{@{}l@{}}wpbc\\  194,33\end{tabular} & 76.29±10.16, 0.0038 & 76.29±10.16, 0.0135 & 76.29±10.16, 0.0027 & 76.29±10.16, 0.0015 & \underline{86.9±5.67}, 0.0052 & \textbf{94.94±2.52}, 0.0011 \\
\begin{tabular}[c]{@{}l@{}}yeast1vs7\\  459,8\end{tabular} & \textbf{94.77±1.18}, 0.0158 & \textbf{94.77±1.18}, 0.0507 & \underline{93.47±2.54}, 0.0026 & \underline{93.47±2.54}, 0.0015 & 76.29±10.16, 0.0062 & 79.38±3.63, 0.0011 \\
\begin{tabular}[c]{@{}l@{}}yeast2vs8\\  483,8\end{tabular} & \textbf{97.73±2.12}, 0.0205 & \textbf{97.73±2.12}, 0.0457 & \underline{97.31±2.15}, 0.0027 & 97.11±2.77, 0.0015 & 93.47±2.54, 0.0054 & 94.77±1.41, 0.0011 \\
\begin{tabular}[c]{@{}l@{}}bank\\  4521,16\end{tabular} & 88.67±0.49, 3.1119 & 89.03±0.42, 4.9583 & 88.48±0.55, 0.0043 & 88.48±0.55, 0.0029 & \underline{96.06±5.13}, 0.0008 & \textbf{98.14±1.7}, 0.0011 \\
\begin{tabular}[c]{@{}l@{}}blood\\  748,4\end{tabular} & 76.64±13.29, 0.0687 & 76.64±13.29, 0.117 & 76.24±14.98, 0.0034 & 76.24±14.98, 0.0028 & \underline{88.48±0.55}, 0.0247 & \textbf{88.5±0.56}, 0.0016 \\
\begin{tabular}[c]{@{}l@{}}breast cancer wisc diag\\  569,30\end{tabular} & 79.44±3.43, 0.0138 & 81.54±5.8, 0.0221 & \textbf{84.27±4.69}, 0.0038 & \underline{81.69±6.92}, 0.0023 & 76.24±14.98, 0.0025 & 78.51±11.76, 0.0012 \\
\begin{tabular}[c]{@{}l@{}}chess krvkp\\  3196,36\end{tabular} & 72.3±27.33, 2.0588 & 75.77±23.19, 3.2068 & 58.7±14.93, 0.0047 & 75.77±23.19, 0.0023 & \underline{80.85±4.6}, 0.0026 & \textbf{87.42±6.62}, 0.0012 \\
\begin{tabular}[c]{@{}l@{}}credit approval\\  690,15\end{tabular} & \textbf{84.06±9.78}, 0.0299 & \textbf{84.06±9.78}, 0.0397 & \underline{77.25±6.85}, 0.003 & 76.38±13.81, 0.0025 & 70.64±7.77, 0.0041 & 75.99±23.24, 0.0014 \\
\begin{tabular}[c]{@{}l@{}}cylinder bands\\  512,35\end{tabular} & 60.87±17.95, 0.0118 & 61.07±17.6, 0.0169 & 65±8.83, 0.0034 & 64.79±14.01, 0.0028 & \underline{76.23±8.33}, 0.0026 & \textbf{82.9±9.86}, 0.0012 \\
\begin{tabular}[c]{@{}l@{}}ilpd indian liver\\  583,9\end{tabular} & \textbf{71.35±5.09}, 0.0774 & \textbf{71.35±5.09}, 0.0235 & \textbf{71.35±5.09}, 0.0032 & \textbf{71.35±5.09}, 0.0028 & 64.04±7.65, 0.0028 & \underline{68.32±10.01}, 0.0012 \\
\begin{tabular}[c]{@{}l@{}}mammographic\\  961,5\end{tabular} & \textbf{77.94±5.78}, 0.0917 & \textbf{77.94±5.78}, 0.138 & 73.15±2.91, 0.0034 & 71.08±2.65, 0.003 & 72.74±5.76, 0.0042 & \underline{77.21±2.02}, 0.0013 \\
\begin{tabular}[c]{@{}l@{}}oocytes trisopterus nucleus 2f\\  912,25\end{tabular} & \textbf{67.99±6.85}, 0.0684 & \textbf{67.99±6.85}, 0.099 & 64.26±7.03, 0.0033 & 59.55±10.69, 0.0026 & 62.38±8.58, 0.0025 & \underline{66.55±7.25}, 0.0014 \\
\begin{tabular}[c]{@{}l@{}}pima\\  768,8\end{tabular} & \textbf{70.58±2.36}, 0.0476 & \textbf{70.58±2.36}, 0.0641 & 65.24±5.69, 0.0035 & 65.1±5.95, 0.0026 & 65.63±5.57, 0.0032 & \underline{69.66±4.88}, 0.0013 \\
\begin{tabular}[c]{@{}l@{}}monk1\\  556,6\end{tabular} & 51.79±3.06, 0.0164 & 52.15±3.1, 0.0697 & 51.97±3.91, 0.0015 & \underline{52.5±4.21}, 0.0015 & 51.96±4.42, 0.0009 & \textbf{53.04±5.14}, 0.0013 \\
\begin{tabular}[c]{@{}l@{}}monk3\\  556,6\end{tabular} & 50.72±1.55, 0.0225 & 50.9±1.42, 0.0476 & 51.44±2.22, 0.0016 & \underline{52.16±3.72}, 0.0015 & 51.8±2.81, 0.0007 & \textbf{53.06±5.3}, 0.0011 \\
\begin{tabular}[c]{@{}l@{}}checkerboard data\\  690,14\end{tabular} & \underline{82.17±2.44}, 0.0656 & \textbf{82.61±2.46}, 0.0944 & 76.67±2.53, 0.0033 & 73.62±4.98, 0.0028 & 78.26±5.05, 0.0043 & 81.01±1.07, 0.0014 \\
\begin{tabular}[c]{@{}l@{}}statlog australian credit\\  690,14\end{tabular} & 67.97±1.65, 0.0416 & 67.97±1.65, 0.0592 & 67.97±1.57, 0.0031 & \textbf{68.55±1.5}, 0.0025 & 67.97±1.94, 0.0033 & \underline{68.41±1.41}, 0.0013 \\
\begin{tabular}[c]{@{}l@{}}transfusion\\  748,4\end{tabular} & \underline{77.3±12.01}, 0.0323 & \underline{77.3±12.01}, 0.0897 & 76.51±14.55, 0.0027 & 76.24±14.98, 0.0015 & 76.51±14.55, 0.0007 & \textbf{78.64±10.99}, 0.0012 \\
\begin{tabular}[c]{@{}l@{}}vowel\\  988,10\end{tabular} & \underline{95.54±2.14}, 0.0604 & \underline{95.54±2.14}, 0.1765 & 94.43±2.42, 0.0026 & 93.01±5.2, 0.0017 & 93.63±0.97, 0.0054 & \textbf{95.95±3.23}, 0.0013 \\
\begin{tabular}[c]{@{}l@{}}yeast-0-2-5-6vs3-7-8-9\\  1004,8\end{tabular} & \textbf{93.22±2.26}, 0.0937 & \textbf{93.22±2.26}, 0.2803 & 90.73±3.25, 0.0025 & 90.54±3.12, 0.0016 & 90.14±2.65, 0.0044 & \underline{92.43±2.06}, 0.0012 \\
\begin{tabular}[c]{@{}l@{}}yeast-0-2-5-7-9vs3-6-8\\  1004,8\end{tabular} & \textbf{96.22±0.9}, 0.0932 & \textbf{96.22±0.9}, 0.2532 & 93.92±1.59, 0.0028 & 93.13±2.2, 0.0017 & 94.52±1.05, 0.0008 & \underline{95.72±0.44}, 0.0012 \\
\begin{tabular}[c]{@{}l@{}}yeast-0-3-5-9vs7-8\\  506,8\end{tabular} & \underline{91.7±2.68}, 0.0197 & \underline{91.7±2.68}, 0.1017 & 91.3±3.66, 0.0021 & 90.71±2.86, 0.0015 & 90.12±2.53, 0.0049 & \textbf{91.9±2.56}, 0.0011 \\
\begin{tabular}[c]{@{}l@{}}yeast-0-5-6-7-9vs4\\  528,8\end{tabular} & \underline{92.42±1.35}, 0.0216 & \underline{92.42±1.35}, 0.0554 & 90.72±1.81, 0.0028 & 90.34±1.82, 0.0015 & 90.72±1.58, 0.007 & \textbf{93.18±1.43}, 0.0012 \\
\begin{tabular}[c]{@{}l@{}}titanic\\  2201,3\end{tabular} & 77.1±15.93, 0.4282 & 77.33±16.02, 0.5734 & \underline{77.92±15.58}, 0.0029 & 77.33±16.02, 0.0027 & 77.87±13.46, 0.0029 & \textbf{79.05±15.04}, 0.0015 \\
\begin{tabular}[c]{@{}l@{}}abalone9-18\\  731,7\end{tabular} & \underline{95.36±3.31}, 0.0326 & \underline{95.36±3.31}, 0.1077 & 94.4±4.5, 0.0042 & 94.95±4.5, 0.0025 & 95.22±3.76, 0.0052 & \textbf{95.9±3.82}, 0.0014 \\
\begin{tabular}[c]{@{}l@{}}aus\\  690,14\end{tabular} & \underline{82.17±2.44}, 0.0285 & \textbf{82.61±2.46}, 0.0756 & 76.67±2.53, 0.0039 & 73.62±4.98, 0.0021 & 78.26±5.05, 0.0042 & 81.01±1.07, 0.0017 \\
\begin{tabular}[c]{@{}l@{}}cmc\\  1473,9\end{tabular} & 69.54±18.95, 0.6173 & 79.99±22.08, 0.5047 & \underline{81.62±20.33}, 0.0032 & \textbf{81.69±20.36}, 0.0017 & \underline{81.62±20.33}, 0.0043 & \underline{81.62±20.33}, 0.0014 \\
\begin{tabular}[c]{@{}l@{}}ripley\\  1250,2\end{tabular} & \underline{59.84±3.37}, 0.27 & \underline{59.84±3.37}, 0.2029 & \underline{59.84±3.37}, 0.0032 & \underline{59.84±3.37}, 0.0017 & \underline{59.84±3.37}, 0.0008 & \textbf{60.4±3.12}, 0.0013 \\
\begin{tabular}[c]{@{}l@{}}yeast5\\  1484,8\end{tabular} & \underline{97.57±2.03}, 0.4438 & \underline{97.57±2.03}, 0.7824 & 97.03±2.45, 0.0036 & 97.17±2.41, 0.0017 & 97.03±2.45, 0.0042 & \textbf{97.78±1.3}, 0.0014 \\
\begin{tabular}[c]{@{}l@{}}ozone\\  2536,72\end{tabular} & \underline{97.12±2.26}, 0.3991 & \underline{97.12±2.26}, 0.4385 & \underline{97.12±2.26}, 0.0043 & \underline{97.12±2.26}, 0.0021 & \underline{97.12±2.26}, 0.0032 & \textbf{97.2±2.11}, 0.002 \\
\begin{tabular}[c]{@{}l@{}}spambase\\  4601,57\end{tabular} & 99.3±1.31, 96.8396 & \textbf{99.39±1.36}, 1.6873 & 77.4±13.77, 0.0033 & \textbf{99.39±1.36}, 0.0019 & 77.75±13.38, 0.0038 & \underline{99.39±3.86}, 0.0016 \\
 \hline
Avg. Acc. $\pm$ Avg. Std. & 83.16$\pm$7.04 & \underline{84.26$\pm$6.44} & 82.53$\pm$7.39 & 82.18$\pm$7.91 & 83.21$\pm$6.6& \textbf{86.35$\pm$5.06} \\ \hline
Avg. time     & 0.1304    & 0.1909     & 0.0031     & \underline{0.0019} & 0.0037    & \textbf{0.0012}     \\ \hline
Avg. rank     & 3.35     & \underline{2.96}     & 3.96     & 4.45 & 4.12    & \textbf{2.16}     \\ \hline
\multicolumn{7}{l}{Here, Avg., Acc. and Std. are acronyms used for average, accuracy, and standard deviation, respectively.}\\
 \multicolumn{7}{l}{$^{\dagger}$ signifies the proposed model while boldface and underline signify the best and second-best models, respectively. }
\end{tabular}%
}
\end{table*}
\end{tiny}

\begin{table*}[]
\caption{The classification accuracies and training times of the proposed $\mathcal{L}_{RoBoSS}$-SVM and baseline models on the EEG dataset.}
\centering
\label{tab:EEG-table}
\resizebox{17cm}{!}{%
\begin{tabular}{lcccccc}
\hline
Model & $\mathcal{L}_{hinge}$-SVM \cite{cortes1995support}             & $\mathcal{L}_{pin}$-SVM \cite{huang2013support}           & $\mathcal{L}_{LINEX}$-SVM \cite{ma2019linex}         & $\mathcal{L}_{qtse}$-SVM \cite{zhao2022asymmetric} & $\mathcal{L}_{wave}$-SVM \cite{akhtar2024advancing} & $\mathcal{L}_{RoBoSS}$-SVM$^{\dagger}$ \\ \hline
Dataset & Acc. $\pm$ Std., time & Acc. $\pm$ Std., time & Acc. $\pm$ Std., time & Acc. $\pm$ Std., time & Acc. $\pm$ Std., time & Acc. $\pm$ Std., time \\ \hline
EEG\_B\_vs\_S\_bhattacharyya\_100 & 69.5±5.97, 0.0049 & 70.5±8.55, 0.0051 & 73.5±7.83, 0.0015 & 55±1.77, 0.0017 & \underline{74±10.4}, 0.0025 & \textbf{77±6.47}, 0.0016 \\
EEG\_B\_vs\_S\_entropy\_100 & 69.5±5.97, 0.0055 & 70.5±8.55, 0.004 & 73.5±7.83, 0.0018 & 55±1.77, 0.0018 & \underline{74±10.4}, 0.004 & \textbf{77±6.47}, 0.0017 \\
EEG\_B\_vs\_S\_roc\_50 & 72.5±3.54, 0.0051 & 73.5±4.87, 0.003 & \underline{74.5±5.7}, 0.0016 & 58±14.3, 0.0019 & 72±5.42, 0.0028 & \textbf{79±9.45}, 0.0015 \\
EEG\_B\_vs\_S\_ttest\_200 & \underline{83.5±5.18}, 0.0059 & \textbf{83.5±4.18}, 0.0038 & 77±6.22, 0.0016 & 55±1.77, 0.0017 & 76±6.98, 0.0039 & 81±7.42, 0.0018 \\
EEG\_B\_vs\_S\_wilcoxon\_100 & 76±2.24, 0.0048 & 76±2.24, 0.0034 & \underline{79.5±4.11}, 0.0019 & 55±1.77, 0.0017 & 76.5±5.18, 0.004 & \textbf{81.5±6.27}, 0.0013 \\
EEG\_B\_vs\_S\_wilcoxon\_200 & \underline{81.5±8.4}, 0.0052 & \underline{81.5±8.4}, 0.0035 & 80±7.07, 0.0018 & 55±1.77, 0.0019 & 79.5±8.18, 0.004 & \textbf{82.5±5.86}, 0.0013 \\
EEG\_A\_vs\_S\_bhattacharyya\_100 & \underline{71.5±4.18}, 0.0057 & \underline{71.5±4.18}, 0.0038 & 68±4.47, 0.0018 & 56±1.37, 0.0018 & 71±6.02, 0.0039 & \textbf{76±8.59}, 0.0013 \\
EEG\_A\_vs\_S\_bhattacharyya\_200 & \underline{78±7.58}, 0.0049 & \underline{78±7.58}, 0.004 & 70±8.29, 0.0016 & 55±1.77, 0.0018 & 70±8.29, 0.0039 & \textbf{79.5±4.11}, 0.0013 \\
EEG\_A\_vs\_S\_entropy\_100 & \underline{71.5±4.18}, 0.0052 & \underline{71.5±4.18}, 0.0037 & 68±4.47, 0.0018 & 56±1.37, 0.0017 & 71±6.02, 0.004 & \textbf{76±8.59}, 0.0012 \\
EEG\_A\_vs\_S\_entropy\_200 & \underline{78±7.58}, 0.0049 & \underline{78±7.58}, 0.004 & 70±8.29, 0.0019 & 55±1.77, 0.0018 & 70±8.29, 0.0026 & \textbf{79.5±4.11}, 0.0013 \\
EEG\_A\_vs\_S\_ttest\_100 & 79±6.75, 0.0051 & \underline{80±6.37}, 0.0034 & 74.5±9.75, 0.0017 & 55±1.77, 0.0018 & 76±4.54, 0.0025 & \textbf{81±5.76}, 0.0014 \\
EEG\_C\_vs\_B\_roc\_200 & 77±4.11, 0.005 & 77±4.11, 0.0036 & 77.5±7.29, 0.0017 & 55±1.77, 0.0018 & \underline{82±4.47}, 0.0025 & \textbf{84.5±3.26}, 0.0013 \\
EEG\_C\_vs\_B\_wilcoxon\_200 & 84.5±4.11, 0.0047 & 86±5.18, 0.0031 & 86±8.4, 0.0015 & 55±1.77, 0.0017 & \underline{88±6.47}, 0.0026 & \textbf{90±5}, 0.0013 \\
EEG\_C\_vs\_A\_entropy\_200 & \underline{77±7.58}, 0.0056 & \underline{77±7.58}, 0.004 & 74±10.55, 0.0016 & 55±1.77, 0.0017 & 76±5.18, 0.0026 & \textbf{80±3.06}, 0.0012 \\
EEG\_C\_vs\_A\_entropy\_50 & 67.5±9.68, 0.0053 & 67.5±9.68, 0.0034 & \underline{69±6.75}, 0.0017 & 55±1.77, 0.0019 & 65.5±11.65, 0.0024 & \textbf{72±12.67}, 0.0012 \\
EEG\_C\_vs\_A\_roc\_150 & 77±5.42, 0.0051 & 77.5±3.95, 0.0039 & \underline{79.5±4.81}, 0.0016 & 55±1.77, 0.0018 & \underline{79.5±4.81}, 0.0028 & \textbf{85±7.29}, 0.0012 \\
EEG\_C\_vs\_A\_roc\_50 & 71.5±3.35, 0.0049 & 71.5±3.35, 0.0043 & \underline{73.5±8.77}, 0.0018 & 55±1.77, 0.0018 & 71.5±6.98, 0.0025 & \textbf{77±5.42}, 0.0015 \\
EEG\_C\_vs\_A\_ttest\_100 & 76.5±7.83, 0.0051 & 76.5±7.83, 0.0036 & \underline{77±4.81}, 0.0016 & 55±1.77, 0.0021 & 76.5±7.2, 0.0025 & \textbf{82±7.79}, 0.0012 \\
EEG\_C\_vs\_A\_ttest\_150 & 76.5±6.75, 0.0056 & 76.5±5.76, 0.0033 & \underline{78.5±4.54}, 0.0017 & 55±1.77, 0.0022 & \underline{78.5±4.54}, 0.0025 & \textbf{84±2.85}, 0.0013 \\
EEG\_C\_vs\_A\_ttest\_200 & \underline{79.5±5.97}, 0.005 & \underline{79.5±5.97}, 0.0036 & 79±3.79, 0.0016 & 55±1.77, 0.0022 & 79±5.76, 0.0025 & \textbf{85±5.59}, 0.0013 \\
EEG\_C\_vs\_A\_ttest\_50 & 74.5±7.79, 0.0051 & 74.5±7.79, 0.0039 & 75±9.01, 0.0017 & 55±9.84, 0.0021 & \underline{76±4.54}, 0.0024 & \textbf{79.5±5.42}, 0.0022 \\
EEG\_C\_vs\_A\_wilcoxon\_50 & 78.5±5.18, 0.0049 & 78.5±4.87, 0.0032 & \underline{79.5±6.94}, 0.0018 & 56±11.81, 0.002 & 79.5±7.79, 0.0029 & \textbf{84±6.98}, 0.0016 \\
EEG\_C\_vs\_S\_bhattacharyya\_100 & 60.5±6.94, 0.005 & 60.5±6.94, 0.0035 & 63.5±9.12, 0.0017 & 55±1.77, 0.0021 & \underline{64.5±6.94}, 0.0047 & \textbf{72±11.1}, 0.0017 \\
EEG\_C\_vs\_S\_bhattacharyya\_150 & 66.5±7.83, 0.0051 & 66.5±7.83, 0.0033 & 68.5±9.78, 0.0016 & 55±1.77, 0.0026 & \underline{69±9.12}, 0.0028 & \textbf{72±2.74}, 0.0012 \\
EEG\_C\_vs\_S\_entropy\_100 & 60.5±6.94, 0.0048 & 60.5±6.94, 0.0038 & 63.5±9.12, 0.0016 & 55±1.77, 0.0024 & \underline{64.5±6.94}, 0.0025 & \textbf{72±11.1}, 0.0012 \\
EEG\_C\_vs\_S\_entropy\_150 & 66.5±7.83, 0.0052 & 66.5±7.83, 0.0037 & 68.5±9.78, 0.0017 & 55±1.77, 0.0024 & \underline{69±9.12}, 0.0025 & \textbf{72±2.74}, 0.0013 \\
EEG\_C\_vs\_S\_ttest\_150 & 68±11.24, 0.0049 & 68±11.24, 0.0035 & 71±6.52, 0.0017 & 55±1.77, 0.0022 & \underline{75.5±9.42}, 0.0034 & \textbf{76.5±7.62}, 0.0012 \\
EEG\_C\_vs\_S\_ttest\_200 & 69.5±8.18, 0.0047 & 69.5±8.18, 0.0036 & \underline{74.5±5.97}, 0.0015 & 55±1.77, 0.0021 & \underline{74.5±5.97}, 0.0038 & \textbf{76.5±6.02}, 0.0012 \\
EEG\_C\_vs\_S\_wilcoxon\_100 & 64.5±8.91, 0.0048 & 64.5±7.79, 0.0031 & \underline{68±10.37}, 0.0017 & 55±1.77, 0.0021 & 67.5±10.31, 0.0025 & \textbf{73.5±5.48}, 0.0013 \\
EEG\_O\_vs\_B\_roc\_150 & 82±2.74, 0.005 & 82±2.74, 0.0037 & 79±1.37, 0.0016 & 55±1.77, 0.0023 & \underline{84±3.79}, 0.0031 & \textbf{86.5±3.79}, 0.0013 \\
EEG\_O\_vs\_B\_ttest\_50 & 71±4.87, 0.0051 & 73±6.22, 0.0029 & \underline{77.5±7.91}, 0.0019 & 55±1.77, 0.0021 & 77±8.91, 0.0025 & \textbf{81±8.02}, 0.0014 \\
EEG\_O\_vs\_B\_wilcoxon\_150 & 82±2.74, 0.0047 & 82±2.74, 0.0037 & 79±1.37, 0.0017 & 55±1.77, 0.0022 & \underline{84±3.79}, 0.003 & \textbf{86.5±3.79}, 0.0013 \\ \hline
Avg Acc. $\pm$ Avg. Std.  & 73.8±6.17 & 74.05±6.29 & 74.06±6.91 & 55.19±2.7 & \underline{74.73±6.98} & \textbf{79.42±6.28} \\ \hline
Avg. time & 0.0051 & 0.0036 & \underline{0.0017} & 0.002 & 0.003 & \textbf{0.0014 }\\ \hline
\multicolumn{7}{l}{Here, Avg., Acc. and Std. are acronyms used for average, accuracy, and standard deviation, respectively.}\\
 \multicolumn{7}{l}{$^{\dagger}$ signifies the proposed model while boldface and underline signify the best and second-best models, respectively. }
\end{tabular}%
}
\end{table*}

\begin{table*}[]
\caption{The classification accuracies and training times of the proposed $\mathcal{L}_{RoBoSS}$-SVM and baseline models on the BreaKHis dataset.}
\centering
\label{tab:breast cancer table}
\resizebox{17cm}{!}{%
\begin{tabular}{lcccccc}
\hline
Model & $\mathcal{L}_{hinge}$-SVM \cite{cortes1995support}             & $\mathcal{L}_{pin}$-SVM \cite{huang2013support}           & $\mathcal{L}_{LINEX}$-SVM \cite{ma2019linex}         & $\mathcal{L}_{qtse}$-SVM \cite{zhao2022asymmetric} & $\mathcal{L}_{wave}$-SVM \cite{akhtar2024advancing} & $\mathcal{L}_{RoBoSS}$-SVM$^{\dagger}$ \\ \hline
Dataset & Acc. $\pm$ Std., time & Acc. $\pm$ Std., time & Acc. $\pm$ Std., time & Acc. $\pm$ Std., time & Acc. $\pm$ Std., time & Acc. $\pm$ Std., time \\ \hline
ADvsDC & \underline{66.87±4.75}, 0.005 & \underline{66.87±4.75}, 0.0169 & 66.24±4.16, 0.0029 & 66.24±4.16, 0.0011 & 66.24±4.16, 0.0055 & \textbf{66.88±4.08}, 0.0021 \\
ADvsLC & 57.21±6.49, 0.0033 & 57.21±6.49, 0.0216 & 58.89±7.25, 0.0036 & 57.61±5.18, 0.0011 & \underline{58.89±5.59}, 0.0041 & \textbf{62.59±5.98}, 0.0017 \\
ADvsMC & 61.82±7.27, 0.0045 & \underline{62.18±6.97}, 0.0147 & 61.45±6.35, 0.0038 & 61.45±6.35, 0.0011 & 61.45±6.35, 0.0063 & \textbf{63.27±4.15}, 0.0017 \\
ADvsPC & 56.54±6.82, 0.0045 & \underline{57.76±4.97}, 0.0174 & \underline{57.76±4.97}, 0.0031 & 57.36±4.69, 0.0011 & \underline{57.76±4.97}, 0.0042 & \textbf{61.91±5.68}, 0.0021 \\
FDvsDC & 53.26±2.04, 0.0104 & 53.26±2.04, 0.0318 & \underline{56.63±2.33}, 0.0035 & 53.26±2.04, 0.0013 & 56.18±2.1, 0.0054 & \textbf{60±4.47}, 0.0018 \\
FDvsLC & \textbf{66.84±3.76}, 0.0077 & \textbf{66.84±3.76}, 0.0128 & 66.3±5.85, 0.0037 & 63.37±3.51, 0.0012 & 65.5±6.06, 0.0043 & \underline{66.57±5.02}, 0.0016 \\
FDvsMC & 58.62±2.23, 0.0082 & 58.62±2.23, 0.0553 & \underline{60.6±3.73}, 0.0036 & 58.37±2.21, 0.0012 & 59.61±3.91, 0.0042 & \textbf{62.56±4.74}, 0.0017 \\
FDvsPC & 63.2±3.96, 0.0071 & 63.2±3.96, 0.0281 & \underline{64±3.89}, 0.0038 & 63.2±3.96, 0.0011 & 63.73±6.56, 0.0037 & \textbf{66.93±5.45}, 0.0016 \\
PTvsDC & \underline{64.73±5.35}, 0.0053 & \underline{64.73±5.35}, 0.017 & 64.41±4.85, 0.0034 & 64.41±4.85, 0.0012 & 64.41±4.85, 0.0037 & \textbf{66.26±2.45}, 0.0015 \\
PTvsLC & 55.11±9.65, 0.0046 & 56.72±6.05, 0.0105 & 59.11±4.6, 0.0034 & 58.31±4.94, 0.0011 & \underline{59.13±4.05}, 0.0075 & \textbf{63.05±7.25}, 0.0014 \\
PTvsMC & \underline{59.5±6.39}, 0.0045 & \underline{59.5±6.39}, 0.0234 & \underline{59.5±6.39}, 0.0036 & \underline{59.5±6.39}, 0.0011 & \underline{59.5±6.39}, 0.0036 & \textbf{62.31±4.44}, 0.0014 \\
PTvsPC & 54.97±8.45, 0.0047 & 57.32±4.51, 0.0837 & 57.32±4.51, 0.0033 & \underline{57.71±4.75}, 0.0011 & 57.32±4.51, 0.0044 & \textbf{60.48±9.87}, 0.0013 \\
TAvsDC & \textbf{64.21±6.99}, 0.0065 & \textbf{64.21±6.99}, 0.0473 & 61.55±5.54, 0.0036 & 61.55±5.54, 0.0011 & 61.55±5.54, 0.0043 & \underline{63.64±5.92}, 0.0016 \\
TAvsLC & 51.28±8.39, 0.0045 & 56.56±4.32, 0.0374 & 59.94±7.93, 0.0033 & 56.56±4.32, 0.0012 & \underline{61.06±6.47}, 0.0038 & \textbf{64.05±1.49}, 0.0014 \\
TAvsMC & \underline{59.85±6.71}, 0.004 & \textbf{60.19±6.4}, 0.0167 & 56.51±6.01, 0.0035 & 56.51±6.01, 0.0011 & 56.51±6.01, 0.0051 & 59.2±5.95, 0.0017 \\
TAvsPC & 54.47±6.54, 0.0053 & 56.35±4.15, 0.0107 & 56.35±4.15, 0.0078 & \underline{58.23±4.76}, 0.0011 & 56.35±4.15, 0.01 & \textbf{62.3±3.51}, 0.0018 \\ \hline
Avg. Acc. $\pm$ Avg. Std. & 59.28±5.99 & 60.09±4.96 & \underline{60.41±5.16} & 59.6±4.6 & 60.32±4.15 & \textbf{63.25±5.03} \\ \hline
Avg. time & 0.0056 & 0.0278 & 0.0037 & \textbf{0.0011} & 0.005 & \underline{0.0016}
\\ \hline
\multicolumn{7}{l}{Here, Avg., Acc. and Std. are acronyms used for average, accuracy, and standard deviation, respectively.}\\
 \multicolumn{7}{l}{$^{\dagger}$ signifies the proposed model while boldface and underline signify the best and second-best models, respectively. }
\end{tabular}%
}
\end{table*}

\begin{tiny}
\begin{table*}[]
 \centering
 \caption{\vspace{0.1mm}The optimal parameters corresponding to the accuracy values of the proposed $\mathcal{L}_{RoBoSS}$-SVM and baseline models across each of the 79 D1 category UCI and KEEL datasets.
}
 \resizebox{15cm}{!}{
\label{tab:UCI-KEEL-optimal-Parameter-table}
\begin{tabular}{lcccccc}
\hline
Dataset\textbackslash{}Model & $\mathcal{L}_{hinge}$-SVM \cite{cortes1995support}             & $\mathcal{L}_{pin}$-SVM \cite{huang2013support}           & $\mathcal{L}_{LINEX}$-SVM \cite{ma2019linex}         & $\mathcal{L}_{qtse}$-SVM \cite{zhao2022asymmetric} & $\mathcal{L}_{wave}$-SVM \cite{akhtar2024advancing} & $\mathcal{L}_{RoBoSS}$-SVM$^{\dagger}$ \\
 & (C, $\sigma$) & (C, $\sigma$, $\tau$) & (a, C, $\sigma$) & (a, C, $\sigma$) & (a, C, $\sigma$) & (a, $\lambda$, C, $\sigma$) \\ \hline
acute\_inflammation & 0.1, 1 & 0.1, 1, 0 & -4, 0.00001, 10 & -1, 1000, 10 & -5, 10, 10 & 1, 1.3, 0.01, 1 \\
balloons & 1, 1 & 100000, 10, 0.5 & -5, 0.001, 10 & -44, 1000, 10 & -4, 0.01, 10 & 1.5, 1.1, 10, 10 \\
fertility & 0.000001, 0.000001 & 10, 1, 0.5 & -7, 0.000001, 0.0001 & -1, 0.00001, 0.000001 & -5, 0.000001, 0.000001 & 1.8, 1.6, 0.001, 1 \\
molec\_biol\_promoter & 10, 0.1 & 1, 10, 0.3 & -1, 0.001, 0.1 & -2, 0.00001, 0.00001 & -5, 0.000001, 1 & 1.1, 0.6, 0.001, 0.1 \\
parkinsons & 10, 1 & 0.1, 10000, 0.3 & -6, 0.000001, 0.001 & -2, 10000, 0.00001 & 3, 0.000001, 1 & 3.3, 0.8, 0.01, 0.1 \\
pittsburg\_bridges\_T\_OR\_D & 0.000001, 0.000001 & 0.000001, 0.000001, 0 & -7, 0.000001, 0.001 & -1, 0.00001, 0.000001 & -5, 0.000001, 0.000001 & 1.7, 1.1, 0.001, 1 \\
breast\_cancer & 0.000001, 0.000001 & 0.000001, 0.000001, 0 & -6, 0.000001, 0.0001 & -4, 0.000001, 1 & 2, 100, 10 & 2.7, 1.9, 1000, 10 \\
breast\_cancer\_wisc\_prog & 0.000001, 0.000001 & 0.000001, 0.000001, 0 & -5, 0.00001, 0.1 & -1, 0.00001, 0.000001 & -5, 0.000001, 0.000001 & 1.6, 0.5, 0.01, 1 \\
congressional\_voting & 10000, 0.1 & 10000, 0.1, 0.7 & -6, 0.000001, 0.1 & -1, 1000, 1 & -5, 0.000001, 0.000001 & 2.4, 0.7, 0.01, 1 \\
echocardiogram & 10, 0.1 & 1, 0.1, 0.5 & -2, 0.0001, 1 & -1, 10000, 10 & 2, 10000, 10 & 2.9, 0.8, 0.001, 1 \\
haberman\_survival & 0.000001, 0.000001 & 0.000001, 0.000001, 0 & -4, 0.000001, 1 & -1, 10, 1 & -5, 0.000001, 0.000001 & 0.5, 1.9, 0.0001, 1 \\
hepatitis & 1, 0.1 & 1, 0.1, 0.9 & -1, 0.0001, 1 & -5, 0.000001, 1 & 3, 0.0001, 1 & 1.9, 0.4, 0.00001, 1 \\
horse\_colic & 0.1, 0.1 & 0.1, 0.1, 0 & -3, 0.0001, 1 & -1, 0.00001, 1 & 4, 0.000001, 1 & 3.9, 1.1, 0.000001, 1 \\
ionosphere & 1, 100 & 1, 1, 0.3 & -1, 0.001, 1 & -3, 0.00001, 1 & -5, 0.000001, 0.000001 & 4.7, 0.7, 10, 0.1 \\
planning & 0.000001, 0.000001 & 0.000001, 0.000001, 0 & -7, 0.000001, 0.01 & -1, 0.00001, 0.00001 & 5, 0.00001, 1 & 1.9, 1.3, 1, 1 \\
spect & 1, 0.1 & 1000000, 0.1, 0.3 & -2, 0.00001, 0.1 & -1, 0.00001, 0.00001 & -5, 0.000001, 0.000001 & 3.2, 1.1, 1, 0.1 \\
spectf & 0.000001, 0.000001 & 0.000001, 0.000001, 0 & -10, 0.000001, 0.000001 & -50, 0.000001, 0.000001 & 3, 0.00001, 1 & 4.1, 1.2, 0.001, 1 \\
statlog\_heart & 1, 0.1 & 1, 0.1, 0 & -3, 0.00001, 1 & -1, 1000000, 10 & -5, 0.000001, 0.000001 & 4.2, 1.4, 0.0001, 1 \\
bupa or liver-disorders & 1, 1 & 1, 1, 0 & -1, 0.00001, 10 & -1, 0.01, 10 & -5, 100000, 10 & 4, 1.8, 0.00001, 1 \\
cleve & 1, 0.1 & 1, 0.1, 0 & -5, 0.00001, 1 & -1, 10000, 10 & 2, 10000, 10 & 1.7, 1.2, 0.000001, 10 \\
crossplane130 & 1000, 1000000 & 1000, 1000000, 0 & -2, 0.000001, 100 & -2, 0.000001, 10000 & -5, 100000, 10 & 0.4, 1.2, 10, 1 \\
crossplane150 & 0.1, 1 & 0.01, 1000, 0.3 & -2, 0.000001, 1000 & -1, 0.001, 10 & 2, 1000000, 1000 & 1.5, 1.7, 0.01, 10 \\
ecoli-0-1-4-6\_vs\_5 & 1, 1 & 1, 1, 0 & -2, 0.00001, 1 & -1, 1000000, 10 & 0, 100000, 100 & 4, 1.6, 0.001, 10 \\
ecoli-0-1-4-7\_vs\_2-3-5-6 & 1, 1 & 1, 1, 0 & -4, 0.00001, 1 & -1, 1, 10 & 4, 100000, 10 & 0, 1.4, 100000, 1 \\
ecoli-0-1-4-7\_vs\_5-6 & 1, 1 & 1, 1, 0 & -1, 0.0001, 1 & -1, 0.000001, 10 & 4, 1000000, 10 & 2.3, 1.8, 1, 1 \\
ecoli-0-1\_vs\_2-3-5 & 1, 1 & 1, 1, 0 & -1, 0.0001, 10 & -1, 1000000, 10 & 5, 100000, 10 & 1.8, 0.3, 0.1, 10 \\
ecoli-0-1\_vs\_5 & 10, 1 & 10, 1, 0 & -1, 0.000001, 1 & -1, 100000, 10 & 4, 10, 10 & 3, 1.3, 0.1, 10 \\
ecoli-0-2-3-4\_vs\_5 & 1, 1 & 1, 1, 0 & -1, 0.000001, 10 & -1, 0.01, 10 & 4, 10000, 10 & 3.1, 1.9, 0.00001, 10 \\
ecoli-0-2-6-7\_vs\_3-5 & 1, 1 & 1, 1, 0 & -4, 0.000001, 10 & -1, 1000, 10 & 5, 10, 10 & 2.6, 1.8, 0.01, 10 \\
ecoli-0-3-4-6\_vs\_5 & 1, 1 & 1, 1, 0 & -3, 0.000001, 10 & -1, 0.001, 10 & 4, 100, 10 & 3.9, 0.4, 0.01, 10 \\
ecoli-0-3-4-7\_vs\_5-6 & 1, 1 & 1, 1, 0 & -1, 0.0001, 10 & -1, 1000000, 10 & 3, 100000, 10 & 0.4, 0.7, 0.1, 10 \\
ecoli-0-4-6\_vs\_5 & 10, 1 & 10, 1, 0 & -1, 0.000001, 10 & -1, 1, 10 & 5, 100000, 10 & 2.5, 1.7, 0.000001, 10 \\
ecoli-0-6-7\_vs\_3-5 & 1, 1 & 1, 1, 0 & -1, 0.000001, 10 & -1, 100000, 10 & 5, 10000, 10 & 0.7, 0.6, 1, 1 \\
ecoli-0-6-7\_vs\_5 & 1, 1 & 1, 1, 0 & -1, 0.000001, 10 & -1, 0.01, 10 & 5, 10000, 10 & 4.7, 2, 0.0001, 10 \\
ecoli0137vs26 & 1, 1 & 0.1, 1, 0.9 & -2, 0.000001, 1 & -1, 1, 10 & 3, 1000000, 10 & 2.2, 0.6, 0.1, 10 \\
ecoli01vs5 & 1, 1 & 1, 1, 0 & -6, 0.000001, 1 & -1, 0.000001, 10 & 3, 1000, 10 & 1.9, 0.4, 0.01, 1 \\
ecoli3 & 1, 1 & 1, 1, 0 & -2, 0.0001, 1 & -1, 0.000001, 10 & 4, 100000, 10 & 0, 1.6, 1, 1 \\
ecoli4 & 1, 1 & 1, 1, 0 & -2, 0.00001, 10 & -2, 0.000001, 10 & 3, 10000, 10 & 0.9, 0.6, 0.1, 1 \\
glass2 & 0.000001, 0.000001 & 0.000001, 0.000001, 0 & -7, 0.000001, 0.000001 & -1, 0.00001, 0.000001 & 5, 1000, 10 & 5, 1.3, 0.001, 1 \\
glass4 & 100, 100 & 100, 100, 0 & -2, 0.00001, 100 & -1, 10000, 100 & 5, 10, 0100 & 0.6, 0.8, 0.01, 100 \\
glass5 & 10, 100 & 10, 100, 0 & -7, 0.000001, 0.000001 & -1, 100, 100 & 5, 0.000001, 100 & 1.9, 0.5, 0.001, 100 \\
haber & 10000, 1 & 10000, 1, 0 & -1, 0.00001, 10 & -1, 0.00001, 0.000001 & -3, 0.0001, 100 & 1.6, 1.9, 0.00001, 100 \\
haberman & 10000, 1 & 10000, 1, 0 & -1, 0.001, 10 & -50, 0.000001, 0.000001 & -5, 0.1, 10 & 0.7, 1, 0.0001, 10 \\
iono & 10, 1 & 10, 1, 0 & -1, 0.0001, 1 & -1, 0.0001, 1 & 4, 0.001, 10 & 0.7, 1, 0.0001, 10 \\
led7digit-0-2-4-5-6-7-8-9\_vs\_1 & 1, 1 & 1, 1, 0 & -6, 0.000001, 1 & -1, 10, 10 & 3, 0.00001, 1 & 3.6, 0.2, 0.000001, 1 \\
new-thyroid1 & 10, 1 & 10, 1, 0 & -1, 0.0001, 100 & -1, 0.01, 100 & 5, 10000, 10 & 3.4, 1.3, 0.0001, 10 \\
shuttle-6\_vs\_2-3 & 1, 1 & 1, 1, 0 & -5, 0.000001, 1 & -2, 0.000001, 10 & -5, 100, 100 & 0.7, 1.1, 1, 100 \\
votes & 0.1, 0.1 & 1000000, 0.1, 0.5 & -2, 0.000001, 1 & -1, 10000, 10 & 5, 100, 10 & 0, 0.1, 0.00001, 1 \\
wpbc & 0.000001, 0.000001 & 0.000001, 0.000001, 0 & -7, 0.000001, 0.000001 & -1, 0.00001, 0.000001 & -4, 1000000, 10 & 5, 0.5, 0.1, 1 \\
yeast1vs7 & 10, 1 & 10, 1, 0 & -7, 0.000001, 0.000001 & -1, 0.00001, 0.000001 & -5, 0.000001, 0.000001 & 4, 1.9, 0.00001, 1 \\
yeast2vs8 & 1, 1 & 1, 1, 0 & -3, 0.00001, 1 & -1, 100000, 10 & -5, 0.000001, 0.000001 & 4.4, 0.5, 0.1, 10 \\
bank & 1, 1 & 1, 1, 0.9 & -7, 0.000001, 0.000001 & -1, 0.00001, 0.000001 & 3, 0.001, 10 & 0, 1.9, 1000000, 1 \\
blood & 10000, 1 & 10000, 1, 0 & -7, 0.000001, 0.000001 & -1, 0.00001, 0.000001 & -5, 0.000001, 0.000001 & 0.3, 0.5, 100, 0.1 \\
breast\_cancer\_wisc\_diag & 0.1, 0.1 & 0.1, 0.1, 0.7 & -2, 0.00001, 1 & -1, 0.000001, 10 & -5, 0.000001, 0.000001 & 0.9, 0.3, 0.00001, 1 \\
chess\_krvkp & 1000, 100 & 0.01, 10, 0.3 & -4, 0.0001, 1 & -2, 0.000001, 0.001 & -4, 0.00001, 1 & 4, 0.4, 10, 1 \\
credit\_approval & 10, 0.1 & 10, 0.1, 0 & -5, 0.00001, 1 & -2, 0.00001, 0.000001 & -2, 0.00001, 1 & 1.7, 0.1, 0.000001, 0.1 \\
cylinder\_bands & 0.000001, 0.000001 & 0.1, 0.1, 0.7 & -4, 0.0001, 1 & -2, 10, 0.001 & 5, 0.0001, 1 & 1.4, 1.8, 1, 1 \\
ilpd\_indian\_liver & 0.000001, 0.000001 & 0.000001, 0.000001, 0 & -6, 0.000001, 0.000001 & -1, 0.00001, 0.000001 & 1, 0.000001, 1 & 4.9, 1, 0.000001, 0.1 \\
mammographic & 1, 1 & 1, 1, 0 & -1, 0.000001, 1 & -1, 10000, 10 & 5, 1000, 10 & 4.9, 1.5, 0.000001, 1 \\
oocytes\_trisopterus\_nucleus\_2f & 1, 1 & 1, 1, 0 & -3, 0.00001, 1 & -4, 0.000001, 1 & 2, 0.00001, 1 & 1.6, 0.9, 0.000001, 1 \\
pima & 1, 1 & 1, 1, 0 & -7, 0.000001, 1 & -1, 0.00001, 0.000001 & 3, 1, 10 & 2.9, 0.6, 0.01, 1 \\
monk1 & 1, 0.1 & 1000, 0.1, 0.3 & -5, 0.00001, 10 & -18, 1000, 10 & 0, 0.00001, 10 & 2.9, 0.4, 100, 10 \\
monk3 & 0.1, 1 & 1, 0.1, 0.5 & -2, 1, 10 & -48, 0.000001, 1 & -2, 0.000001, 10 & 3.7, 0.3, 10000, 10 \\
checkerboard\_Data & 1, 1 & 1, 1, 0.5 & -6, 0.000001, 1 & -3, 0.000001, 1 & 5, 0.00001, 1 & 1.7, 1.2, 0.000001, 1 \\
statlog\_australian\_credit & 1000, 10 & 1000, 10, 0 & -8, 0.000001, 1 & -27, 0.01, 1 & 1, 100, 1 & 1.7, 1.1, 0.001, 0.01 \\
transfusion & 0.1, 100 & 0.1, 100, 0 & -2, 0.000001, 100 & -2, 0.00001, 0.1 & 3, 0.000001, 100 & 2.9, 0.1, 1, 100 \\
vowel & 10, 0.1 & 10, 0.1, 0 & -2, 0.00001, 1 & -1, 10, 10 & 5, 100000, 10 & 1.7, 1.6, 0.001, 10 \\
yeast-0-2-5-6\_vs\_3-7-8-9 & 10, 1 & 10, 1, 0 & -1, 0.000001, 1 & -1, 100, 10 & -5, 0.000001, 0.000001 & 0.9, 0.8, 1, 10 \\
yeast-0-2-5-7-9\_vs\_3-6-8 & 10, 1 & 10, 1, 0 & -1, 0.00001, 10 & -1, 1000000, 10 & 5, 0.000001, 10 & 3, 0.3, 0.001, 10 \\
yeast-0-3-5-9\_vs\_7-8 & 10, 0.1 & 10, 0.1, 0 & -2, 0.00001, 1 & -4, 0.000001, 1 & -5, 0.000001, 0.000001 & 2.1, 1, 0.00001, 1 \\
yeast-0-5-6-7-9\_vs\_4 & 1, 1 & 1, 1, 0 & -2, 0.0001, 1 & -1, 0.00001, 0.000001 & 3, 10000, 10 & 2.2, 1.8, 0.00001, 10 \\
titanic & 0.01, 1 & 0.1, 1, 0.3 & -1, 0.0001, 1 & -1, 1000000, 10 & 5, 1000, 10 & 0, 1.4, 0.001, 1 \\
abalone9-18 & 10, 1 & 10, 1, 0 & -2, 0.000001, 10 & -1, 1000000, 10 & 4, 1000, 10 & 2.7, 1.7, 0.000001, 10 \\
aus & 1, 1 & 1, 1, 0.5 & -6, 0.000001, 1 & -3, 0.000001, 1 & 5, 0.00001, 1 & 1.7, 1.2, 0.000001, 1 \\
cmc & 10000, 1 & 10, 10, 0.3 & -10, 10, 1 & -1, 0.00001, 1 & 2, 0.00001, 0.1 & 0.2, 1.3, 10000, 1 \\
ripley & 0.001, 10 & 0.001, 10, 0 & -4, 0.000001, 10 & -1, 0.001, 1000 & -5, 0.000001, 100 & 1.6, 1.3, 10, 1000000 \\
yeast5 & 10, 10 & 10, 10, 0 & -7, 0.000001, 0.000001 & -1, 0.01, 10 & -5, 0.000001, 0.000001 & 1.4, 0.1, 0.00001, 10 \\
ozone & 0.000001, 0.000001 & 0.000001, 0.000001, 0 & -7, 0.000001, 0.000001 & -1, 0.00001, 0.000001 & -5, 0.000001, 0.000001 & 3.6, 1.9, 10, 0.1 \\
spambase & 100000, 1000 & 0.01, 10000, 0.7 & -2, 0.000001, 1 & -2, 0.000001, 0.000001 & 3, 0.000001, 1 & 3.4, 1, 0.001, 1 \\ \hline
\end{tabular}%
}
\end{table*}
\end{tiny}

\begin{table*}[]
\caption{The optimal parameters corresponding to the accuracy values of the proposed $\mathcal{L}_{RoBoSS}$-SVM and baseline models across each of the 32 EEG datasets.}
\centering
\label{tab:EEG-optimal-parameters}
\resizebox{17cm}{!}{%
\begin{tabular}{lcccccc}
\hline
Dataset\textbackslash{}Model & $\mathcal{L}_{hinge}$-SVM \cite{cortes1995support}             & $\mathcal{L}_{pin}$-SVM \cite{huang2013support}           & $\mathcal{L}_{LINEX}$-SVM \cite{ma2019linex}         & $\mathcal{L}_{qtse}$-SVM \cite{zhao2022asymmetric} & $\mathcal{L}_{wave}$-SVM \cite{akhtar2024advancing} & $\mathcal{L}_{RoBoSS}$-SVM$^{\dagger}$ \\
 & (C, $\sigma$) & (C, $\sigma$, $\tau$) & (a, C, $\sigma$) & (a, C, $\sigma$) & (a, C, $\sigma$) & (a, $\lambda$, C, $\sigma$) \\ \hline
EEG\_f\_vs\_s\_bhattacharyya\_100 & 1, 0.1 & 0.1, 0.1, 0.5 & -1, 0.0001, 1 & -2, 0.00001, 0.0001 & -4, 0.00001, 1 & 3, 2, 0.00001, 1 \\
EEG\_f\_vs\_s\_entropy\_100 & 1, 0.1 & 0.1, 0.1, 0.5 & -1, 0.0001, 1 & -2, 0.00001, 0.0001 & -4, 0.00001, 1 & 3, 2, 0.00001, 1 \\
EEG\_f\_vs\_s\_roc\_50 & 1, 0.1 & 0.1, 0.1, 0.5 & -5, 0.00001, 1 & -1, 0.00001, 1 & 0, 0.00001, 1 & 0.3, 1.4, 0.000001, 1 \\
EEG\_f\_vs\_s\_ttest\_200 & 1, 0.1 & 0.1, 0.1, 0.5 & -3, 0.00001, 1 & -2, 0.00001, 0.0001 & 2, 0.00001, 1 & 3.2, 1.3, 0.000001, 1 \\
EEG\_f\_vs\_s\_wilcoxon\_100 & 1, 0.1 & 1, 0.1, 0 & -5, 0.00001, 1 & -2, 0.00001, 0.0001 & -4, 0.00001, 1 & 2.8, 1.8, 0.0001, 1 \\
EEG\_f\_vs\_s\_wilcoxon\_200 & 1, 0.1 & 1, 0.1, 0 & -5, 0.00001, 1 & -2, 0.00001, 0.0001 & 0, 0.00001, 1 & 0.5, 0.9, 0.0001, 1 \\
EEG\_n\_vs\_s\_bhattacharyya\_100 & 1, 0.1 & 1, 0.1, 0 & -7, 0.000001, 1 & -2, 1000000, 0.1 & 5, 0.00001, 1 & 0.4, 0.9, 10, 1 \\
EEG\_n\_vs\_s\_bhattacharyya\_200 & 1, 0.1 & 1, 0.1, 0 & -1, 0.000001, 1 & -2, 0.00001, 0.0001 & 4, 0.000001, 1 & 0.8, 1.5, 0.0001, 1 \\
EEG\_n\_vs\_s\_entropy\_100 & 1, 0.1 & 1, 0.1, 0 & -7, 0.000001, 1 & -2, 1000000, 0.1 & 5, 0.00001, 1 & 0.4, 0.9, 10, 1 \\
EEG\_n\_vs\_s\_entropy\_200 & 1, 0.1 & 1, 0.1, 0 & -1, 0.000001, 1 & -2, 0.00001, 0.0001 & 4, 0.000001, 1 & 0.8, 1.5, 0.0001, 1 \\
EEG\_n\_vs\_s\_ttest\_100 & 1, 0.1 & 0.1, 0.1, 0.5 & -2, 0.000001, 1 & -2, 0.00001, 0.0001 & -5, 0.000001, 1 & 0.7, 0.7, 0.01, 1 \\
EEG\_o\_vs\_f\_roc\_200 & 1, 0.1 & 1, 0.1, 0 & -5, 0.000001, 1 & -2, 0.00001, 0.0001 & -5, 0.0001, 0.1 & 0.7, 0.1, 100, 1 \\
EEG\_o\_vs\_f\_wilcoxon\_200 & 1, 0.1 & 0.1, 0.1, 0.5 & -5, 0.000001, 1 & -2, 0.00001, 0.0001 & -4, 0.00001, 1 & 1.1, 1.3, 10, 1 \\
EEG\_o\_vs\_n\_entropy\_200 & 1, 0.1 & 1, 0.1, 0 & -3, 0.0001, 1 & -2, 0.00001, 0.0001 & -4, 0.0001, 1 & 3.8, 1.9, 0.001, 1 \\
EEG\_o\_vs\_n\_entropy\_50 & 1, 0.1 & 1, 0.1, 0 & -1, 0.01, 1 & -2, 0.00001, 0.0001 & -2, 0.000001, 1 & 0.1, 1.4, 0.0001, 1 \\
EEG\_o\_vs\_n\_roc\_150 & 1, 0.1 & 0.1, 0.1, 0.5 & -3, 0.000001, 1 & -2, 0.00001, 0.0001 & 2, 0.000001, 1 & 4.4, 2, 0.1, 1 \\
EEG\_o\_vs\_n\_roc\_50 & 1, 0.1 & 1, 0.1, 0 & -1, 0.01, 1 & -2, 0.00001, 0.0001 & 5, 0.000001, 1 & 4.8, 0.4, 100, 1 \\
EEG\_o\_vs\_n\_ttest\_100 & 1, 0.1 & 1, 0.1, 0 & -3, 0.0001, 1 & -2, 0.00001, 0.0001 & 1, 0.00001, 1 & 3.8, 1.7, 0.0001, 1 \\
EEG\_o\_vs\_n\_ttest\_150 & 1, 0.1 & 0.1, 0.1, 0.5 & -5, 0.000001, 1 & -2, 0.00001, 0.0001 & 0, 0.000001, 1 & 4.9, 1.9, 0.000001, 1 \\
EEG\_o\_vs\_n\_ttest\_200 & 1, 0.1 & 1, 0.1, 0 & -2, 0.000001, 1 & -2, 0.00001, 0.0001 & -4, 0.0001, 1 & 4.2, 0.3, 100, 1 \\
EEG\_o\_vs\_n\_ttest\_50 & 1, 0.1 & 1, 0.1, 0 & -4, 0.00001, 1 & -5, 0.000001, 1 & -4, 0.0001, 1 & 4.5, 1.4, 0.000001, 1 \\
EEG\_o\_vs\_n\_wilcoxon\_50 & 1, 0.1 & 0.1, 0.1, 0.5 & -4, 0.00001, 1 & -5, 0.000001, 1 & 1, 0.00001, 1 & 1.2, 1.9, 1, 1 \\
EEG\_o\_vs\_s\_bhattacharyya\_100 & 1, 0.1 & 1, 0.1, 0 & -2, 0.000001, 1 & -2, 0.00001, 0.0001 & -5, 0.000001, 1 & 0.1, 0.3, 100, 1 \\
EEG\_o\_vs\_s\_bhattacharyya\_150 & 1, 0.1 & 1, 0.1, 0 & -6, 0.000001, 1 & -2, 0.00001, 0.0001 & -1, 0.000001, 1 & 3.8, 0.7, 1, 1 \\
EEG\_o\_vs\_s\_entropy\_100 & 1, 0.1 & 1, 0.1, 0 & -2, 0.000001, 1 & -2, 0.00001, 0.0001 & -5, 0.000001, 1 & 0.1, 0.3, 100, 1 \\
EEG\_o\_vs\_s\_entropy\_150 & 1, 0.1 & 1, 0.1, 0 & -6, 0.000001, 1 & -2, 0.00001, 0.0001 & -1, 0.000001, 1 & 3.8, 0.7, 1, 1 \\
EEG\_o\_vs\_s\_ttest\_150 & 1, 0.1 & 1, 0.1, 0 & -1, 0.00001, 1 & -2, 0.00001, 0.0001 & -4, 0.000001, 1 & 4, 1.2, 0.000001, 1 \\
EEG\_o\_vs\_s\_ttest\_200 & 1, 0.1 & 1, 0.1, 0 & -1, 0.000001, 1 & -2, 0.00001, 0.0001 & 4, 0.000001, 1 & 0.3, 2, 0.0001, 1 \\
EEG\_o\_vs\_s\_wilcoxon\_100 & 1, 0.1 & 0.1, 0.1, 0.5 & -7, 0.000001, 1 & -2, 0.00001, 0.0001 & -2, 0.000001, 1 & 0.1, 0.4, 1000, 1 \\
EEG\_z\_vs\_f\_roc\_150 & 1, 0.1 & 1, 0.1, 0 & -3, 0.000001, 1 & -2, 0.00001, 0.0001 & -5, 0.0001, 0.1 & 3.4, 1.6, 0.001, 1 \\
EEG\_z\_vs\_f\_ttest\_50 & 1, 0.1 & 0.1, 0.1, 0.5 & -7, 0.000001, 1 & -2, 0.00001, 0.0001 & -2, 0.000001, 1 & 3.4, 2, 10, 1 \\
EEG\_z\_vs\_f\_wilcoxon\_150 & 1, 0.1 & 1, 0.1, 0 & -3, 0.000001, 1 & -2, 0.00001, 0.0001 & -5, 0.0001, 0.1 & 3.4, 1.6, 0.001, 1 \\ \hline
\end{tabular}%
}
\end{table*}


\begin{table*}[]
\caption{The optimal parameters corresponding to the accuracy values of the proposed $\mathcal{L}_{RoBoSS}$-SVM and baseline models across each of the 16 BreaKHis datasets.}
\centering
\label{tab:BC-optimal-parameters}
\resizebox{17cm}{!}{%
\begin{tabular}{lllllll}
\hline
Dataset\textbackslash{}Model & $\mathcal{L}_{hinge}$-SVM \cite{cortes1995support}             & $\mathcal{L}_{pin}$-SVM \cite{huang2013support}           & $\mathcal{L}_{LINEX}$-SVM \cite{ma2019linex}         & $\mathcal{L}_{qtse}$-SVM \cite{zhao2022asymmetric} & $\mathcal{L}_{wave}$-SVM \cite{akhtar2024advancing} & $\mathcal{L}_{RoBoSS}$-SVM$^{\dagger}$ \\
 & (C, $\sigma$) & (C, $\sigma$, $\tau$) & (a, C, $\sigma$) & (a, C, $\sigma$) & (a, C, $\sigma$) & (a, $\lambda$, C, $\sigma$) \\ \hline
ADvsDC & 1, 1 & 1, 1, 0 & -6, 0.000001, 0.000001 & -1, 0.00001, 0.0001 & -5, 0.000001, 0.00001 & 3, 1.4, 0.0001, 0.1 \\
ADvsLC & 10, 1 & 1, 1, 0.5 & -1, 0.000001, 0.1 & -2, 100000, 0.000001 & 5, 0.0001, 1 & 0.4, 2, 0.00001, 1 \\
ADvsMC & 1, 1 & 1, 1, 0.5 & -6, 0.000001, 0.00001 & -1, 0.00001, 0.000001 & -5, 0.000001, 0.000001 & 4, 0.4, 0.000001, 0.1 \\
ADvsPC & 0.000001, 0.000001 & 1, 10, 0.3 & -3, 0.000001, 0.000001 & -2, 10000, 0.001 & -2, 0.00001, 0.00001 & 2, 1.3, 0.001, 1 \\
FDvsDC & 0.000001, 0.000001 & 0.000001, 0.000001, 0 & -5, 0.000001, 1 & -50, 0.000001, 0.000001 & 0, 0.000001, 1 & 0.3, 2, 0.001, 1 \\
FDvsLC & 0.1, 0.1 & 0.1, 0.1, 0 & -1, 0.01, 1 & -50, 0.000001, 0.000001 & -1, 0.00001, 1 & 1.8, 1, 0.1, 1 \\
FDvsMC & 1, 0.1 & 1, 0.1, 0 & -2, 0.001, 1 & -50, 0.000001, 0.000001 & 5, 0.000001, 1 & 0.3, 0.1, 0.01, 1 \\
FDvsPC & 0.000001, 0.000001 & 0.000001, 0.000001, 0 & -4, 0.00001, 1 & -50, 0.000001, 0.000001 & 1, 0.00001, 1 & 3, 2, 0.000001, 1 \\
PTvsDC & 10, 1 & 10, 1, 0 & -6, 0.000001, 0.00001 & -1, 0.00001, 0.000001 & -5, 0.000001, 0.00001 & 2.1, 1.8, 0.001, 0.1 \\
PTvsLC & 1, 1 & 1, 100000, 0.5 & -4, 0.000001, 1 & -2, 10000, 1 & 2, 0.000001, 1 & 2.3, 1.9, 0.0001, 1 \\
PTvsMC & 0.000001, 0.000001 & 0.000001, 0.000001, 0 & -6, 0.000001, 0.00001 & -1, 0.00001, 0.0001 & -5, 0.000001, 0.00001 & 0.5, 0.7, 0.00001, 0.01 \\
PTvsPC & 1, 1 & 1, 10, 0.5 & -2, 0.00001, 0.000001 & -2, 1000, 1 & -5, 0.00001, 0.1 & 1.3, 0.8, 0.000001, 1 \\
TAvsDC & 1, 1 & 1, 1, 0 & -6, 0.000001, 0.00001 & -1, 0.00001, 0.00001 & -5, 0.000001, 0.00001 & 1, 0.4, 0.1, 0.1 \\
TAvsLC & 0.01, 0.1 & 0.1, 100, 0.3 & -3, 0.00001, 1 & -2, 0.00001, 0.00001 & -3, 0.000001, 1 & 4.7, 0.3, 0.000001, 1 \\
TAvsMC & 10, 1 & 1, 1, 0.5 & -6, 0.000001, 0.1 & -2, 0.0001, 0.000001 & -5, 0.000001, 0.01 & 5, 0.9, 0.001, 0.1 \\
TAvsPC & 10000, 100 & 0.1, 10, 0.7 & -1, 0.0001, 0.00001 & -2, 0.1, 1 & -1, 0.000001, 0.00001 & 2.9, 0.8, 0.00001, 1 \\ \hline
\end{tabular}%
}
\end{table*}

\clearpage
\clearpage
\bibliographystyle{IEEEtranN}
\bibliography{refs.bib}